%% file: main.tex
\documentclass[]{amsart}
\usepackage{amsaddr}

\usepackage[hscale=.75]{geometry}

\usepackage[]{xcolor}
\usepackage[T1]{fontenc}

\usepackage{graphicx}

\usepackage[caption=false]{subfig}
\graphicspath{{./figures/}} % MUST be terminated by /

\usepackage{xparse}
\NewDocumentCommand\soi{}{\textsc{soi}}
\NewDocumentCommand\eg{}{\textit{e.g.}}

\NewDocumentCommand\afinn{}{\textsc{Afinn}}
\NewDocumentCommand\minos{}{\textsc{Minos}}
\NewDocumentCommand\vader{}{\textsc{Vader}}

\usepackage[colorlinks]{hyperref}

\usepackage[
    backend=biber,
    style=numeric-comp,
    sorting=none,
    sortcites=true
]{biblatex}
\usepackage[nameinlink,capitalise]{cleveref}

\title{Data Science Approaches to Evaluating Honours Candidates}
\thanks{This is an enhanced author version of \cite{10.1007/978-3-032-11442-6_23}: the version of record is available at \url{https://doi.org/10.1007/978-3-032-11442-6_23}.}

\author{Francesca von Braun-Bates}
\address{Ministry of Justice, London, United Kingdom}
\address{Joint Counter-Terrorism Prisons and Probation Hub, London, United Kingdom}
\email{francesca.von.braun-bates@justice.gov.uk}

\author{Sunreeta Sen}
\address{Arndit Ltd., Cambridge, United Kingdom}

\author{Indraayudh Talukdar}
\address{Indian Institute of Technology Delhi, New Delhi, India}

\author{Anirban Lahiri}
\address{Kainos, London, United Kingdom}

\begin{document}

\begin{abstract}
This paper introduces the first application of data science to the UK Honours system. We present a comprehensive Natural Language Processing methodology for evaluating public sentiment of Honours recipients.  In order to form an opinion about applicants for the UK King's Honours, we evaluated two existing sentiment algorithms (\afinn{} and \vader{}) and then developed our own novel algorithm, \minos{}.  The promising results in this work indicate that this system can be used to augment human evaluation to better judge whether a current recipient has maintained, or a prospective recipient is likely to maintain, the high standards of conduct demanded by the Honours system.  
Our novel approach is generalisable to any individual with a sufficient internet footprint and has applications in many fields including recruitment, national security and investigative journalism.
\end{abstract}

% In AMS document classes abstract preceeds maketitel
\maketitle

\input{Introduction}
\input{Background}
\input{Selection}
\input{Methodology}
\input{Results}
\input{conclusions_and_future_work}

\section*{Acknowledgements}

This research originated in the Data Science Accelerator programme, with advocacy from Stephanie Karpetas \textsc{obe}, Alec Waterhouse \textsc{fors} and Christalla Kyriacou.  We are grateful for the support of the Government Digital Service, the Office for National Statistics, the Government Operational Research Service, the Honours and Appointments Secretariat of the Cabinet Office, the (then) Department for Business, Energy and Industrial Strategy and the Ministry of Justice.

\printbibliography

\end{document}

%% file: introduction.tex
\section{Introduction}
\label{sec:introduction}

The Honours system has recognised individuals who have contributed exceptionally to  the United Kingdom for nearly a thousand years \cite{phillips-review-honours}.  This complex and ancient system needs to leverage cutting-edge technology to remain fair and transparent in the 21st century.  This paper is (to our knowledge) the first application of data science to the Honours System.  This paper demonstrates how combining a variety of natural language processing techniques creates a thorough and open-source intelligence picture from which to measure public opinion of Honours recipients.

This paper improves on the state of the art in four ways:
\begin{itemize}
     \item it uses a broad range of sources rather than purely social media or news articles;
     \item it illustrates a novel algorithm for identifying positive or negative personality traits as compared to two existing algorithms for sentiment analysis \cite{vader-paper,afinn-paper};
     \item we scrutinise the relevance of our input using co-reference resolution;
     \item we probe \soi{}s from many different fields and with varying degrees of celebrity (or non-celebrity) status
\end{itemize}
The individual techniques we apply to this problem are not new.  Web-scraping has existed in various forms since the earliest days of the internet \cite{web-scraping-history} including web crawling, web indexing and archiving of web content.  Tokenisation has been a core component of natural language processing for over 30 years (see \cite{mielke2021wordscharactersbriefhistory} for a history). Co-reference resolution is the task of associating different linguistic expressions which refer to the same entity \cite{hirst-thesis}, a linguistic problem well-known as early as Cicero in the first century \textsc{bc}.  Many sentiment analysis approaches exist (see \cite{comprehensive-setiment-analysis-techniques} for a selection).  We use two off-the-shelf models---\afinn{} \cite{afinn-blog} and \vader{} \cite{vader-paper}---before constructing our own algorithm, \minos{}, in \cref{sec:methodology}.  What is novel is our use of these techniques to solve a whole problem for users in a challenging context.

The structure of this paper is as follows.  In \cref{sec:background} we summarise the Honours system. We define our subjects of interest in \cref{sec:selection}.  Our methodology is described in \cref{sec:methodology}, including data collection, cleaning, tokenisation, co-reference resolution, sentiment analysis and extraction of results.  \cref{sec:results} shows the first advantage of web-scraping over manual research by examining the fraction of web-scraped text which survived our relevance criteria.  We summarise our sentiment results  in \cref{sec:results}.  We conclude in \cref{sec:conclusions} with a summary of our workflow and results, followed by a discussion of challenges addressed in future papers.

%% file: background.tex
\section{The Honours System}
\label{sec:background}

This section covers the essentials of the Honours System, in particular the concept of forfeiture.

The modern Honours System recognises individuals who have made an outstanding contribution to public life in the United Kingdom \cite{commons-library-paper-reviews} .  It encompasses diverse fields such as charity and voluntary service, education, science, arts, and business \cite{commons-library-paper-reviews}. The system bestows various levels of awards \cite{types-of-honours} to reflect a recipient's level of impact and time of sustained contributions to their field \cite{types-of-honours}.  Perhaps uniquely, individuals must maintain high standards of conduct \emph{after} receiving the Honour in order to retain it.  Anyone falling short of these standards risks having the Honour revoked (even posthumously), a process known as forfeiture \cite{forfeiture_guidance}.

Honours nominations are gathered from both public submissions and the Civil Service, pass through a series of sift committees, then undergo vetting before receiving final royal approval (details in \cite{honours-report-2008,honours-report-2011,honours-report-2013}).  This vetting consumes considerable resource to confirm that any nomination details are  factually correct, to conduct criminal record checks and probity checks with professional bodies and government departments \cite{forfeiture-cse-inquiry}.  

Maintaining public confidence in the Honours system requires award recipients to consistently uphold high standards. When an awardee's conduct brings the system into disrepute, the award may be forfeited \cite{forfeiture_guidance} or in rare cases, posthumously revoked \cite{forfeiture_posthumous_press_release}. Forfeiture is "almost certain" in the event of severe criminal convictions and/or professional disbarment \cite{forfeiture-cse-inquiry}.  Other causes for potential disrepute often result in nuanced and complex cases requiring careful consideration, \eg{} \cite{honours-report-2012-pt-1}.  Historically, awards were conferred post-career to minimize forfeiture risk \cite{honours-hmg-response-2005}.  More recently, awards focus on timely recognition of achievements \cite{honours-report-2008}, which may or may not be causally-connected to the recent increase in referrals to the Forfeiture Committee \cite{forfeiture-cse-inquiry}.

Forfeiture decisions are handled by an independent Forfeiture Committee, which lacks investigative powers \cite{honours-report-2012-pt-1}.  Instead it relies on evidence including official investigations and court proceedings \cite{forfeiture-cse-inquiry}.  In this high-risk environment, a clear and robust intelligence picture is essential to judge each case on its own merits.  Automated data-driven methods could significantly enhance proactive detection of potential misconduct, supporting timely interventions to uphold the integrity of the system.

%% file: selection.tex
\section{Selecting Subjects of Interest} % (fold)
\label{sec:selection}

To robustly test our novel approach, we selected two control groups as well as the forfeiture group:

\begin{itemize}
    \item ``Infamous'' people whose behaviour stands in direct contrast to enhancing public good;
    \item ``Awarded'' subjects who have received an Honour and retained it since, suggesting continued behaviour which supports the public good;
    \item ``Forfeited'' subjects who forfeited their Honour, or who were posthumously stripped of the Honour by the Committee.
\end{itemize}

We selected twenty \soi{}s per group.

The awarded and forfeited groups were selected from the announcements of the London Gazette \cite{gazette-honours-lists-1940-2022}.  We selected a range of award levels, from the highest (Knight/Dame) to the lowest (Member) \cite{types-of-honours}.  We were careful to include a broad range of fields and achievements across the ten Honours Committees, ranging from the arts and sciences to business, sport and civil service \cite{everything-you-need-to-know-nomination}.  We also used a mixture of male and female recipients approximating the underlying gender split.  The differentiating factor between them is that --- unlike the ``awarded'' group --- all forfeited \soi{}s were stripped of their Honours for poor conduct.  This level of poor conduct varies greatly in degree of criminality, the length and nature of the offence, and the delay between award and forfeiture.

The infamous group contains \soi{}s who far exceed the ``criminal conviction'' threshold for forfeiture.  We selected individuals whose crimes (and, where possible, convictions) are in the public domain.  Again we selected a mixture of male and female individuals.

We expect a well-behaved algorithm and a well-constructed method to clearly distinguish between each group:
\begin{itemize}
  \item Highly positive for the awarded group
  \item Approaching zero for the forfeited group
  \item Highly negative for the infamous group
\end{itemize}
Without this distinction, we cannot invert the mapping for new \soi{}.  In other words, we cannot train the algorithm on this group of known individuals to accurately classify any \soi{} in a test set.

% subsection selecting_subjects_of_interest (end)

Now we have a specific problem to solve and a set of subjects on which to test our solution.  We detail our approach to solving this problem in \cref{sec:methodology}.

%% file: methodology.tex
\section{Methodology}
\label{sec:methodology}

\begin{figure}
  \centering
\includegraphics[height=.9\textheight]{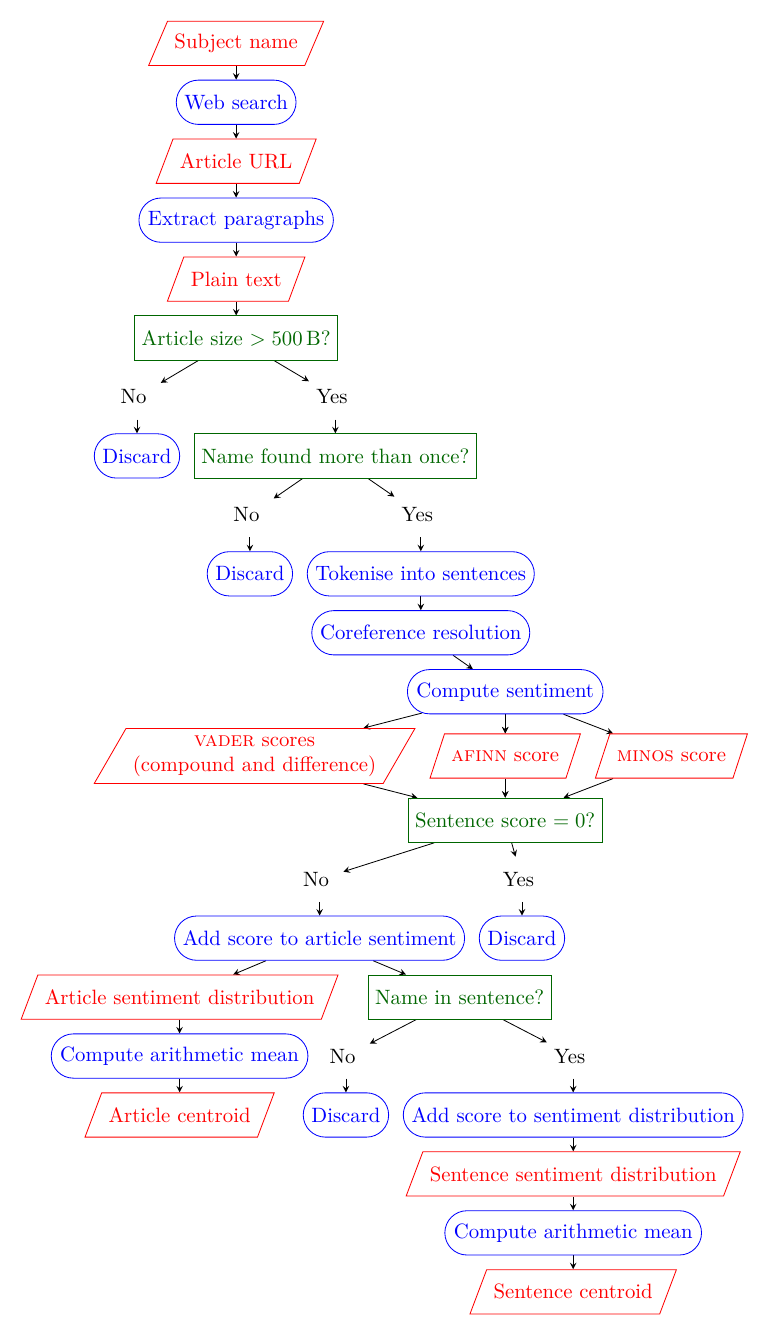}
\caption{Control flow of our methodology.  Trapezia represent outputs (except the starting input), rectangles are binary decisions and rounded rectangles are outcomes.} \label{fig:control_flow}
\end{figure}

This section covers the key stages in our algorithm, summarised in \cref{fig:control_flow}.

\subsection{Data Collection, Processing and Filtering}

The initial phase of our process involves automating online searches to collect publicly accessible data about candidates. To simulate human research, we employed the Selenium web-driver \cite{selenium} to automatically query Google with candidates' full names, extracted plain text from the results using BeautifulSoup \cite{richardson2007beautiful} and preserved the original sources and their \textsc{url}s to ensure traceability and auditability.

We encountered several limitations: the inability to access paywalled or subscription-based content, some pages blocked by compliance with a site's \texttt{robots.txt}, and incomplete retrieval of dynamically-generated JavaScript content.  Nonetheless, the top 60-70 search results returned sufficient text for our proof-of-concept.

We separated the plain text into sentences using \textsc{nltk} \cite{nltk}, which effectively handled linguistic complexities such as abbreviations, ellipses, and unconventional punctuation, ensuring robust sentence-level processing for subsequent analysis.

We discarded articles which were either under 500 characters long, or did not contain the \soi{}'s "best-known full name" (not always equivalent to the \emph{entire}, legal full name used in the search). This filtering strategy adeptly managed issues such as multi-part surnames, middle names, peerages, titles, and informal name variants.

The resulting dataset provided a focused and relevant base for sentiment analysis, significantly reducing noise and increasing the accuracy of subsequent analytical steps.

\subsection{Co-reference Resolution}

Co-reference resolution was applied to address the frequent occurrence of indirect candidate mentions in textual data. This critical step significantly improved the reliability and accuracy of sentiment analysis by ensuring appropriate attribution of content to specific candidates.

Using a supervised learning-based Python algorithm trained on extensive English text collections, our approach systematically identified and linked pronouns and abbreviated references back to the candidates' full names. In case there are multiple people having the same name and surname combination, some additional distinguishing attributes can be used, which would further help disambiguation during coreference resolution, which was extensively validated. 

\subsection{Sentiment Algorithms}

Our sentiment analysis involved evaluating multiple algorithms to determine their suitability for accurately assessing candidate-related text. Initially, we assessed the lexical approach \afinn{}, straightforwardly assigning integer scores from -5 to +5 for individual words but limited by a lack of contextual understanding. Next, we explored \vader{}, a rule-based model optimized for informal social media texts, adept at handling intensifiers and idioms but still challenged by formal content.

Recognizing these limitations, we developed \minos{}, a tailored sentiment algorithm explicitly designed for the honours evaluation context. \minos{} incorporated comprehensive lexicons of positive and negative words, strongly emphasizing terms related to misconduct and criminality. The initial lexicons were taken from \afinn{} and \vader{} algorithms and then augmented with words often used to describe individuals either positively or negatively in news articles, assessed by linguists and other domain experts \cite{MinosDataSetl2025}.
Sentences containing negative terms automatically received negative scores irrespective of any positive language, effectively mirroring the Honours system’s stringent standards on recipient conduct.
Describing in detail the scoring and decision-making rules implemented in the MINOS algorithm is beyond the scope of this paper and we refer the reader to the code for details \cite{MinosCode2025}.

\subsection{Posterior Marginalisation}

Sentiment analysis produced a multi-modal distribution averaged over all surviving text for each \soi{}.  We integrated over these distributions to produce a centroid to convey the "average" sentiment per individual.

Although these can be compared for the same algorithm, we cannot compare between algorithms.  This is because the different approaches have different (and sometimes unbounded) priors which cannot all be normalised to the unit interval.

The final output also identified and highlighted sentences with significant positive or negative sentiment and the source \textsc{url}.  Thus an Honours sifter had access to the summary centroid, the full posterior, outlying values and the number of articles and sentences which survived filtering from the original web search.  This approach facilitated efficient human evaluation, enabling committees to quickly verify the accuracy, relevance, and reliability of presented evidence, ensuring decisions remained informed, reproducible, and auditable.

%% file: results.tex
\section{Results, Analysis and Discussion}
\label{sec:results}

\subsection{Relevance Filtering}

\begin{figure}
  \centering
  \subfloat[Awarded]{
    \label{f:file-size-awarded}
    \resizebox{.6\textwidth}{!}{%
      \input{figures/bar_file_size_by_soi_awarded}
    }
  }  \\
  \subfloat[Forfeited]{
    \label{f:file-size-forfeited}
    \resizebox{.6\textwidth}{!}{%
      \input{figures/bar_file_size_by_soi_forfeited}
    }
  } \\
  \subfloat[Infamous]{
    \label{f:file-size-infamous}
    \resizebox{.6\textwidth}{!}{%
      \input{figures/bar_file_size_by_soi_infamous}
    }
  }
  \caption{File size distribution per \soi{} group.  The colour scale represents the (logarithmically-binned) file size in bytes, with the height of each stack is the number of articles per \soi{}.}
  \label{fig:hist-file-size-by-soi}
\end{figure}
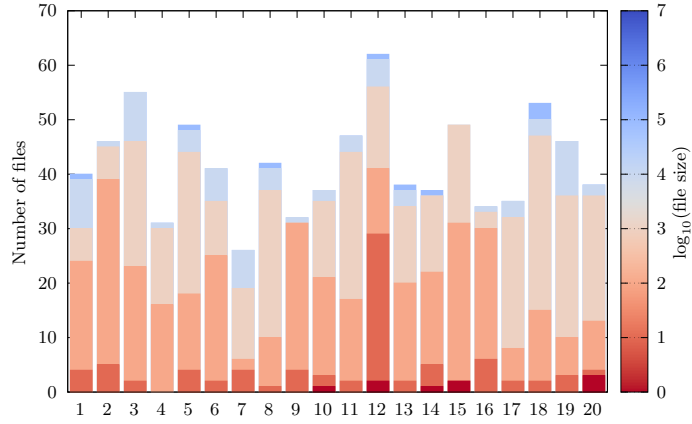
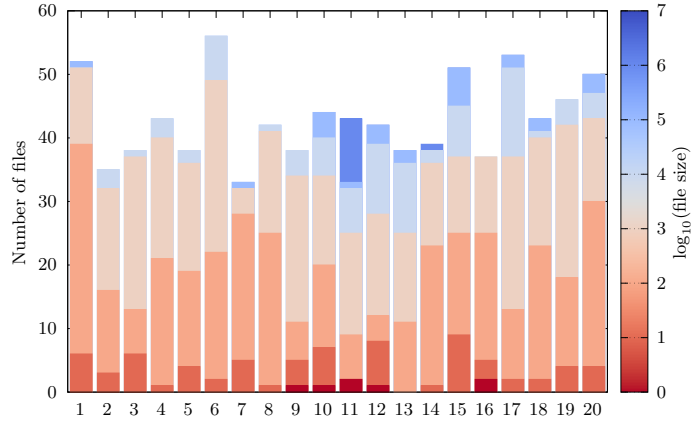
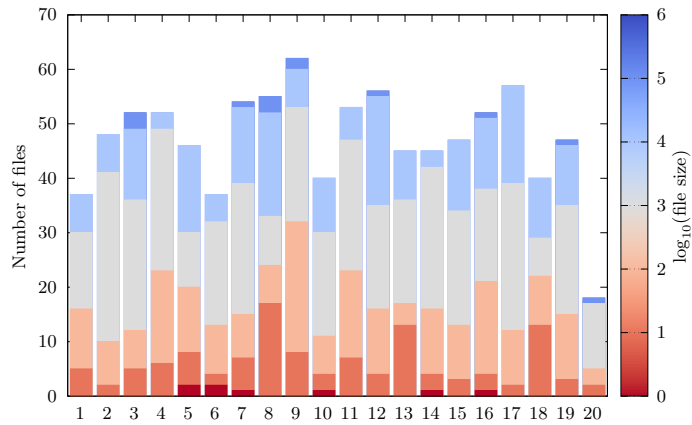

Filtering web-sourced articles is essential to retain only relevant content for each individual. As illustrated in \cref{fig:hist-file-size-by-soi} article file sizes vary by seven orders of magnitude.  Short files (dark red) were largely spurious browser messages.  Extremely large files (up to $10^7$ characters) swamped any signal about the desired \soi{} with noise about other individuals and topics. The most meaningful content usually falls within the mid-size range (1 kB – 1 MB), which balances depth and relevance. 

This distribution highlights the importance of machine-assisted filtering, as the volume of material far exceeds what a human analyst could process, ensuring accuracy in sentiment analysis and interpretation.

\subsection{Marginalised Centroids}

This section compares centroid results from different sentiment analysis algorithms applied to individuals across the awarded, forfeited, and infamous groups.

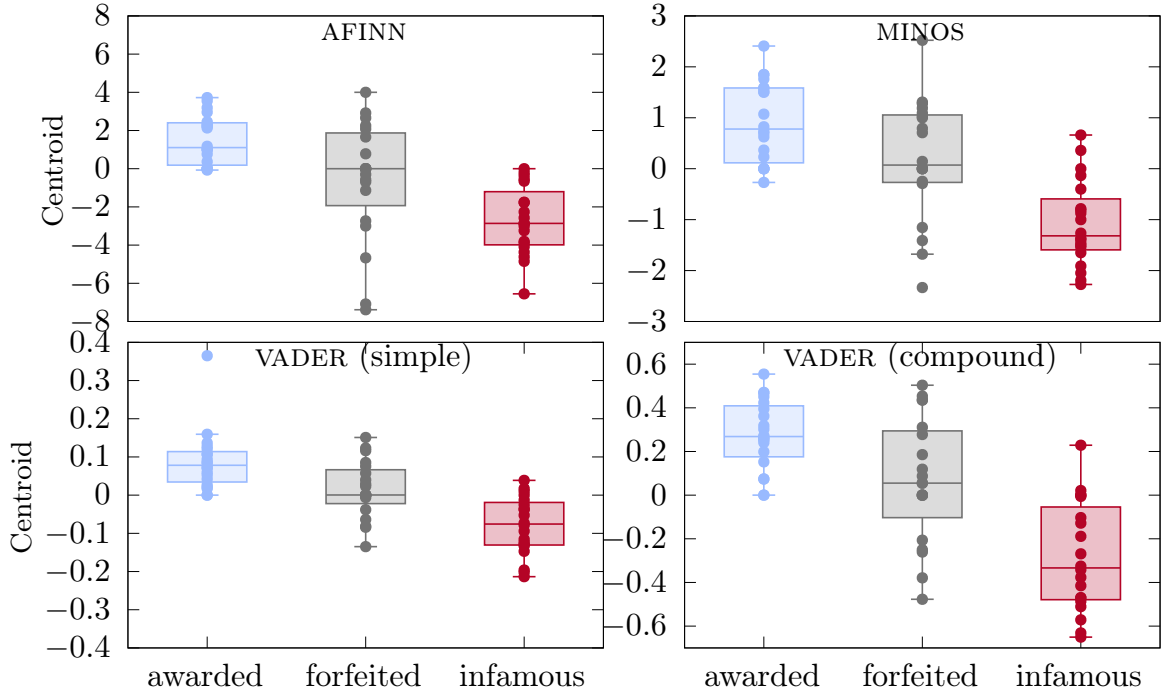
\begin{figure}
  \centering
  \resizebox{.95\textwidth}{!}{%
  \input{figures/centroid_boxplot_03_coref_resolved_data_sentences_filtered}}
  \caption{Box plot of centroid distributions from coreference-resolved data, including only sentences which contain the \soi{} name.}
  \label{fig:centroids_coref_sentences}
\end{figure}

The results focus only on sentences that explicitly reference the individual once the co-reference resolution has been performed.  This filters out as much noise as possible---which we know is not attributable to the \soi{}---while retaining as much information as possible thanks to co-reference resolution.  We have four results (one per sentiment algorithm) for each of the three \soi{} groups.  \cref{fig:centroids_coref_sentences} shows box plots, wherein each datum is the centroid for one \soi{} in that group.  The boxes show the inter-quartile range of the \soi{} group, with the mean shown as a horizontal line.  Red, gray and blue show infamous, forfeited and awarded groups respectively.

Generally, our results demonstrate the desirable properties defined in \cref{sec:selection}.  Independent of the sentiment algorithm, the arithmetic means and inter-quartile ranges are clearly positive for awarded, clearly negative for infamous and span zero for the forfeited groups.  Some undesirable results also occur, \eg{} only \vader{}'s compound score returns a positive value for all awarded \soi{}, whereas no algorithm returns a negative score for all infamous \soi{}.   The prevalence of outliers makes assigning an individual to a group based on the centroid a difficult prospect for a classification or clustering algorithm.  Only \minos{} shows a clear distinction between the infamous group's inter-quartile range and the other two groups, which do not overlap it at all.  These mixed results suggest that we have integrated over important detail in the posterior distributions to arrive at a scalar summary result.

\subsection{Selected Posteriors}

\begin{figure}
\resizebox{.49\textwidth}{!}{\input{figures/bhattacharyya_sushantha_kumar/afinn}}
\resizebox{.49\textwidth}{!}{\input{figures/bhattacharyya_sushantha_kumar/minos}}
\\
\resizebox{.49\textwidth}{!}{\input{figures/bhattacharyya_sushantha_kumar/vader_compound}}
\resizebox{.49\textwidth}{!}{\input{figures/bhattacharyya_sushantha_kumar/vader_simple}} 
\caption{Sentiment posteriors for Kumar Bhattacharyya.}
\label{fig:sentiment_posteriors_for_kumar_bhattacharyya}
\end{figure}
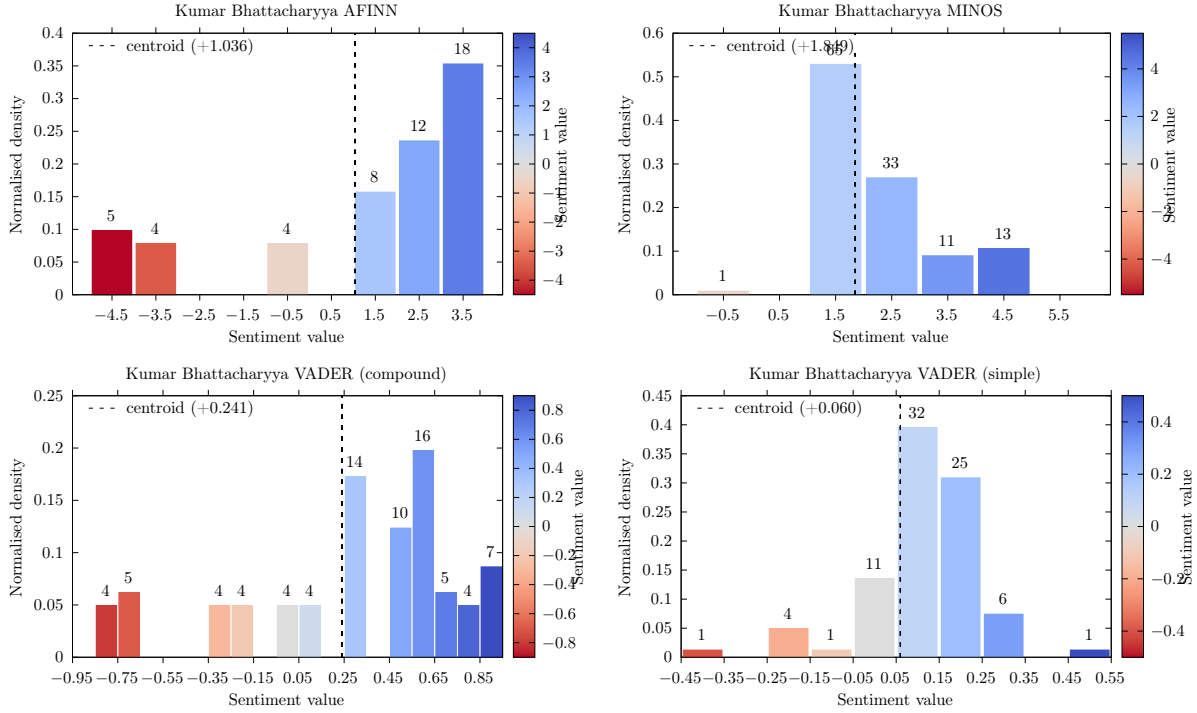

\begin{figure}
\resizebox{.49\textwidth}{!}{\input{figures/charles_sobhraj/afinn}}
\resizebox{.49\textwidth}{!}{\input{figures/charles_sobhraj/minos}}
\\
\resizebox{.49\textwidth}{!}{\input{figures/charles_sobhraj/vader_compound}}
\resizebox{.49\textwidth}{!}{\input{figures/charles_sobhraj/vader_simple}} 
\caption{Sentiment posteriors for Charles Sobhraj.}
\label{fig:sentiment_posteriors_for_charles_sobhraj}
\end{figure}
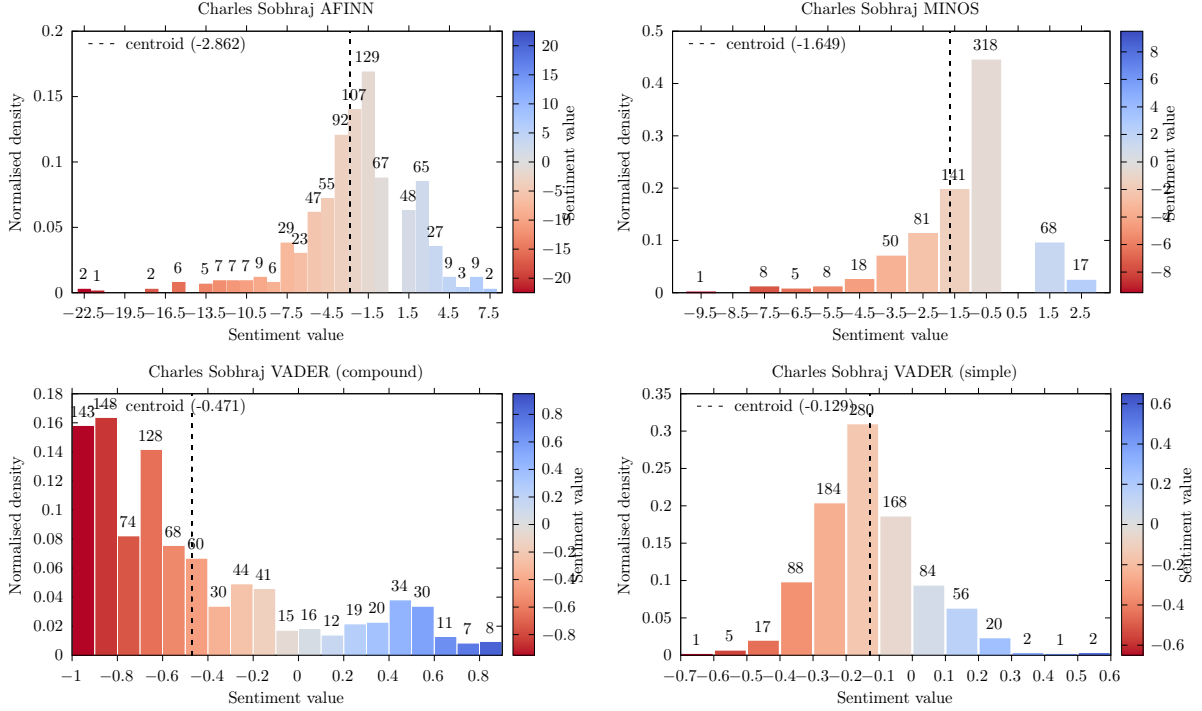

\begin{figure}
\resizebox{.49\textwidth}{!}{\input{figures/rolf_harris/afinn}}
\resizebox{.49\textwidth}{!}{\input{figures/rolf_harris/minos}}
\\
\resizebox{.49\textwidth}{!}{\input{figures/rolf_harris/vader_compound}}
\resizebox{.49\textwidth}{!}{\input{figures/rolf_harris/vader_simple}} 
\caption{Sentiment posteriors for Rolf Harris.}
\label{fig:sentiment_posteriors_for_rolf_harris}
\end{figure}
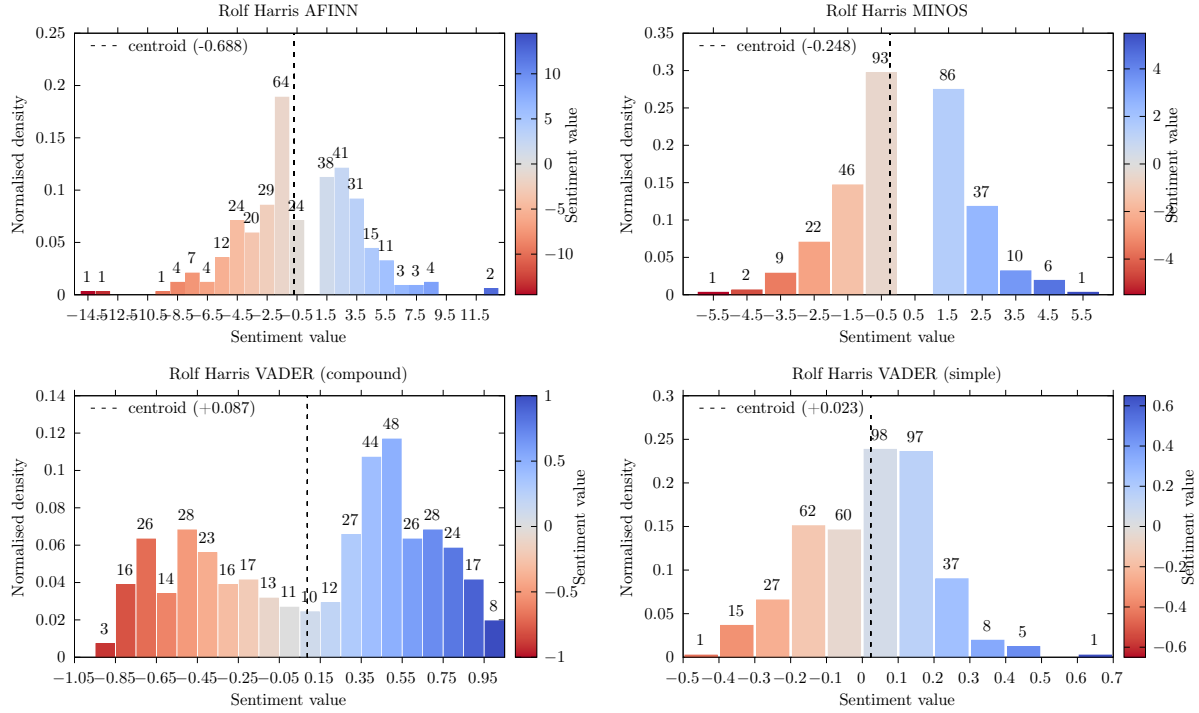

This section shows the distribution functions for individual \soi{}.  We selected one from each group as an illustration of a typical posterior:
\begin{enumerate}
  \item Kumar Bhattacharyya for awarded in \cref{fig:sentiment_posteriors_for_kumar_bhattacharyya}
  \item Rolf Harris for forfeited in \cref{fig:sentiment_posteriors_for_rolf_harris}
  \item Charles Sobhraj for infamous in \cref{fig:sentiment_posteriors_for_charles_sobhraj}
\end{enumerate}
All histograms integrate to unity, with the the number of sentences in each bin shown above the bar for that bin.  Red results are negative, whereas blue are positive.  The zero bin (which would be grey) is excluded by our prior distribution to avoid washing out the results to zero everywhere else in the parameter space.

\cref{fig:sentiment_posteriors_for_kumar_bhattacharyya} shows broadly positive sentiment for our awarded \soi{}.  The centroids are positive for all sentiment algorithms.  However, the shape of the distributions differs.  Only \minos{} delivers a consistently positive result in which the centroid is firmly in the mode of the distribution, with few negative sentences.  The other three algorithms show a long negative tail which we would not expect from an awarded \soi{}. 

For infamous individuals, \cref{fig:sentiment_posteriors_for_charles_sobhraj} shows negative centroids independent of the sentiment algorithm.  Three distributions have similar shapes, apart from \vader{} compound which is strongly peaked at the negative extremum.  Although this last algorithm seems desirable, it also has some extremely positive results (over $80\%$ positive), which seem inexplicable for a prolific serial killer.  In contrast, the other distributions have a much longer negative tail than their positive component, which agrees with the logic for this group.  Thus in this group, our example shows that three of the four distribution have desirable properties throughout the posterior as well as in the summary centroid.

\cref{fig:sentiment_posteriors_for_rolf_harris} shows the mixture of positive and negative sentiment expected of a forfeited \soi{}.  Whereas \afinn{} and \minos{} produce negative centroids, \vader{} produces neutral and slightly positive centroids.  All the distributions have different shapes: \vader{} simple peaks in the zero bin.  Since all zero scores are removed by convolving with the prior, this only occurs when a sentence has elements of both positive and negative sentiment which "cancels out" at close to zero (rather than a neutral sentence without any sentiment).  This is a nuance which is potentially misleading at first glance, so this algorithm is undesirable.  The compound \vader{} score shows a bimodal distribution which is strongly skewed towards the positive.  Thus it picks up Harris' positive actions (which may have contributed towards the award of his \textsc{cbe}), but it fails to adequately convey the prolifically-publicised offences for which his \textsc{cbe} was forfeited.  Both \afinn{} and \minos{} capture a range of sentiment, some extremes as well as mixed sentiment.  However, \minos{} delivers a clearer positive peak, resulting in a more balanced distribution, whereas \afinn{} captures the positive side less clearly.   Apart from \vader{} compound, the centroids are all in or very close to the edge of the mode bin, which indicates that the summary is a reliable approximation of the whole distribution.

Across our \soi{} groups, \minos{} delivers the results which are most consistent with our prior knowledge of the \soi{}.  This makes it the most promising algorithm for wider application.

%% file: figures/bar_file_size_by_soi_awarded.tex
% GNUPLOT: LaTeX picture with Postscript
\begingroup
\small
  \makeatletter
  \providecommand\color[2][]{%
    \GenericError{(gnuplot) \space\space\space\@spaces}{%
      Package color not loaded in conjunction with
      terminal option `colourtext'%
    }{See the gnuplot documentation for explanation.%
    }{Either use 'blacktext' in gnuplot or load the package
      color.sty in LaTeX.}%
    \renewcommand\color[2][]{}%
  }%
  \providecommand\includegraphics[2][]{%
    \GenericError{(gnuplot) \space\space\space\@spaces}{%
      Package graphicx or graphics not loaded%
    }{See the gnuplot documentation for explanation.%
    }{The gnuplot epslatex terminal needs graphicx.sty or graphics.sty.}%
    \renewcommand\includegraphics[2][]{}%
  }%
  \providecommand\rotatebox[2]{#2}%
  \@ifundefined{ifGPcolor}{%
    \newif\ifGPcolor
    \GPcolortrue
  }{}%
  \@ifundefined{ifGPblacktext}{%
    \newif\ifGPblacktext
    \GPblacktextfalse
  }{}%
  % define a \g@addto@macro without @ in the name:
  \let\gplgaddtomacro\g@addto@macro
  % define empty templates for all commands taking text:
  \gdef\gplbacktext{}%
  \gdef\gplfronttext{}%
  \makeatother
  \ifGPblacktext
    % no textcolor at all
    \def\colorrgb#1{}%
    \def\colorgray#1{}%
  \else
    % gray or color?
    \ifGPcolor
      \def\colorrgb#1{\color[rgb]{#1}}%
      \def\colorgray#1{\color[gray]{#1}}%
      \expandafter\def\csname LTw\endcsname{\color{white}}%
      \expandafter\def\csname LTb\endcsname{\color{black}}%
      \expandafter\def\csname LTa\endcsname{\color{black}}%
      \expandafter\def\csname LT0\endcsname{\color[rgb]{1,0,0}}%
      \expandafter\def\csname LT1\endcsname{\color[rgb]{0,1,0}}%
      \expandafter\def\csname LT2\endcsname{\color[rgb]{0,0,1}}%
      \expandafter\def\csname LT3\endcsname{\color[rgb]{1,0,1}}%
      \expandafter\def\csname LT4\endcsname{\color[rgb]{0,1,1}}%
      \expandafter\def\csname LT5\endcsname{\color[rgb]{1,1,0}}%
      \expandafter\def\csname LT6\endcsname{\color[rgb]{0,0,0}}%
      \expandafter\def\csname LT7\endcsname{\color[rgb]{1,0.3,0}}%
      \expandafter\def\csname LT8\endcsname{\color[rgb]{0.5,0.5,0.5}}%
    \else
      % gray
      \def\colorrgb#1{\color{black}}%
      \def\colorgray#1{\color[gray]{#1}}%
      \expandafter\def\csname LTw\endcsname{\color{white}}%
      \expandafter\def\csname LTb\endcsname{\color{black}}%
      \expandafter\def\csname LTa\endcsname{\color{black}}%
      \expandafter\def\csname LT0\endcsname{\color{black}}%
      \expandafter\def\csname LT1\endcsname{\color{black}}%
      \expandafter\def\csname LT2\endcsname{\color{black}}%
      \expandafter\def\csname LT3\endcsname{\color{black}}%
      \expandafter\def\csname LT4\endcsname{\color{black}}%
      \expandafter\def\csname LT5\endcsname{\color{black}}%
      \expandafter\def\csname LT6\endcsname{\color{black}}%
      \expandafter\def\csname LT7\endcsname{\color{black}}%
      \expandafter\def\csname LT8\endcsname{\color{black}}%
    \fi
  \fi
    \setlength{\unitlength}{0.0500bp}%
    \ifx\gptboxheight\undefined%
      \newlength{\gptboxheight}%
      \newlength{\gptboxwidth}%
      \newsavebox{\gptboxtext}%
    \fi%
    \setlength{\fboxrule}{0.5pt}%
    \setlength{\fboxsep}{1pt}%
    \definecolor{tbcol}{rgb}{1,1,1}%
\begin{picture}(7200.00,4320.00)%
    \gplgaddtomacro\gplbacktext{%
      \csname LTb\endcsname%%
      \put(518,351){\makebox(0,0)[r]{\strut{}$0$}}%
      \csname LTb\endcsname%%
      \put(518,890){\makebox(0,0)[r]{\strut{}$10$}}%
      \csname LTb\endcsname%%
      \put(518,1429){\makebox(0,0)[r]{\strut{}$20$}}%
      \csname LTb\endcsname%%
      \put(518,1968){\makebox(0,0)[r]{\strut{}$30$}}%
      \csname LTb\endcsname%%
      \put(518,2507){\makebox(0,0)[r]{\strut{}$40$}}%
      \csname LTb\endcsname%%
      \put(518,3046){\makebox(0,0)[r]{\strut{}$50$}}%
      \csname LTb\endcsname%%
      \put(518,3585){\makebox(0,0)[r]{\strut{}$60$}}%
      \csname LTb\endcsname%%
      \put(518,4124){\makebox(0,0)[r]{\strut{}$70$}}%
      \csname LTb\endcsname%%
      \put(749,175){\makebox(0,0){\strut{}$1$}}%
      \csname LTb\endcsname%%
      \put(1016,175){\makebox(0,0){\strut{}$2$}}%
      \csname LTb\endcsname%%
      \put(1283,175){\makebox(0,0){\strut{}$3$}}%
      \csname LTb\endcsname%%
      \put(1550,175){\makebox(0,0){\strut{}$4$}}%
      \csname LTb\endcsname%%
      \put(1817,175){\makebox(0,0){\strut{}$5$}}%
      \csname LTb\endcsname%%
      \put(2084,175){\makebox(0,0){\strut{}$6$}}%
      \csname LTb\endcsname%%
      \put(2351,175){\makebox(0,0){\strut{}$7$}}%
      \csname LTb\endcsname%%
      \put(2618,175){\makebox(0,0){\strut{}$8$}}%
      \csname LTb\endcsname%%
      \put(2885,175){\makebox(0,0){\strut{}$9$}}%
      \csname LTb\endcsname%%
      \put(3152,175){\makebox(0,0){\strut{}$10$}}%
      \csname LTb\endcsname%%
      \put(3419,175){\makebox(0,0){\strut{}$11$}}%
      \csname LTb\endcsname%%
      \put(3686,175){\makebox(0,0){\strut{}$12$}}%
      \csname LTb\endcsname%%
      \put(3953,175){\makebox(0,0){\strut{}$13$}}%
      \csname LTb\endcsname%%
      \put(4220,175){\makebox(0,0){\strut{}$14$}}%
      \csname LTb\endcsname%%
      \put(4487,175){\makebox(0,0){\strut{}$15$}}%
      \csname LTb\endcsname%%
      \put(4754,175){\makebox(0,0){\strut{}$16$}}%
      \csname LTb\endcsname%%
      \put(5021,175){\makebox(0,0){\strut{}$17$}}%
      \csname LTb\endcsname%%
      \put(5288,175){\makebox(0,0){\strut{}$18$}}%
      \csname LTb\endcsname%%
      \put(5555,175){\makebox(0,0){\strut{}$19$}}%
      \csname LTb\endcsname%%
      \put(5822,175){\makebox(0,0){\strut{}$20$}}%
    }%
    \gplgaddtomacro\gplfronttext{%
      \csname LTb\endcsname%%
      \put(161,2237){\rotatebox{-270}{\makebox(0,0){\strut{}Number of files}}}%
      \csname LTb\endcsname%%
      \put(6454,351){\makebox(0,0)[l]{\strut{}$0$}}%
      \csname LTb\endcsname%%
      \put(6454,890){\makebox(0,0)[l]{\strut{}$1$}}%
      \csname LTb\endcsname%%
      \put(6454,1429){\makebox(0,0)[l]{\strut{}$2$}}%
      \csname LTb\endcsname%%
      \put(6454,1968){\makebox(0,0)[l]{\strut{}$3$}}%
      \csname LTb\endcsname%%
      \put(6454,2507){\makebox(0,0)[l]{\strut{}$4$}}%
      \csname LTb\endcsname%%
      \put(6454,3046){\makebox(0,0)[l]{\strut{}$5$}}%
      \csname LTb\endcsname%%
      \put(6454,3585){\makebox(0,0)[l]{\strut{}$6$}}%
      \csname LTb\endcsname%%
      \put(6454,4124){\makebox(0,0)[l]{\strut{}$7$}}%
      \csname LTb\endcsname%%
      \put(6698,2237){\rotatebox{-270}{\makebox(0,0){\strut{}$\log_{10}$(file size)}}}%
    }%
    \gplbacktext
    \put(0,0){\includegraphics[width={360.00bp},height={216.00bp}]{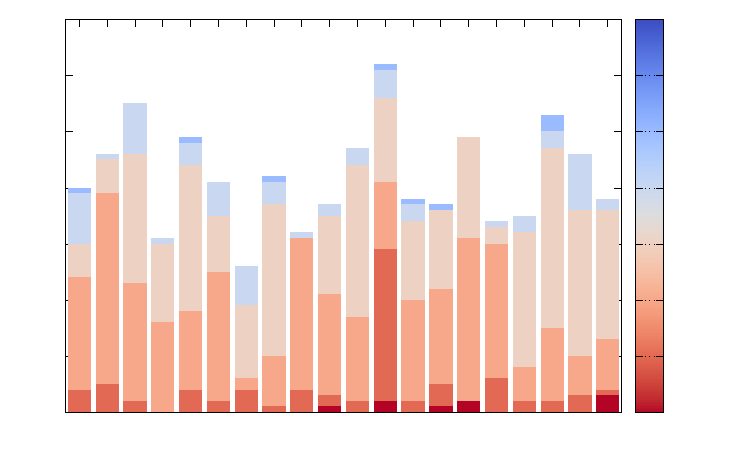}}%
    \gplfronttext
  \end{picture}%
\endgroup

%% file: figures/bar_file_size_by_soi_forfeited.tex
% GNUPLOT: LaTeX picture with Postscript
\begingroup
\small
  \makeatletter
  \providecommand\color[2][]{%
    \GenericError{(gnuplot) \space\space\space\@spaces}{%
      Package color not loaded in conjunction with
      terminal option `colourtext'%
    }{See the gnuplot documentation for explanation.%
    }{Either use 'blacktext' in gnuplot or load the package
      color.sty in LaTeX.}%
    \renewcommand\color[2][]{}%
  }%
  \providecommand\includegraphics[2][]{%
    \GenericError{(gnuplot) \space\space\space\@spaces}{%
      Package graphicx or graphics not loaded%
    }{See the gnuplot documentation for explanation.%
    }{The gnuplot epslatex terminal needs graphicx.sty or graphics.sty.}%
    \renewcommand\includegraphics[2][]{}%
  }%
  \providecommand\rotatebox[2]{#2}%
  \@ifundefined{ifGPcolor}{%
    \newif\ifGPcolor
    \GPcolortrue
  }{}%
  \@ifundefined{ifGPblacktext}{%
    \newif\ifGPblacktext
    \GPblacktextfalse
  }{}%
  % define a \g@addto@macro without @ in the name:
  \let\gplgaddtomacro\g@addto@macro
  % define empty templates for all commands taking text:
  \gdef\gplbacktext{}%
  \gdef\gplfronttext{}%
  \makeatother
  \ifGPblacktext
    % no textcolor at all
    \def\colorrgb#1{}%
    \def\colorgray#1{}%
  \else
    % gray or color?
    \ifGPcolor
      \def\colorrgb#1{\color[rgb]{#1}}%
      \def\colorgray#1{\color[gray]{#1}}%
      \expandafter\def\csname LTw\endcsname{\color{white}}%
      \expandafter\def\csname LTb\endcsname{\color{black}}%
      \expandafter\def\csname LTa\endcsname{\color{black}}%
      \expandafter\def\csname LT0\endcsname{\color[rgb]{1,0,0}}%
      \expandafter\def\csname LT1\endcsname{\color[rgb]{0,1,0}}%
      \expandafter\def\csname LT2\endcsname{\color[rgb]{0,0,1}}%
      \expandafter\def\csname LT3\endcsname{\color[rgb]{1,0,1}}%
      \expandafter\def\csname LT4\endcsname{\color[rgb]{0,1,1}}%
      \expandafter\def\csname LT5\endcsname{\color[rgb]{1,1,0}}%
      \expandafter\def\csname LT6\endcsname{\color[rgb]{0,0,0}}%
      \expandafter\def\csname LT7\endcsname{\color[rgb]{1,0.3,0}}%
      \expandafter\def\csname LT8\endcsname{\color[rgb]{0.5,0.5,0.5}}%
    \else
      % gray
      \def\colorrgb#1{\color{black}}%
      \def\colorgray#1{\color[gray]{#1}}%
      \expandafter\def\csname LTw\endcsname{\color{white}}%
      \expandafter\def\csname LTb\endcsname{\color{black}}%
      \expandafter\def\csname LTa\endcsname{\color{black}}%
      \expandafter\def\csname LT0\endcsname{\color{black}}%
      \expandafter\def\csname LT1\endcsname{\color{black}}%
      \expandafter\def\csname LT2\endcsname{\color{black}}%
      \expandafter\def\csname LT3\endcsname{\color{black}}%
      \expandafter\def\csname LT4\endcsname{\color{black}}%
      \expandafter\def\csname LT5\endcsname{\color{black}}%
      \expandafter\def\csname LT6\endcsname{\color{black}}%
      \expandafter\def\csname LT7\endcsname{\color{black}}%
      \expandafter\def\csname LT8\endcsname{\color{black}}%
    \fi
  \fi
    \setlength{\unitlength}{0.0500bp}%
    \ifx\gptboxheight\undefined%
      \newlength{\gptboxheight}%
      \newlength{\gptboxwidth}%
      \newsavebox{\gptboxtext}%
    \fi%
    \setlength{\fboxrule}{0.5pt}%
    \setlength{\fboxsep}{1pt}%
    \definecolor{tbcol}{rgb}{1,1,1}%
\begin{picture}(7200.00,4320.00)%
    \gplgaddtomacro\gplbacktext{%
      \csname LTb\endcsname%%
      \put(518,351){\makebox(0,0)[r]{\strut{}$0$}}%
      \csname LTb\endcsname%%
      \put(518,980){\makebox(0,0)[r]{\strut{}$10$}}%
      \csname LTb\endcsname%%
      \put(518,1609){\makebox(0,0)[r]{\strut{}$20$}}%
      \csname LTb\endcsname%%
      \put(518,2237){\makebox(0,0)[r]{\strut{}$30$}}%
      \csname LTb\endcsname%%
      \put(518,2866){\makebox(0,0)[r]{\strut{}$40$}}%
      \csname LTb\endcsname%%
      \put(518,3495){\makebox(0,0)[r]{\strut{}$50$}}%
      \csname LTb\endcsname%%
      \put(518,4124){\makebox(0,0)[r]{\strut{}$60$}}%
      \csname LTb\endcsname%%
      \put(749,175){\makebox(0,0){\strut{}$1$}}%
      \csname LTb\endcsname%%
      \put(1016,175){\makebox(0,0){\strut{}$2$}}%
      \csname LTb\endcsname%%
      \put(1283,175){\makebox(0,0){\strut{}$3$}}%
      \csname LTb\endcsname%%
      \put(1550,175){\makebox(0,0){\strut{}$4$}}%
      \csname LTb\endcsname%%
      \put(1817,175){\makebox(0,0){\strut{}$5$}}%
      \csname LTb\endcsname%%
      \put(2084,175){\makebox(0,0){\strut{}$6$}}%
      \csname LTb\endcsname%%
      \put(2351,175){\makebox(0,0){\strut{}$7$}}%
      \csname LTb\endcsname%%
      \put(2618,175){\makebox(0,0){\strut{}$8$}}%
      \csname LTb\endcsname%%
      \put(2885,175){\makebox(0,0){\strut{}$9$}}%
      \csname LTb\endcsname%%
      \put(3152,175){\makebox(0,0){\strut{}$10$}}%
      \csname LTb\endcsname%%
      \put(3419,175){\makebox(0,0){\strut{}$11$}}%
      \csname LTb\endcsname%%
      \put(3686,175){\makebox(0,0){\strut{}$12$}}%
      \csname LTb\endcsname%%
      \put(3953,175){\makebox(0,0){\strut{}$13$}}%
      \csname LTb\endcsname%%
      \put(4220,175){\makebox(0,0){\strut{}$14$}}%
      \csname LTb\endcsname%%
      \put(4487,175){\makebox(0,0){\strut{}$15$}}%
      \csname LTb\endcsname%%
      \put(4754,175){\makebox(0,0){\strut{}$16$}}%
      \csname LTb\endcsname%%
      \put(5021,175){\makebox(0,0){\strut{}$17$}}%
      \csname LTb\endcsname%%
      \put(5288,175){\makebox(0,0){\strut{}$18$}}%
      \csname LTb\endcsname%%
      \put(5555,175){\makebox(0,0){\strut{}$19$}}%
      \csname LTb\endcsname%%
      \put(5822,175){\makebox(0,0){\strut{}$20$}}%
    }%
    \gplgaddtomacro\gplfronttext{%
      \csname LTb\endcsname%%
      \put(161,2237){\rotatebox{-270}{\makebox(0,0){\strut{}Number of files}}}%
      \csname LTb\endcsname%%
      \put(6454,351){\makebox(0,0)[l]{\strut{}$0$}}%
      \csname LTb\endcsname%%
      \put(6454,890){\makebox(0,0)[l]{\strut{}$1$}}%
      \csname LTb\endcsname%%
      \put(6454,1429){\makebox(0,0)[l]{\strut{}$2$}}%
      \csname LTb\endcsname%%
      \put(6454,1968){\makebox(0,0)[l]{\strut{}$3$}}%
      \csname LTb\endcsname%%
      \put(6454,2507){\makebox(0,0)[l]{\strut{}$4$}}%
      \csname LTb\endcsname%%
      \put(6454,3046){\makebox(0,0)[l]{\strut{}$5$}}%
      \csname LTb\endcsname%%
      \put(6454,3585){\makebox(0,0)[l]{\strut{}$6$}}%
      \csname LTb\endcsname%%
      \put(6454,4124){\makebox(0,0)[l]{\strut{}$7$}}%
      \csname LTb\endcsname%%
      \put(6698,2237){\rotatebox{-270}{\makebox(0,0){\strut{}$\log_{10}$(file size)}}}%
    }%
    \gplbacktext
    \put(0,0){\includegraphics[width={360.00bp},height={216.00bp}]{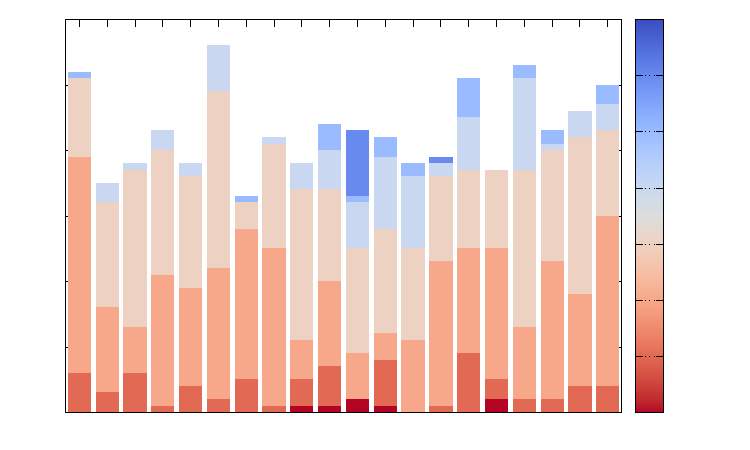}}%
    \gplfronttext
  \end{picture}%
\endgroup

%% file: figures/bar_file_size_by_soi_infamous.tex
% GNUPLOT: LaTeX picture with Postscript
\begingroup
\small
  \makeatletter
  \providecommand\color[2][]{%
    \GenericError{(gnuplot) \space\space\space\@spaces}{%
      Package color not loaded in conjunction with
      terminal option `colourtext'%
    }{See the gnuplot documentation for explanation.%
    }{Either use 'blacktext' in gnuplot or load the package
      color.sty in LaTeX.}%
    \renewcommand\color[2][]{}%
  }%
  \providecommand\includegraphics[2][]{%
    \GenericError{(gnuplot) \space\space\space\@spaces}{%
      Package graphicx or graphics not loaded%
    }{See the gnuplot documentation for explanation.%
    }{The gnuplot epslatex terminal needs graphicx.sty or graphics.sty.}%
    \renewcommand\includegraphics[2][]{}%
  }%
  \providecommand\rotatebox[2]{#2}%
  \@ifundefined{ifGPcolor}{%
    \newif\ifGPcolor
    \GPcolortrue
  }{}%
  \@ifundefined{ifGPblacktext}{%
    \newif\ifGPblacktext
    \GPblacktextfalse
  }{}%
  % define a \g@addto@macro without @ in the name:
  \let\gplgaddtomacro\g@addto@macro
  % define empty templates for all commands taking text:
  \gdef\gplbacktext{}%
  \gdef\gplfronttext{}%
  \makeatother
  \ifGPblacktext
    % no textcolor at all
    \def\colorrgb#1{}%
    \def\colorgray#1{}%
  \else
    % gray or color?
    \ifGPcolor
      \def\colorrgb#1{\color[rgb]{#1}}%
      \def\colorgray#1{\color[gray]{#1}}%
      \expandafter\def\csname LTw\endcsname{\color{white}}%
      \expandafter\def\csname LTb\endcsname{\color{black}}%
      \expandafter\def\csname LTa\endcsname{\color{black}}%
      \expandafter\def\csname LT0\endcsname{\color[rgb]{1,0,0}}%
      \expandafter\def\csname LT1\endcsname{\color[rgb]{0,1,0}}%
      \expandafter\def\csname LT2\endcsname{\color[rgb]{0,0,1}}%
      \expandafter\def\csname LT3\endcsname{\color[rgb]{1,0,1}}%
      \expandafter\def\csname LT4\endcsname{\color[rgb]{0,1,1}}%
      \expandafter\def\csname LT5\endcsname{\color[rgb]{1,1,0}}%
      \expandafter\def\csname LT6\endcsname{\color[rgb]{0,0,0}}%
      \expandafter\def\csname LT7\endcsname{\color[rgb]{1,0.3,0}}%
      \expandafter\def\csname LT8\endcsname{\color[rgb]{0.5,0.5,0.5}}%
    \else
      % gray
      \def\colorrgb#1{\color{black}}%
      \def\colorgray#1{\color[gray]{#1}}%
      \expandafter\def\csname LTw\endcsname{\color{white}}%
      \expandafter\def\csname LTb\endcsname{\color{black}}%
      \expandafter\def\csname LTa\endcsname{\color{black}}%
      \expandafter\def\csname LT0\endcsname{\color{black}}%
      \expandafter\def\csname LT1\endcsname{\color{black}}%
      \expandafter\def\csname LT2\endcsname{\color{black}}%
      \expandafter\def\csname LT3\endcsname{\color{black}}%
      \expandafter\def\csname LT4\endcsname{\color{black}}%
      \expandafter\def\csname LT5\endcsname{\color{black}}%
      \expandafter\def\csname LT6\endcsname{\color{black}}%
      \expandafter\def\csname LT7\endcsname{\color{black}}%
      \expandafter\def\csname LT8\endcsname{\color{black}}%
    \fi
  \fi
    \setlength{\unitlength}{0.0500bp}%
    \ifx\gptboxheight\undefined%
      \newlength{\gptboxheight}%
      \newlength{\gptboxwidth}%
      \newsavebox{\gptboxtext}%
    \fi%
    \setlength{\fboxrule}{0.5pt}%
    \setlength{\fboxsep}{1pt}%
    \definecolor{tbcol}{rgb}{1,1,1}%
\begin{picture}(7200.00,4320.00)%
    \gplgaddtomacro\gplbacktext{%
      \csname LTb\endcsname%%
      \put(518,351){\makebox(0,0)[r]{\strut{}$0$}}%
      \csname LTb\endcsname%%
      \put(518,890){\makebox(0,0)[r]{\strut{}$10$}}%
      \csname LTb\endcsname%%
      \put(518,1429){\makebox(0,0)[r]{\strut{}$20$}}%
      \csname LTb\endcsname%%
      \put(518,1968){\makebox(0,0)[r]{\strut{}$30$}}%
      \csname LTb\endcsname%%
      \put(518,2507){\makebox(0,0)[r]{\strut{}$40$}}%
      \csname LTb\endcsname%%
      \put(518,3046){\makebox(0,0)[r]{\strut{}$50$}}%
      \csname LTb\endcsname%%
      \put(518,3585){\makebox(0,0)[r]{\strut{}$60$}}%
      \csname LTb\endcsname%%
      \put(518,4124){\makebox(0,0)[r]{\strut{}$70$}}%
      \csname LTb\endcsname%%
      \put(749,175){\makebox(0,0){\strut{}$1$}}%
      \csname LTb\endcsname%%
      \put(1016,175){\makebox(0,0){\strut{}$2$}}%
      \csname LTb\endcsname%%
      \put(1283,175){\makebox(0,0){\strut{}$3$}}%
      \csname LTb\endcsname%%
      \put(1550,175){\makebox(0,0){\strut{}$4$}}%
      \csname LTb\endcsname%%
      \put(1817,175){\makebox(0,0){\strut{}$5$}}%
      \csname LTb\endcsname%%
      \put(2084,175){\makebox(0,0){\strut{}$6$}}%
      \csname LTb\endcsname%%
      \put(2351,175){\makebox(0,0){\strut{}$7$}}%
      \csname LTb\endcsname%%
      \put(2618,175){\makebox(0,0){\strut{}$8$}}%
      \csname LTb\endcsname%%
      \put(2885,175){\makebox(0,0){\strut{}$9$}}%
      \csname LTb\endcsname%%
      \put(3152,175){\makebox(0,0){\strut{}$10$}}%
      \csname LTb\endcsname%%
      \put(3419,175){\makebox(0,0){\strut{}$11$}}%
      \csname LTb\endcsname%%
      \put(3686,175){\makebox(0,0){\strut{}$12$}}%
      \csname LTb\endcsname%%
      \put(3953,175){\makebox(0,0){\strut{}$13$}}%
      \csname LTb\endcsname%%
      \put(4220,175){\makebox(0,0){\strut{}$14$}}%
      \csname LTb\endcsname%%
      \put(4487,175){\makebox(0,0){\strut{}$15$}}%
      \csname LTb\endcsname%%
      \put(4754,175){\makebox(0,0){\strut{}$16$}}%
      \csname LTb\endcsname%%
      \put(5021,175){\makebox(0,0){\strut{}$17$}}%
      \csname LTb\endcsname%%
      \put(5288,175){\makebox(0,0){\strut{}$18$}}%
      \csname LTb\endcsname%%
      \put(5555,175){\makebox(0,0){\strut{}$19$}}%
      \csname LTb\endcsname%%
      \put(5822,175){\makebox(0,0){\strut{}$20$}}%
    }%
    \gplgaddtomacro\gplfronttext{%
      \csname LTb\endcsname%%
      \put(161,2237){\rotatebox{-270}{\makebox(0,0){\strut{}Number of files}}}%
      \csname LTb\endcsname%%
      \put(6454,351){\makebox(0,0)[l]{\strut{}$0$}}%
      \csname LTb\endcsname%%
      \put(6454,980){\makebox(0,0)[l]{\strut{}$1$}}%
      \csname LTb\endcsname%%
      \put(6454,1609){\makebox(0,0)[l]{\strut{}$2$}}%
      \csname LTb\endcsname%%
      \put(6454,2237){\makebox(0,0)[l]{\strut{}$3$}}%
      \csname LTb\endcsname%%
      \put(6454,2866){\makebox(0,0)[l]{\strut{}$4$}}%
      \csname LTb\endcsname%%
      \put(6454,3495){\makebox(0,0)[l]{\strut{}$5$}}%
      \csname LTb\endcsname%%
      \put(6454,4124){\makebox(0,0)[l]{\strut{}$6$}}%
      \csname LTb\endcsname%%
      \put(6698,2237){\rotatebox{-270}{\makebox(0,0){\strut{}$\log_{10}$(file size)}}}%
    }%
    \gplbacktext
    \put(0,0){\includegraphics[width={360.00bp},height={216.00bp}]{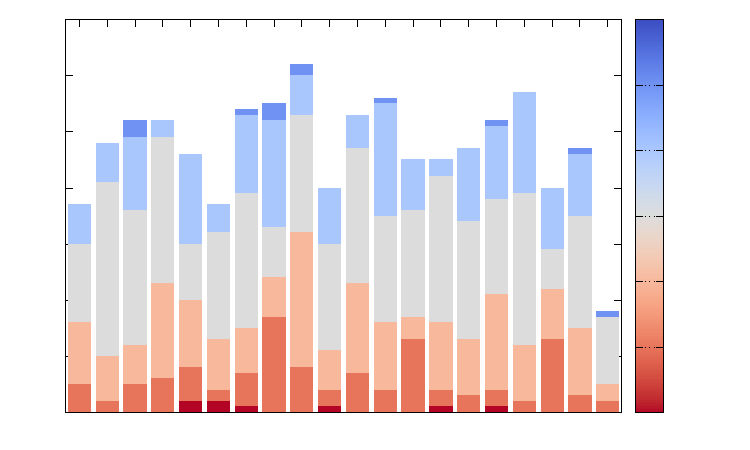}}%
    \gplfronttext
  \end{picture}%
\endgroup

%% file: figures/centroid_boxplot_03_coref_resolved_data_sentences_filtered.tex
% GNUPLOT: LaTeX picture with Postscript
\begingroup
  \makeatletter
  \providecommand\color[2][]{%
    \GenericError{(gnuplot) \space\space\space\@spaces}{%
      Package color not loaded in conjunction with
      terminal option `colourtext'%
    }{See the gnuplot documentation for explanation.%
    }{Either use 'blacktext' in gnuplot or load the package
      color.sty in LaTeX.}%
    \renewcommand\color[2][]{}%
  }%
  \providecommand\includegraphics[2][]{%
    \GenericError{(gnuplot) \space\space\space\@spaces}{%
      Package graphicx or graphics not loaded%
    }{See the gnuplot documentation for explanation.%
    }{The gnuplot epslatex terminal needs graphicx.sty or graphics.sty.}%
    \renewcommand\includegraphics[2][]{}%
  }%
  \providecommand\rotatebox[2]{#2}%
  \@ifundefined{ifGPcolor}{%
    \newif\ifGPcolor
    \GPcolortrue
  }{}%
  \@ifundefined{ifGPblacktext}{%
    \newif\ifGPblacktext
    \GPblacktextfalse
  }{}%
  % define a \g@addto@macro without @ in the name:
  \let\gplgaddtomacro\g@addto@macro
  % define empty templates for all commands taking text:
  \gdef\gplbacktext{}%
  \gdef\gplfronttext{}%
  \makeatother
  \ifGPblacktext
    % no textcolor at all
    \def\colorrgb#1{}%
    \def\colorgray#1{}%
  \else
    % gray or color?
    \ifGPcolor
      \def\colorrgb#1{\color[rgb]{#1}}%
      \def\colorgray#1{\color[gray]{#1}}%
      \expandafter\def\csname LTw\endcsname{\color{white}}%
      \expandafter\def\csname LTb\endcsname{\color{black}}%
      \expandafter\def\csname LTa\endcsname{\color{black}}%
      \expandafter\def\csname LT0\endcsname{\color[rgb]{1,0,0}}%
      \expandafter\def\csname LT1\endcsname{\color[rgb]{0,1,0}}%
      \expandafter\def\csname LT2\endcsname{\color[rgb]{0,0,1}}%
      \expandafter\def\csname LT3\endcsname{\color[rgb]{1,0,1}}%
      \expandafter\def\csname LT4\endcsname{\color[rgb]{0,1,1}}%
      \expandafter\def\csname LT5\endcsname{\color[rgb]{1,1,0}}%
      \expandafter\def\csname LT6\endcsname{\color[rgb]{0,0,0}}%
      \expandafter\def\csname LT7\endcsname{\color[rgb]{1,0.3,0}}%
      \expandafter\def\csname LT8\endcsname{\color[rgb]{0.5,0.5,0.5}}%
    \else
      % gray
      \def\colorrgb#1{\color{black}}%
      \def\colorgray#1{\color[gray]{#1}}%
      \expandafter\def\csname LTw\endcsname{\color{white}}%
      \expandafter\def\csname LTb\endcsname{\color{black}}%
      \expandafter\def\csname LTa\endcsname{\color{black}}%
      \expandafter\def\csname LT0\endcsname{\color{black}}%
      \expandafter\def\csname LT1\endcsname{\color{black}}%
      \expandafter\def\csname LT2\endcsname{\color{black}}%
      \expandafter\def\csname LT3\endcsname{\color{black}}%
      \expandafter\def\csname LT4\endcsname{\color{black}}%
      \expandafter\def\csname LT5\endcsname{\color{black}}%
      \expandafter\def\csname LT6\endcsname{\color{black}}%
      \expandafter\def\csname LT7\endcsname{\color{black}}%
      \expandafter\def\csname LT8\endcsname{\color{black}}%
    \fi
  \fi
    \setlength{\unitlength}{0.0500bp}%
    \ifx\gptboxheight\undefined%
      \newlength{\gptboxheight}%
      \newlength{\gptboxwidth}%
      \newsavebox{\gptboxtext}%
    \fi%
    \setlength{\fboxrule}{0.5pt}%
    \setlength{\fboxsep}{1pt}%
    \definecolor{tbcol}{rgb}{1,1,1}%
\begin{picture}(7200.00,4320.00)%
    \gplgaddtomacro\gplbacktext{%
      \csname LTb\endcsname%%
      \put(620,2322){\makebox(0,0)[r]{\strut{}$-8$}}%
      \csname LTb\endcsname%%
      \put(620,2558){\makebox(0,0)[r]{\strut{}$-6$}}%
      \csname LTb\endcsname%%
      \put(620,2795){\makebox(0,0)[r]{\strut{}$-4$}}%
      \csname LTb\endcsname%%
      \put(620,3031){\makebox(0,0)[r]{\strut{}$-2$}}%
      \csname LTb\endcsname%%
      \put(620,3268){\makebox(0,0)[r]{\strut{}$0$}}%
      \csname LTb\endcsname%%
      \put(620,3504){\makebox(0,0)[r]{\strut{}$2$}}%
      \csname LTb\endcsname%%
      \put(620,3740){\makebox(0,0)[r]{\strut{}$4$}}%
      \csname LTb\endcsname%%
      \put(620,3977){\makebox(0,0)[r]{\strut{}$6$}}%
      \csname LTb\endcsname%%
      \put(620,4213){\makebox(0,0)[r]{\strut{}$8$}}%
    }%
    \gplgaddtomacro\gplfronttext{%
      \csname LTb\endcsname%%
      \put(262,3267){\rotatebox{-270}{\makebox(0,0){\small Centroid}}}%
      \csname LTb\endcsname%%
      \put(2189,4125){\makebox(0,0){\strut{}\textsc{afinn}}}%
    }%
    \gplgaddtomacro\gplbacktext{%
      \csname LTb\endcsname%%
      \put(4066,2322){\makebox(0,0)[r]{\strut{}$-3$}}%
      \csname LTb\endcsname%%
      \put(4066,2637){\makebox(0,0)[r]{\strut{}$-2$}}%
      \csname LTb\endcsname%%
      \put(4066,2952){\makebox(0,0)[r]{\strut{}$-1$}}%
      \csname LTb\endcsname%%
      \put(4066,3268){\makebox(0,0)[r]{\strut{}$0$}}%
      \csname LTb\endcsname%%
      \put(4066,3583){\makebox(0,0)[r]{\strut{}$1$}}%
      \csname LTb\endcsname%%
      \put(4066,3898){\makebox(0,0)[r]{\strut{}$2$}}%
      \csname LTb\endcsname%%
      \put(4066,4213){\makebox(0,0)[r]{\strut{}$3$}}%
    }%
    \gplgaddtomacro\gplfronttext{%
      \csname LTb\endcsname%%
      \put(5636,4125){\makebox(0,0){\strut{}\textsc{minos}}}%
    }%
    \gplgaddtomacro\gplbacktext{%
      \csname LTb\endcsname%%
      \put(620,301){\makebox(0,0)[r]{\strut{}$-0.4$}}%
      \csname LTb\endcsname%%
      \put(620,537){\makebox(0,0)[r]{\strut{}$-0.3$}}%
      \csname LTb\endcsname%%
      \put(620,774){\makebox(0,0)[r]{\strut{}$-0.2$}}%
      \csname LTb\endcsname%%
      \put(620,1010){\makebox(0,0)[r]{\strut{}$-0.1$}}%
      \csname LTb\endcsname%%
      \put(620,1247){\makebox(0,0)[r]{\strut{}$0$}}%
      \csname LTb\endcsname%%
      \put(620,1483){\makebox(0,0)[r]{\strut{}$0.1$}}%
      \csname LTb\endcsname%%
      \put(620,1719){\makebox(0,0)[r]{\strut{}$0.2$}}%
      \csname LTb\endcsname%%
      \put(620,1956){\makebox(0,0)[r]{\strut{}$0.3$}}%
      \csname LTb\endcsname%%
      \put(620,2192){\makebox(0,0)[r]{\strut{}$0.4$}}%
      \csname LTb\endcsname%%
      \put(1208,125){\makebox(0,0){\strut{}awarded}}%
      \csname LTb\endcsname%%
      \put(2189,125){\makebox(0,0){\strut{}forfeited}}%
      \csname LTb\endcsname%%
      \put(3171,125){\makebox(0,0){\strut{}infamous}}%
    }%
    \gplgaddtomacro\gplfronttext{%
      \csname LTb\endcsname%%
      \put(67,1246){\rotatebox{-270}{\makebox(0,0){\small Centroid}}}%
      \csname LTb\endcsname%%
      \put(2189,2104){\makebox(0,0){\strut{}\textsc{vader} (simple)}}%
    }%
    \gplgaddtomacro\gplbacktext{%
      \csname LTb\endcsname%%
      \put(4066,436){\makebox(0,0)[r]{\strut{}$-0.6$}}%
      \csname LTb\endcsname%%
      \put(4066,706){\makebox(0,0)[r]{\strut{}$-0.4$}}%
      \csname LTb\endcsname%%
      \put(4066,976){\makebox(0,0)[r]{\strut{}$-0.2$}}%
      \csname LTb\endcsname%%
      \put(4066,1246){\makebox(0,0)[r]{\strut{}$0$}}%
      \csname LTb\endcsname%%
      \put(4066,1517){\makebox(0,0)[r]{\strut{}$0.2$}}%
      \csname LTb\endcsname%%
      \put(4066,1787){\makebox(0,0)[r]{\strut{}$0.4$}}%
      \csname LTb\endcsname%%
      \put(4066,2057){\makebox(0,0)[r]{\strut{}$0.6$}}%
      \csname LTb\endcsname%%
      \put(4655,125){\makebox(0,0){\strut{}awarded}}%
      \csname LTb\endcsname%%
      \put(5636,125){\makebox(0,0){\strut{}forfeited}}%
      \csname LTb\endcsname%%
      \put(6617,125){\makebox(0,0){\strut{}infamous}}%
    }%
    \gplgaddtomacro\gplfronttext{%
      \csname LTb\endcsname%%
      \put(5636,2104){\makebox(0,0){\strut{}\textsc{vader} (compound)}}%
    }%
    \gplbacktext
    \put(0,0){\includegraphics[width={360.00bp},height={216.00bp}]{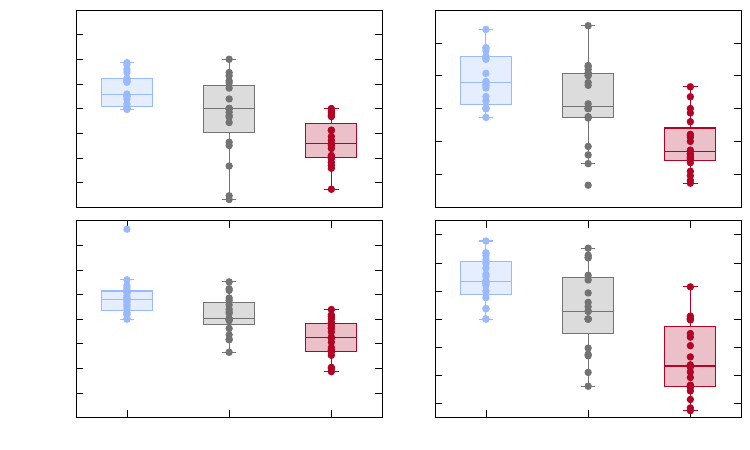}}%
    \gplfronttext
  \end{picture}%
\endgroup

%% file: figures/bhattacharyya_sushantha_kumar/afinn.tex
% GNUPLOT: LaTeX picture with Postscript
\begingroup
  \makeatletter
  \providecommand\color[2][]{%
    \GenericError{(gnuplot) \space\space\space\@spaces}{%
      Package color not loaded in conjunction with
      terminal option `colourtext'%
    }{See the gnuplot documentation for explanation.%
    }{Either use 'blacktext' in gnuplot or load the package
      color.sty in LaTeX.}%
    \renewcommand\color[2][]{}%
  }%
  \providecommand\includegraphics[2][]{%
    \GenericError{(gnuplot) \space\space\space\@spaces}{%
      Package graphicx or graphics not loaded%
    }{See the gnuplot documentation for explanation.%
    }{The gnuplot epslatex terminal needs graphicx.sty or graphics.sty.}%
    \renewcommand\includegraphics[2][]{}%
  }%
  \providecommand\rotatebox[2]{#2}%
  \@ifundefined{ifGPcolor}{%
    \newif\ifGPcolor
    \GPcolortrue
  }{}%
  \@ifundefined{ifGPblacktext}{%
    \newif\ifGPblacktext
    \GPblacktextfalse
  }{}%
  % define a \g@addto@macro without @ in the name:
  \let\gplgaddtomacro\g@addto@macro
  % define empty templates for all commands taking text:
  \gdef\gplbacktext{}%
  \gdef\gplfronttext{}%
  \makeatother
  \ifGPblacktext
    % no textcolor at all
    \def\colorrgb#1{}%
    \def\colorgray#1{}%
  \else
    % gray or color?
    \ifGPcolor
      \def\colorrgb#1{\color[rgb]{#1}}%
      \def\colorgray#1{\color[gray]{#1}}%
      \expandafter\def\csname LTw\endcsname{\color{white}}%
      \expandafter\def\csname LTb\endcsname{\color{black}}%
      \expandafter\def\csname LTa\endcsname{\color{black}}%
      \expandafter\def\csname LT0\endcsname{\color[rgb]{1,0,0}}%
      \expandafter\def\csname LT1\endcsname{\color[rgb]{0,1,0}}%
      \expandafter\def\csname LT2\endcsname{\color[rgb]{0,0,1}}%
      \expandafter\def\csname LT3\endcsname{\color[rgb]{1,0,1}}%
      \expandafter\def\csname LT4\endcsname{\color[rgb]{0,1,1}}%
      \expandafter\def\csname LT5\endcsname{\color[rgb]{1,1,0}}%
      \expandafter\def\csname LT6\endcsname{\color[rgb]{0,0,0}}%
      \expandafter\def\csname LT7\endcsname{\color[rgb]{1,0.3,0}}%
      \expandafter\def\csname LT8\endcsname{\color[rgb]{0.5,0.5,0.5}}%
    \else
      % gray
      \def\colorrgb#1{\color{black}}%
      \def\colorgray#1{\color[gray]{#1}}%
      \expandafter\def\csname LTw\endcsname{\color{white}}%
      \expandafter\def\csname LTb\endcsname{\color{black}}%
      \expandafter\def\csname LTa\endcsname{\color{black}}%
      \expandafter\def\csname LT0\endcsname{\color{black}}%
      \expandafter\def\csname LT1\endcsname{\color{black}}%
      \expandafter\def\csname LT2\endcsname{\color{black}}%
      \expandafter\def\csname LT3\endcsname{\color{black}}%
      \expandafter\def\csname LT4\endcsname{\color{black}}%
      \expandafter\def\csname LT5\endcsname{\color{black}}%
      \expandafter\def\csname LT6\endcsname{\color{black}}%
      \expandafter\def\csname LT7\endcsname{\color{black}}%
      \expandafter\def\csname LT8\endcsname{\color{black}}%
    \fi
  \fi
    \setlength{\unitlength}{0.0500bp}%
    \ifx\gptboxheight\undefined%
      \newlength{\gptboxheight}%
      \newlength{\gptboxwidth}%
      \newsavebox{\gptboxtext}%
    \fi%
    \setlength{\fboxrule}{0.5pt}%
    \setlength{\fboxsep}{1pt}%
    \definecolor{tbcol}{rgb}{1,1,1}%
\begin{picture}(7200.00,4320.00)%
    \gplgaddtomacro\gplbacktext{%
      \csname LTb\endcsname%%
      \put(714,633){\makebox(0,0)[r]{\strut{}$0$}}%
      \csname LTb\endcsname%%
      \put(714,1025){\makebox(0,0)[r]{\strut{}$0.05$}}%
      \csname LTb\endcsname%%
      \put(714,1417){\makebox(0,0)[r]{\strut{}$0.1$}}%
      \csname LTb\endcsname%%
      \put(714,1810){\makebox(0,0)[r]{\strut{}$0.15$}}%
      \csname LTb\endcsname%%
      \put(714,2202){\makebox(0,0)[r]{\strut{}$0.2$}}%
      \csname LTb\endcsname%%
      \put(714,2595){\makebox(0,0)[r]{\strut{}$0.25$}}%
      \csname LTb\endcsname%%
      \put(714,2987){\makebox(0,0)[r]{\strut{}$0.3$}}%
      \csname LTb\endcsname%%
      \put(714,3379){\makebox(0,0)[r]{\strut{}$0.35$}}%
      \csname LTb\endcsname%%
      \put(714,3772){\makebox(0,0)[r]{\strut{}$0.4$}}%
      \csname LTb\endcsname%%
      \put(1286,386){\makebox(0,0){\strut{}$-4.5$}}%
      \csname LTb\endcsname%%
      \put(1813,386){\makebox(0,0){\strut{}$-3.5$}}%
      \csname LTb\endcsname%%
      \put(2340,386){\makebox(0,0){\strut{}$-2.5$}}%
      \csname LTb\endcsname%%
      \put(2866,386){\makebox(0,0){\strut{}$-1.5$}}%
      \csname LTb\endcsname%%
      \put(3393,386){\makebox(0,0){\strut{}$-0.5$}}%
      \csname LTb\endcsname%%
      \put(3920,386){\makebox(0,0){\strut{}$0.5$}}%
      \csname LTb\endcsname%%
      \put(4447,386){\makebox(0,0){\strut{}$1.5$}}%
      \csname LTb\endcsname%%
      \put(4974,386){\makebox(0,0){\strut{}$2.5$}}%
      \csname LTb\endcsname%%
      \put(5501,386){\makebox(0,0){\strut{}$3.5$}}%
    }%
    \gplgaddtomacro\gplfronttext{%
      \csname LTb\endcsname%%
      \put(1286,1578){\makebox(0,0){\strut{}5}}%
      \csname LTb\endcsname%%
      \put(1813,1424){\makebox(0,0){\strut{}4}}%
      \csname LTb\endcsname%%
      \put(3393,1424){\makebox(0,0){\strut{}4}}%
      \csname LTb\endcsname%%
      \put(4447,2040){\makebox(0,0){\strut{}8}}%
      \csname LTb\endcsname%%
      \put(4974,2655){\makebox(0,0){\strut{}12}}%
      \csname LTb\endcsname%%
      \put(5501,3578){\makebox(0,0){\strut{}18}}%
      \csname LTb\endcsname%%
      \put(1469,3622){\makebox(0,0)[l]{\strut{}centroid (+1.036)}}%
      \csname LTb\endcsname%%
      \put(161,2202){\rotatebox{-270.00}{\makebox(0,0){\strut{}Normalised density}}}%
      \csname LTb\endcsname%%
      \put(3393,123){\makebox(0,0){\strut{}Sentiment value}}%
      \csname LTb\endcsname%%
      \put(6460,807){\makebox(0,0)[l]{\strut{}$-4$}}%
      \csname LTb\endcsname%%
      \put(6460,1156){\makebox(0,0)[l]{\strut{}$-3$}}%
      \csname LTb\endcsname%%
      \put(6460,1505){\makebox(0,0)[l]{\strut{}$-2$}}%
      \csname LTb\endcsname%%
      \put(6460,1853){\makebox(0,0)[l]{\strut{}$-1$}}%
      \csname LTb\endcsname%%
      \put(6460,2202){\makebox(0,0)[l]{\strut{}$0$}}%
      \csname LTb\endcsname%%
      \put(6460,2551){\makebox(0,0)[l]{\strut{}$1$}}%
      \csname LTb\endcsname%%
      \put(6460,2900){\makebox(0,0)[l]{\strut{}$2$}}%
      \csname LTb\endcsname%%
      \put(6460,3249){\makebox(0,0)[l]{\strut{}$3$}}%
      \csname LTb\endcsname%%
      \put(6460,3597){\makebox(0,0)[l]{\strut{}$4$}}%
      \csname LTb\endcsname%%
      \put(6705,2202){\rotatebox{-270.00}{\makebox(0,0){\strut{}Sentiment value}}}%
      \csname LTb\endcsname%%
      \put(3393,4036){\makebox(0,0){\strut{}Kumar Bhattacharyya AFINN}}%
    }%
    \gplbacktext
    \put(0,0){\includegraphics[width={360.00bp},height={216.00bp}]{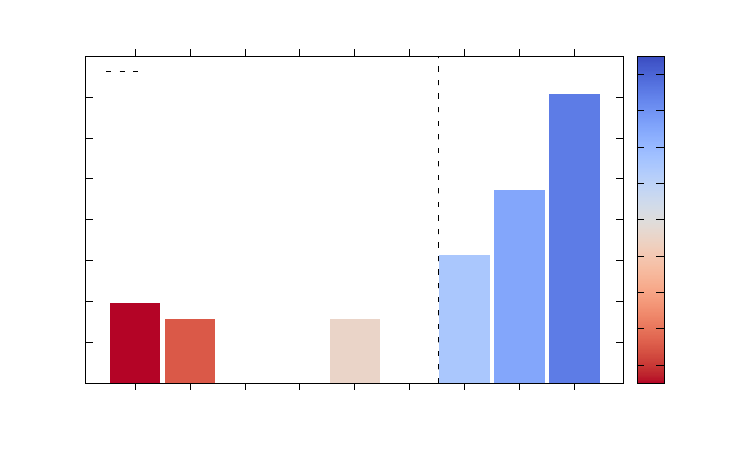}}%
    \gplfronttext
  \end{picture}%
\endgroup

%% file: figures/bhattacharyya_sushantha_kumar/minos.tex
% GNUPLOT: LaTeX picture with Postscript
\begingroup
  \makeatletter
  \providecommand\color[2][]{%
    \GenericError{(gnuplot) \space\space\space\@spaces}{%
      Package color not loaded in conjunction with
      terminal option `colourtext'%
    }{See the gnuplot documentation for explanation.%
    }{Either use 'blacktext' in gnuplot or load the package
      color.sty in LaTeX.}%
    \renewcommand\color[2][]{}%
  }%
  \providecommand\includegraphics[2][]{%
    \GenericError{(gnuplot) \space\space\space\@spaces}{%
      Package graphicx or graphics not loaded%
    }{See the gnuplot documentation for explanation.%
    }{The gnuplot epslatex terminal needs graphicx.sty or graphics.sty.}%
    \renewcommand\includegraphics[2][]{}%
  }%
  \providecommand\rotatebox[2]{#2}%
  \@ifundefined{ifGPcolor}{%
    \newif\ifGPcolor
    \GPcolortrue
  }{}%
  \@ifundefined{ifGPblacktext}{%
    \newif\ifGPblacktext
    \GPblacktextfalse
  }{}%
  % define a \g@addto@macro without @ in the name:
  \let\gplgaddtomacro\g@addto@macro
  % define empty templates for all commands taking text:
  \gdef\gplbacktext{}%
  \gdef\gplfronttext{}%
  \makeatother
  \ifGPblacktext
    % no textcolor at all
    \def\colorrgb#1{}%
    \def\colorgray#1{}%
  \else
    % gray or color?
    \ifGPcolor
      \def\colorrgb#1{\color[rgb]{#1}}%
      \def\colorgray#1{\color[gray]{#1}}%
      \expandafter\def\csname LTw\endcsname{\color{white}}%
      \expandafter\def\csname LTb\endcsname{\color{black}}%
      \expandafter\def\csname LTa\endcsname{\color{black}}%
      \expandafter\def\csname LT0\endcsname{\color[rgb]{1,0,0}}%
      \expandafter\def\csname LT1\endcsname{\color[rgb]{0,1,0}}%
      \expandafter\def\csname LT2\endcsname{\color[rgb]{0,0,1}}%
      \expandafter\def\csname LT3\endcsname{\color[rgb]{1,0,1}}%
      \expandafter\def\csname LT4\endcsname{\color[rgb]{0,1,1}}%
      \expandafter\def\csname LT5\endcsname{\color[rgb]{1,1,0}}%
      \expandafter\def\csname LT6\endcsname{\color[rgb]{0,0,0}}%
      \expandafter\def\csname LT7\endcsname{\color[rgb]{1,0.3,0}}%
      \expandafter\def\csname LT8\endcsname{\color[rgb]{0.5,0.5,0.5}}%
    \else
      % gray
      \def\colorrgb#1{\color{black}}%
      \def\colorgray#1{\color[gray]{#1}}%
      \expandafter\def\csname LTw\endcsname{\color{white}}%
      \expandafter\def\csname LTb\endcsname{\color{black}}%
      \expandafter\def\csname LTa\endcsname{\color{black}}%
      \expandafter\def\csname LT0\endcsname{\color{black}}%
      \expandafter\def\csname LT1\endcsname{\color{black}}%
      \expandafter\def\csname LT2\endcsname{\color{black}}%
      \expandafter\def\csname LT3\endcsname{\color{black}}%
      \expandafter\def\csname LT4\endcsname{\color{black}}%
      \expandafter\def\csname LT5\endcsname{\color{black}}%
      \expandafter\def\csname LT6\endcsname{\color{black}}%
      \expandafter\def\csname LT7\endcsname{\color{black}}%
      \expandafter\def\csname LT8\endcsname{\color{black}}%
    \fi
  \fi
    \setlength{\unitlength}{0.0500bp}%
    \ifx\gptboxheight\undefined%
      \newlength{\gptboxheight}%
      \newlength{\gptboxwidth}%
      \newsavebox{\gptboxtext}%
    \fi%
    \setlength{\fboxrule}{0.5pt}%
    \setlength{\fboxsep}{1pt}%
    \definecolor{tbcol}{rgb}{1,1,1}%
\begin{picture}(7200.00,4320.00)%
    \gplgaddtomacro\gplbacktext{%
      \csname LTb\endcsname%%
      \put(616,633){\makebox(0,0)[r]{\strut{}$0$}}%
      \csname LTb\endcsname%%
      \put(616,1156){\makebox(0,0)[r]{\strut{}$0.1$}}%
      \csname LTb\endcsname%%
      \put(616,1679){\makebox(0,0)[r]{\strut{}$0.2$}}%
      \csname LTb\endcsname%%
      \put(616,2202){\makebox(0,0)[r]{\strut{}$0.3$}}%
      \csname LTb\endcsname%%
      \put(616,2725){\makebox(0,0)[r]{\strut{}$0.4$}}%
      \csname LTb\endcsname%%
      \put(616,3249){\makebox(0,0)[r]{\strut{}$0.5$}}%
      \csname LTb\endcsname%%
      \put(616,3772){\makebox(0,0)[r]{\strut{}$0.6$}}%
      \csname LTb\endcsname%%
      \put(1320,386){\makebox(0,0){\strut{}$-0.5$}}%
      \csname LTb\endcsname%%
      \put(1993,386){\makebox(0,0){\strut{}$0.5$}}%
      \csname LTb\endcsname%%
      \put(2666,386){\makebox(0,0){\strut{}$1.5$}}%
      \csname LTb\endcsname%%
      \put(3339,386){\makebox(0,0){\strut{}$2.5$}}%
      \csname LTb\endcsname%%
      \put(4013,386){\makebox(0,0){\strut{}$3.5$}}%
      \csname LTb\endcsname%%
      \put(4686,386){\makebox(0,0){\strut{}$4.5$}}%
      \csname LTb\endcsname%%
      \put(5359,386){\makebox(0,0){\strut{}$5.5$}}%
    }%
    \gplgaddtomacro\gplfronttext{%
      \csname LTb\endcsname%%
      \put(1320,851){\makebox(0,0){\strut{}1}}%
      \csname LTb\endcsname%%
      \put(2666,3573){\makebox(0,0){\strut{}65}}%
      \csname LTb\endcsname%%
      \put(3339,2212){\makebox(0,0){\strut{}33}}%
      \csname LTb\endcsname%%
      \put(4013,1276){\makebox(0,0){\strut{}11}}%
      \csname LTb\endcsname%%
      \put(4686,1362){\makebox(0,0){\strut{}13}}%
      \csname LTb\endcsname%%
      \put(1372,3622){\makebox(0,0)[l]{\strut{}centroid (+1.849)}}%
      \csname LTb\endcsname%%
      \put(161,2202){\rotatebox{-270.00}{\makebox(0,0){\strut{}Normalised density}}}%
      \csname LTb\endcsname%%
      \put(3339,123){\makebox(0,0){\strut{}Sentiment value}}%
      \csname LTb\endcsname%%
      \put(6457,1061){\makebox(0,0)[l]{\strut{}$-4$}}%
      \csname LTb\endcsname%%
      \put(6457,1631){\makebox(0,0)[l]{\strut{}$-2$}}%
      \csname LTb\endcsname%%
      \put(6457,2202){\makebox(0,0)[l]{\strut{}$0$}}%
      \csname LTb\endcsname%%
      \put(6457,2773){\makebox(0,0)[l]{\strut{}$2$}}%
      \csname LTb\endcsname%%
      \put(6457,3344){\makebox(0,0)[l]{\strut{}$4$}}%
      \csname LTb\endcsname%%
      \put(6701,2202){\rotatebox{-270.00}{\makebox(0,0){\strut{}Sentiment value}}}%
      \csname LTb\endcsname%%
      \put(3339,4036){\makebox(0,0){\strut{}Kumar Bhattacharyya MINOS}}%
    }%
    \gplbacktext
    \put(0,0){\includegraphics[width={360.00bp},height={216.00bp}]{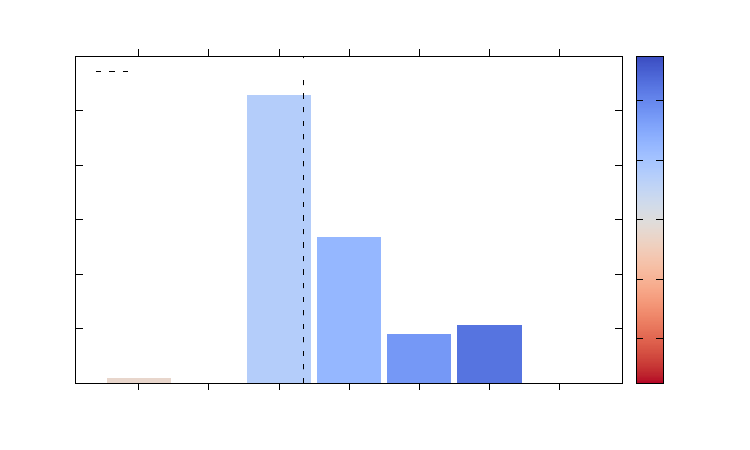}}%
    \gplfronttext
  \end{picture}%
\endgroup

%% file: figures/bhattacharyya_sushantha_kumar/vader_compound.tex
% GNUPLOT: LaTeX picture with Postscript
\begingroup
  \makeatletter
  \providecommand\color[2][]{%
    \GenericError{(gnuplot) \space\space\space\@spaces}{%
      Package color not loaded in conjunction with
      terminal option `colourtext'%
    }{See the gnuplot documentation for explanation.%
    }{Either use 'blacktext' in gnuplot or load the package
      color.sty in LaTeX.}%
    \renewcommand\color[2][]{}%
  }%
  \providecommand\includegraphics[2][]{%
    \GenericError{(gnuplot) \space\space\space\@spaces}{%
      Package graphicx or graphics not loaded%
    }{See the gnuplot documentation for explanation.%
    }{The gnuplot epslatex terminal needs graphicx.sty or graphics.sty.}%
    \renewcommand\includegraphics[2][]{}%
  }%
  \providecommand\rotatebox[2]{#2}%
  \@ifundefined{ifGPcolor}{%
    \newif\ifGPcolor
    \GPcolortrue
  }{}%
  \@ifundefined{ifGPblacktext}{%
    \newif\ifGPblacktext
    \GPblacktextfalse
  }{}%
  % define a \g@addto@macro without @ in the name:
  \let\gplgaddtomacro\g@addto@macro
  % define empty templates for all commands taking text:
  \gdef\gplbacktext{}%
  \gdef\gplfronttext{}%
  \makeatother
  \ifGPblacktext
    % no textcolor at all
    \def\colorrgb#1{}%
    \def\colorgray#1{}%
  \else
    % gray or color?
    \ifGPcolor
      \def\colorrgb#1{\color[rgb]{#1}}%
      \def\colorgray#1{\color[gray]{#1}}%
      \expandafter\def\csname LTw\endcsname{\color{white}}%
      \expandafter\def\csname LTb\endcsname{\color{black}}%
      \expandafter\def\csname LTa\endcsname{\color{black}}%
      \expandafter\def\csname LT0\endcsname{\color[rgb]{1,0,0}}%
      \expandafter\def\csname LT1\endcsname{\color[rgb]{0,1,0}}%
      \expandafter\def\csname LT2\endcsname{\color[rgb]{0,0,1}}%
      \expandafter\def\csname LT3\endcsname{\color[rgb]{1,0,1}}%
      \expandafter\def\csname LT4\endcsname{\color[rgb]{0,1,1}}%
      \expandafter\def\csname LT5\endcsname{\color[rgb]{1,1,0}}%
      \expandafter\def\csname LT6\endcsname{\color[rgb]{0,0,0}}%
      \expandafter\def\csname LT7\endcsname{\color[rgb]{1,0.3,0}}%
      \expandafter\def\csname LT8\endcsname{\color[rgb]{0.5,0.5,0.5}}%
    \else
      % gray
      \def\colorrgb#1{\color{black}}%
      \def\colorgray#1{\color[gray]{#1}}%
      \expandafter\def\csname LTw\endcsname{\color{white}}%
      \expandafter\def\csname LTb\endcsname{\color{black}}%
      \expandafter\def\csname LTa\endcsname{\color{black}}%
      \expandafter\def\csname LT0\endcsname{\color{black}}%
      \expandafter\def\csname LT1\endcsname{\color{black}}%
      \expandafter\def\csname LT2\endcsname{\color{black}}%
      \expandafter\def\csname LT3\endcsname{\color{black}}%
      \expandafter\def\csname LT4\endcsname{\color{black}}%
      \expandafter\def\csname LT5\endcsname{\color{black}}%
      \expandafter\def\csname LT6\endcsname{\color{black}}%
      \expandafter\def\csname LT7\endcsname{\color{black}}%
      \expandafter\def\csname LT8\endcsname{\color{black}}%
    \fi
  \fi
    \setlength{\unitlength}{0.0500bp}%
    \ifx\gptboxheight\undefined%
      \newlength{\gptboxheight}%
      \newlength{\gptboxwidth}%
      \newsavebox{\gptboxtext}%
    \fi%
    \setlength{\fboxrule}{0.5pt}%
    \setlength{\fboxsep}{1pt}%
    \definecolor{tbcol}{rgb}{1,1,1}%
\begin{picture}(7200.00,4320.00)%
    \gplgaddtomacro\gplbacktext{%
      \csname LTb\endcsname%%
      \put(714,633){\makebox(0,0)[r]{\strut{}$0$}}%
      \csname LTb\endcsname%%
      \put(714,1260){\makebox(0,0)[r]{\strut{}$0.05$}}%
      \csname LTb\endcsname%%
      \put(714,1888){\makebox(0,0)[r]{\strut{}$0.1$}}%
      \csname LTb\endcsname%%
      \put(714,2516){\makebox(0,0)[r]{\strut{}$0.15$}}%
      \csname LTb\endcsname%%
      \put(714,3144){\makebox(0,0)[r]{\strut{}$0.2$}}%
      \csname LTb\endcsname%%
      \put(714,3772){\makebox(0,0)[r]{\strut{}$0.25$}}%
      \csname LTb\endcsname%%
      \put(812,386){\makebox(0,0){\strut{}$-0.95$}}%
      \csname LTb\endcsname%%
      \put(1355,386){\makebox(0,0){\strut{}$-0.75$}}%
      \csname LTb\endcsname%%
      \put(1899,386){\makebox(0,0){\strut{}$-0.55$}}%
      \csname LTb\endcsname%%
      \put(2442,386){\makebox(0,0){\strut{}$-0.35$}}%
      \csname LTb\endcsname%%
      \put(2986,386){\makebox(0,0){\strut{}$-0.15$}}%
      \csname LTb\endcsname%%
      \put(3529,386){\makebox(0,0){\strut{}$0.05$}}%
      \csname LTb\endcsname%%
      \put(4073,386){\makebox(0,0){\strut{}$0.25$}}%
      \csname LTb\endcsname%%
      \put(4616,386){\makebox(0,0){\strut{}$0.45$}}%
      \csname LTb\endcsname%%
      \put(5160,386){\makebox(0,0){\strut{}$0.65$}}%
      \csname LTb\endcsname%%
      \put(5703,386){\makebox(0,0){\strut{}$0.85$}}%
    }%
    \gplgaddtomacro\gplfronttext{%
      \csname LTb\endcsname%%
      \put(1219,1429){\makebox(0,0){\strut{}4}}%
      \csname LTb\endcsname%%
      \put(1491,1584){\makebox(0,0){\strut{}5}}%
      \csname LTb\endcsname%%
      \put(2578,1429){\makebox(0,0){\strut{}4}}%
      \csname LTb\endcsname%%
      \put(2850,1429){\makebox(0,0){\strut{}4}}%
      \csname LTb\endcsname%%
      \put(3393,1429){\makebox(0,0){\strut{}4}}%
      \csname LTb\endcsname%%
      \put(3665,1429){\makebox(0,0){\strut{}4}}%
      \csname LTb\endcsname%%
      \put(4209,2979){\makebox(0,0){\strut{}14}}%
      \csname LTb\endcsname%%
      \put(4752,2359){\makebox(0,0){\strut{}10}}%
      \csname LTb\endcsname%%
      \put(5024,3289){\makebox(0,0){\strut{}16}}%
      \csname LTb\endcsname%%
      \put(5295,1584){\makebox(0,0){\strut{}5}}%
      \csname LTb\endcsname%%
      \put(5567,1429){\makebox(0,0){\strut{}4}}%
      \csname LTb\endcsname%%
      \put(5839,1894){\makebox(0,0){\strut{}7}}%
      \csname LTb\endcsname%%
      \put(1469,3622){\makebox(0,0)[l]{\strut{}centroid (+0.241)}}%
      \csname LTb\endcsname%%
      \put(161,2202){\rotatebox{-270.00}{\makebox(0,0){\strut{}Normalised density}}}%
      \csname LTb\endcsname%%
      \put(3393,123){\makebox(0,0){\strut{}Sentiment value}}%
      \csname LTb\endcsname%%
      \put(6460,807){\makebox(0,0)[l]{\strut{}$-0.8$}}%
      \csname LTb\endcsname%%
      \put(6460,1156){\makebox(0,0)[l]{\strut{}$-0.6$}}%
      \csname LTb\endcsname%%
      \put(6460,1505){\makebox(0,0)[l]{\strut{}$-0.4$}}%
      \csname LTb\endcsname%%
      \put(6460,1853){\makebox(0,0)[l]{\strut{}$-0.2$}}%
      \csname LTb\endcsname%%
      \put(6460,2202){\makebox(0,0)[l]{\strut{}$0$}}%
      \csname LTb\endcsname%%
      \put(6460,2551){\makebox(0,0)[l]{\strut{}$0.2$}}%
      \csname LTb\endcsname%%
      \put(6460,2900){\makebox(0,0)[l]{\strut{}$0.4$}}%
      \csname LTb\endcsname%%
      \put(6460,3249){\makebox(0,0)[l]{\strut{}$0.6$}}%
      \csname LTb\endcsname%%
      \put(6460,3597){\makebox(0,0)[l]{\strut{}$0.8$}}%
      \csname LTb\endcsname%%
      \put(6900,2202){\rotatebox{-270.00}{\makebox(0,0){\strut{}Sentiment value}}}%
      \csname LTb\endcsname%%
      \put(3393,4036){\makebox(0,0){\strut{}Kumar Bhattacharyya VADER (compound)}}%
    }%
    \gplbacktext
    \put(0,0){\includegraphics[width={360.00bp},height={216.00bp}]{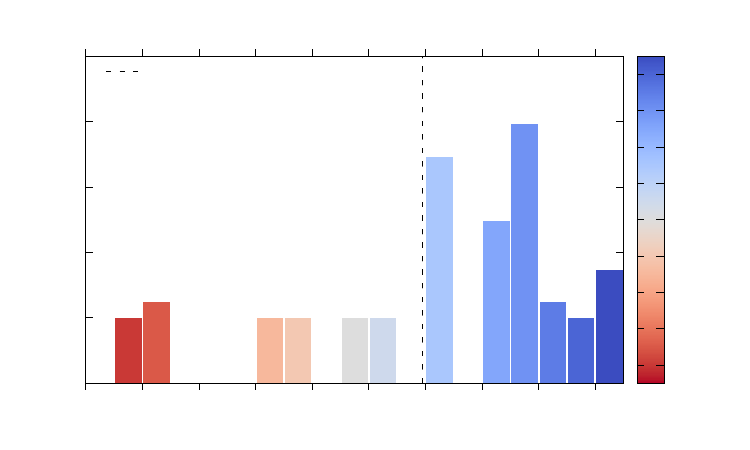}}%
    \gplfronttext
  \end{picture}%
\endgroup

%% file: figures/bhattacharyya_sushantha_kumar/vader_simple.tex
% GNUPLOT: LaTeX picture with Postscript
\begingroup
  \makeatletter
  \providecommand\color[2][]{%
    \GenericError{(gnuplot) \space\space\space\@spaces}{%
      Package color not loaded in conjunction with
      terminal option `colourtext'%
    }{See the gnuplot documentation for explanation.%
    }{Either use 'blacktext' in gnuplot or load the package
      color.sty in LaTeX.}%
    \renewcommand\color[2][]{}%
  }%
  \providecommand\includegraphics[2][]{%
    \GenericError{(gnuplot) \space\space\space\@spaces}{%
      Package graphicx or graphics not loaded%
    }{See the gnuplot documentation for explanation.%
    }{The gnuplot epslatex terminal needs graphicx.sty or graphics.sty.}%
    \renewcommand\includegraphics[2][]{}%
  }%
  \providecommand\rotatebox[2]{#2}%
  \@ifundefined{ifGPcolor}{%
    \newif\ifGPcolor
    \GPcolortrue
  }{}%
  \@ifundefined{ifGPblacktext}{%
    \newif\ifGPblacktext
    \GPblacktextfalse
  }{}%
  % define a \g@addto@macro without @ in the name:
  \let\gplgaddtomacro\g@addto@macro
  % define empty templates for all commands taking text:
  \gdef\gplbacktext{}%
  \gdef\gplfronttext{}%
  \makeatother
  \ifGPblacktext
    % no textcolor at all
    \def\colorrgb#1{}%
    \def\colorgray#1{}%
  \else
    % gray or color?
    \ifGPcolor
      \def\colorrgb#1{\color[rgb]{#1}}%
      \def\colorgray#1{\color[gray]{#1}}%
      \expandafter\def\csname LTw\endcsname{\color{white}}%
      \expandafter\def\csname LTb\endcsname{\color{black}}%
      \expandafter\def\csname LTa\endcsname{\color{black}}%
      \expandafter\def\csname LT0\endcsname{\color[rgb]{1,0,0}}%
      \expandafter\def\csname LT1\endcsname{\color[rgb]{0,1,0}}%
      \expandafter\def\csname LT2\endcsname{\color[rgb]{0,0,1}}%
      \expandafter\def\csname LT3\endcsname{\color[rgb]{1,0,1}}%
      \expandafter\def\csname LT4\endcsname{\color[rgb]{0,1,1}}%
      \expandafter\def\csname LT5\endcsname{\color[rgb]{1,1,0}}%
      \expandafter\def\csname LT6\endcsname{\color[rgb]{0,0,0}}%
      \expandafter\def\csname LT7\endcsname{\color[rgb]{1,0.3,0}}%
      \expandafter\def\csname LT8\endcsname{\color[rgb]{0.5,0.5,0.5}}%
    \else
      % gray
      \def\colorrgb#1{\color{black}}%
      \def\colorgray#1{\color[gray]{#1}}%
      \expandafter\def\csname LTw\endcsname{\color{white}}%
      \expandafter\def\csname LTb\endcsname{\color{black}}%
      \expandafter\def\csname LTa\endcsname{\color{black}}%
      \expandafter\def\csname LT0\endcsname{\color{black}}%
      \expandafter\def\csname LT1\endcsname{\color{black}}%
      \expandafter\def\csname LT2\endcsname{\color{black}}%
      \expandafter\def\csname LT3\endcsname{\color{black}}%
      \expandafter\def\csname LT4\endcsname{\color{black}}%
      \expandafter\def\csname LT5\endcsname{\color{black}}%
      \expandafter\def\csname LT6\endcsname{\color{black}}%
      \expandafter\def\csname LT7\endcsname{\color{black}}%
      \expandafter\def\csname LT8\endcsname{\color{black}}%
    \fi
  \fi
    \setlength{\unitlength}{0.0500bp}%
    \ifx\gptboxheight\undefined%
      \newlength{\gptboxheight}%
      \newlength{\gptboxwidth}%
      \newsavebox{\gptboxtext}%
    \fi%
    \setlength{\fboxrule}{0.5pt}%
    \setlength{\fboxsep}{1pt}%
    \definecolor{tbcol}{rgb}{1,1,1}%
\begin{picture}(7200.00,4320.00)%
    \gplgaddtomacro\gplbacktext{%
      \csname LTb\endcsname%%
      \put(714,633){\makebox(0,0)[r]{\strut{}$0$}}%
      \csname LTb\endcsname%%
      \put(714,981){\makebox(0,0)[r]{\strut{}$0.05$}}%
      \csname LTb\endcsname%%
      \put(714,1330){\makebox(0,0)[r]{\strut{}$0.1$}}%
      \csname LTb\endcsname%%
      \put(714,1679){\makebox(0,0)[r]{\strut{}$0.15$}}%
      \csname LTb\endcsname%%
      \put(714,2028){\makebox(0,0)[r]{\strut{}$0.2$}}%
      \csname LTb\endcsname%%
      \put(714,2377){\makebox(0,0)[r]{\strut{}$0.25$}}%
      \csname LTb\endcsname%%
      \put(714,2725){\makebox(0,0)[r]{\strut{}$0.3$}}%
      \csname LTb\endcsname%%
      \put(714,3074){\makebox(0,0)[r]{\strut{}$0.35$}}%
      \csname LTb\endcsname%%
      \put(714,3423){\makebox(0,0)[r]{\strut{}$0.4$}}%
      \csname LTb\endcsname%%
      \put(714,3772){\makebox(0,0)[r]{\strut{}$0.45$}}%
      \csname LTb\endcsname%%
      \put(812,386){\makebox(0,0){\strut{}$-0.45$}}%
      \csname LTb\endcsname%%
      \put(1328,386){\makebox(0,0){\strut{}$-0.35$}}%
      \csname LTb\endcsname%%
      \put(1844,386){\makebox(0,0){\strut{}$-0.25$}}%
      \csname LTb\endcsname%%
      \put(2361,386){\makebox(0,0){\strut{}$-0.15$}}%
      \csname LTb\endcsname%%
      \put(2877,386){\makebox(0,0){\strut{}$-0.05$}}%
      \csname LTb\endcsname%%
      \put(3393,386){\makebox(0,0){\strut{}$0.05$}}%
      \csname LTb\endcsname%%
      \put(3910,386){\makebox(0,0){\strut{}$0.15$}}%
      \csname LTb\endcsname%%
      \put(4426,386){\makebox(0,0){\strut{}$0.25$}}%
      \csname LTb\endcsname%%
      \put(4942,386){\makebox(0,0){\strut{}$0.35$}}%
      \csname LTb\endcsname%%
      \put(5459,386){\makebox(0,0){\strut{}$0.45$}}%
      \csname LTb\endcsname%%
      \put(5975,386){\makebox(0,0){\strut{}$0.55$}}%
    }%
    \gplgaddtomacro\gplfronttext{%
      \csname LTb\endcsname%%
      \put(1070,895){\makebox(0,0){\strut{}1}}%
      \csname LTb\endcsname%%
      \put(2103,1153){\makebox(0,0){\strut{}4}}%
      \csname LTb\endcsname%%
      \put(2619,895){\makebox(0,0){\strut{}1}}%
      \csname LTb\endcsname%%
      \put(3135,1756){\makebox(0,0){\strut{}11}}%
      \csname LTb\endcsname%%
      \put(3651,3564){\makebox(0,0){\strut{}32}}%
      \csname LTb\endcsname%%
      \put(4168,2962){\makebox(0,0){\strut{}25}}%
      \csname LTb\endcsname%%
      \put(4684,1325){\makebox(0,0){\strut{}6}}%
      \csname LTb\endcsname%%
      \put(5717,895){\makebox(0,0){\strut{}1}}%
      \csname LTb\endcsname%%
      \put(1469,3622){\makebox(0,0)[l]{\strut{}centroid (+0.060)}}%
      \csname LTb\endcsname%%
      \put(161,2202){\rotatebox{-270.00}{\makebox(0,0){\strut{}Normalised density}}}%
      \csname LTb\endcsname%%
      \put(3393,123){\makebox(0,0){\strut{}Sentiment value}}%
      \csname LTb\endcsname%%
      \put(6460,947){\makebox(0,0)[l]{\strut{}$-0.4$}}%
      \csname LTb\endcsname%%
      \put(6460,1574){\makebox(0,0)[l]{\strut{}$-0.2$}}%
      \csname LTb\endcsname%%
      \put(6460,2202){\makebox(0,0)[l]{\strut{}$0$}}%
      \csname LTb\endcsname%%
      \put(6460,2830){\makebox(0,0)[l]{\strut{}$0.2$}}%
      \csname LTb\endcsname%%
      \put(6460,3458){\makebox(0,0)[l]{\strut{}$0.4$}}%
      \csname LTb\endcsname%%
      \put(6900,2202){\rotatebox{-270.00}{\makebox(0,0){\strut{}Sentiment value}}}%
      \csname LTb\endcsname%%
      \put(3393,4036){\makebox(0,0){\strut{}Kumar Bhattacharyya VADER (simple)}}%
    }%
    \gplbacktext
    \put(0,0){\includegraphics[width={360.00bp},height={216.00bp}]{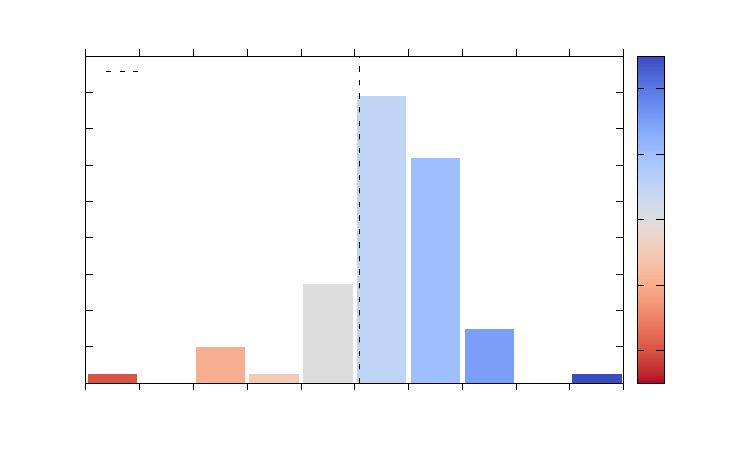}}%
    \gplfronttext
  \end{picture}%
\endgroup

%% file: figures/charles_sobhraj/afinn.tex
% GNUPLOT: LaTeX picture with Postscript
\begingroup
  \makeatletter
  \providecommand\color[2][]{%
    \GenericError{(gnuplot) \space\space\space\@spaces}{%
      Package color not loaded in conjunction with
      terminal option `colourtext'%
    }{See the gnuplot documentation for explanation.%
    }{Either use 'blacktext' in gnuplot or load the package
      color.sty in LaTeX.}%
    \renewcommand\color[2][]{}%
  }%
  \providecommand\includegraphics[2][]{%
    \GenericError{(gnuplot) \space\space\space\@spaces}{%
      Package graphicx or graphics not loaded%
    }{See the gnuplot documentation for explanation.%
    }{The gnuplot epslatex terminal needs graphicx.sty or graphics.sty.}%
    \renewcommand\includegraphics[2][]{}%
  }%
  \providecommand\rotatebox[2]{#2}%
  \@ifundefined{ifGPcolor}{%
    \newif\ifGPcolor
    \GPcolortrue
  }{}%
  \@ifundefined{ifGPblacktext}{%
    \newif\ifGPblacktext
    \GPblacktextfalse
  }{}%
  % define a \g@addto@macro without @ in the name:
  \let\gplgaddtomacro\g@addto@macro
  % define empty templates for all commands taking text:
  \gdef\gplbacktext{}%
  \gdef\gplfronttext{}%
  \makeatother
  \ifGPblacktext
    % no textcolor at all
    \def\colorrgb#1{}%
    \def\colorgray#1{}%
  \else
    % gray or color?
    \ifGPcolor
      \def\colorrgb#1{\color[rgb]{#1}}%
      \def\colorgray#1{\color[gray]{#1}}%
      \expandafter\def\csname LTw\endcsname{\color{white}}%
      \expandafter\def\csname LTb\endcsname{\color{black}}%
      \expandafter\def\csname LTa\endcsname{\color{black}}%
      \expandafter\def\csname LT0\endcsname{\color[rgb]{1,0,0}}%
      \expandafter\def\csname LT1\endcsname{\color[rgb]{0,1,0}}%
      \expandafter\def\csname LT2\endcsname{\color[rgb]{0,0,1}}%
      \expandafter\def\csname LT3\endcsname{\color[rgb]{1,0,1}}%
      \expandafter\def\csname LT4\endcsname{\color[rgb]{0,1,1}}%
      \expandafter\def\csname LT5\endcsname{\color[rgb]{1,1,0}}%
      \expandafter\def\csname LT6\endcsname{\color[rgb]{0,0,0}}%
      \expandafter\def\csname LT7\endcsname{\color[rgb]{1,0.3,0}}%
      \expandafter\def\csname LT8\endcsname{\color[rgb]{0.5,0.5,0.5}}%
    \else
      % gray
      \def\colorrgb#1{\color{black}}%
      \def\colorgray#1{\color[gray]{#1}}%
      \expandafter\def\csname LTw\endcsname{\color{white}}%
      \expandafter\def\csname LTb\endcsname{\color{black}}%
      \expandafter\def\csname LTa\endcsname{\color{black}}%
      \expandafter\def\csname LT0\endcsname{\color{black}}%
      \expandafter\def\csname LT1\endcsname{\color{black}}%
      \expandafter\def\csname LT2\endcsname{\color{black}}%
      \expandafter\def\csname LT3\endcsname{\color{black}}%
      \expandafter\def\csname LT4\endcsname{\color{black}}%
      \expandafter\def\csname LT5\endcsname{\color{black}}%
      \expandafter\def\csname LT6\endcsname{\color{black}}%
      \expandafter\def\csname LT7\endcsname{\color{black}}%
      \expandafter\def\csname LT8\endcsname{\color{black}}%
    \fi
  \fi
    \setlength{\unitlength}{0.0500bp}%
    \ifx\gptboxheight\undefined%
      \newlength{\gptboxheight}%
      \newlength{\gptboxwidth}%
      \newsavebox{\gptboxtext}%
    \fi%
    \setlength{\fboxrule}{0.5pt}%
    \setlength{\fboxsep}{1pt}%
    \definecolor{tbcol}{rgb}{1,1,1}%
\begin{picture}(7200.00,4320.00)%
    \gplgaddtomacro\gplbacktext{%
      \csname LTb\endcsname%%
      \put(714,633){\makebox(0,0)[r]{\strut{}$0$}}%
      \csname LTb\endcsname%%
      \put(714,1417){\makebox(0,0)[r]{\strut{}$0.05$}}%
      \csname LTb\endcsname%%
      \put(714,2202){\makebox(0,0)[r]{\strut{}$0.1$}}%
      \csname LTb\endcsname%%
      \put(714,2987){\makebox(0,0)[r]{\strut{}$0.15$}}%
      \csname LTb\endcsname%%
      \put(714,3772){\makebox(0,0)[r]{\strut{}$0.2$}}%
      \csname LTb\endcsname%%
      \put(958,386){\makebox(0,0){\strut{}$-22.5$}}%
      \csname LTb\endcsname%%
      \put(1445,386){\makebox(0,0){\strut{}$-19.5$}}%
      \csname LTb\endcsname%%
      \put(1932,386){\makebox(0,0){\strut{}$-16.5$}}%
      \csname LTb\endcsname%%
      \put(2419,386){\makebox(0,0){\strut{}$-13.5$}}%
      \csname LTb\endcsname%%
      \put(2906,386){\makebox(0,0){\strut{}$-10.5$}}%
      \csname LTb\endcsname%%
      \put(3393,386){\makebox(0,0){\strut{}$-7.5$}}%
      \csname LTb\endcsname%%
      \put(3880,386){\makebox(0,0){\strut{}$-4.5$}}%
      \csname LTb\endcsname%%
      \put(4367,386){\makebox(0,0){\strut{}$-1.5$}}%
      \csname LTb\endcsname%%
      \put(4855,386){\makebox(0,0){\strut{}$1.5$}}%
      \csname LTb\endcsname%%
      \put(5342,386){\makebox(0,0){\strut{}$4.5$}}%
      \csname LTb\endcsname%%
      \put(5829,386){\makebox(0,0){\strut{}$7.5$}}%
    }%
    \gplgaddtomacro\gplfronttext{%
      \csname LTb\endcsname%%
      \put(958,850){\makebox(0,0){\strut{}2}}%
      \csname LTb\endcsname%%
      \put(1120,829){\makebox(0,0){\strut{}1}}%
      \csname LTb\endcsname%%
      \put(1770,850){\makebox(0,0){\strut{}2}}%
      \csname LTb\endcsname%%
      \put(2094,932){\makebox(0,0){\strut{}6}}%
      \csname LTb\endcsname%%
      \put(2419,911){\makebox(0,0){\strut{}5}}%
      \csname LTb\endcsname%%
      \put(2581,952){\makebox(0,0){\strut{}7}}%
      \csname LTb\endcsname%%
      \put(2744,952){\makebox(0,0){\strut{}7}}%
      \csname LTb\endcsname%%
      \put(2906,952){\makebox(0,0){\strut{}7}}%
      \csname LTb\endcsname%%
      \put(3069,993){\makebox(0,0){\strut{}9}}%
      \csname LTb\endcsname%%
      \put(3231,932){\makebox(0,0){\strut{}6}}%
      \csname LTb\endcsname%%
      \put(3393,1404){\makebox(0,0){\strut{}29}}%
      \csname LTb\endcsname%%
      \put(3556,1281){\makebox(0,0){\strut{}23}}%
      \csname LTb\endcsname%%
      \put(3718,1774){\makebox(0,0){\strut{}47}}%
      \csname LTb\endcsname%%
      \put(3880,1938){\makebox(0,0){\strut{}55}}%
      \csname LTb\endcsname%%
      \put(4043,2699){\makebox(0,0){\strut{}92}}%
      \csname LTb\endcsname%%
      \put(4205,3007){\makebox(0,0){\strut{}107}}%
      \csname LTb\endcsname%%
      \put(4367,3459){\makebox(0,0){\strut{}129}}%
      \csname LTb\endcsname%%
      \put(4530,2185){\makebox(0,0){\strut{}67}}%
      \csname LTb\endcsname%%
      \put(4855,1795){\makebox(0,0){\strut{}48}}%
      \csname LTb\endcsname%%
      \put(5017,2144){\makebox(0,0){\strut{}65}}%
      \csname LTb\endcsname%%
      \put(5179,1363){\makebox(0,0){\strut{}27}}%
      \csname LTb\endcsname%%
      \put(5342,993){\makebox(0,0){\strut{}9}}%
      \csname LTb\endcsname%%
      \put(5504,870){\makebox(0,0){\strut{}3}}%
      \csname LTb\endcsname%%
      \put(5666,993){\makebox(0,0){\strut{}9}}%
      \csname LTb\endcsname%%
      \put(5829,850){\makebox(0,0){\strut{}2}}%
      \csname LTb\endcsname%%
      \put(1469,3622){\makebox(0,0)[l]{\strut{}centroid (-2.862)}}%
      \csname LTb\endcsname%%
      \put(161,2202){\rotatebox{-270.00}{\makebox(0,0){\strut{}Normalised density}}}%
      \csname LTb\endcsname%%
      \put(3393,123){\makebox(0,0){\strut{}Sentiment value}}%
      \csname LTb\endcsname%%
      \put(6460,807){\makebox(0,0)[l]{\strut{}$-20$}}%
      \csname LTb\endcsname%%
      \put(6460,1156){\makebox(0,0)[l]{\strut{}$-15$}}%
      \csname LTb\endcsname%%
      \put(6460,1505){\makebox(0,0)[l]{\strut{}$-10$}}%
      \csname LTb\endcsname%%
      \put(6460,1853){\makebox(0,0)[l]{\strut{}$-5$}}%
      \csname LTb\endcsname%%
      \put(6460,2202){\makebox(0,0)[l]{\strut{}$0$}}%
      \csname LTb\endcsname%%
      \put(6460,2551){\makebox(0,0)[l]{\strut{}$5$}}%
      \csname LTb\endcsname%%
      \put(6460,2900){\makebox(0,0)[l]{\strut{}$10$}}%
      \csname LTb\endcsname%%
      \put(6460,3249){\makebox(0,0)[l]{\strut{}$15$}}%
      \csname LTb\endcsname%%
      \put(6460,3597){\makebox(0,0)[l]{\strut{}$20$}}%
      \csname LTb\endcsname%%
      \put(6803,2202){\rotatebox{-270.00}{\makebox(0,0){\strut{}Sentiment value}}}%
      \csname LTb\endcsname%%
      \put(3393,4036){\makebox(0,0){\strut{}Charles Sobhraj AFINN}}%
    }%
    \gplbacktext
    \put(0,0){\includegraphics[width={360.00bp},height={216.00bp}]{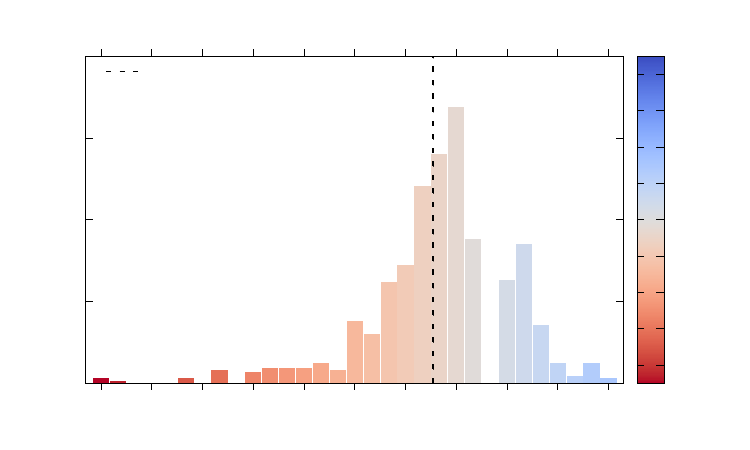}}%
    \gplfronttext
  \end{picture}%
\endgroup

%% file: figures/charles_sobhraj/minos.tex
% GNUPLOT: LaTeX picture with Postscript
\begingroup
  \makeatletter
  \providecommand\color[2][]{%
    \GenericError{(gnuplot) \space\space\space\@spaces}{%
      Package color not loaded in conjunction with
      terminal option `colourtext'%
    }{See the gnuplot documentation for explanation.%
    }{Either use 'blacktext' in gnuplot or load the package
      color.sty in LaTeX.}%
    \renewcommand\color[2][]{}%
  }%
  \providecommand\includegraphics[2][]{%
    \GenericError{(gnuplot) \space\space\space\@spaces}{%
      Package graphicx or graphics not loaded%
    }{See the gnuplot documentation for explanation.%
    }{The gnuplot epslatex terminal needs graphicx.sty or graphics.sty.}%
    \renewcommand\includegraphics[2][]{}%
  }%
  \providecommand\rotatebox[2]{#2}%
  \@ifundefined{ifGPcolor}{%
    \newif\ifGPcolor
    \GPcolortrue
  }{}%
  \@ifundefined{ifGPblacktext}{%
    \newif\ifGPblacktext
    \GPblacktextfalse
  }{}%
  % define a \g@addto@macro without @ in the name:
  \let\gplgaddtomacro\g@addto@macro
  % define empty templates for all commands taking text:
  \gdef\gplbacktext{}%
  \gdef\gplfronttext{}%
  \makeatother
  \ifGPblacktext
    % no textcolor at all
    \def\colorrgb#1{}%
    \def\colorgray#1{}%
  \else
    % gray or color?
    \ifGPcolor
      \def\colorrgb#1{\color[rgb]{#1}}%
      \def\colorgray#1{\color[gray]{#1}}%
      \expandafter\def\csname LTw\endcsname{\color{white}}%
      \expandafter\def\csname LTb\endcsname{\color{black}}%
      \expandafter\def\csname LTa\endcsname{\color{black}}%
      \expandafter\def\csname LT0\endcsname{\color[rgb]{1,0,0}}%
      \expandafter\def\csname LT1\endcsname{\color[rgb]{0,1,0}}%
      \expandafter\def\csname LT2\endcsname{\color[rgb]{0,0,1}}%
      \expandafter\def\csname LT3\endcsname{\color[rgb]{1,0,1}}%
      \expandafter\def\csname LT4\endcsname{\color[rgb]{0,1,1}}%
      \expandafter\def\csname LT5\endcsname{\color[rgb]{1,1,0}}%
      \expandafter\def\csname LT6\endcsname{\color[rgb]{0,0,0}}%
      \expandafter\def\csname LT7\endcsname{\color[rgb]{1,0.3,0}}%
      \expandafter\def\csname LT8\endcsname{\color[rgb]{0.5,0.5,0.5}}%
    \else
      % gray
      \def\colorrgb#1{\color{black}}%
      \def\colorgray#1{\color[gray]{#1}}%
      \expandafter\def\csname LTw\endcsname{\color{white}}%
      \expandafter\def\csname LTb\endcsname{\color{black}}%
      \expandafter\def\csname LTa\endcsname{\color{black}}%
      \expandafter\def\csname LT0\endcsname{\color{black}}%
      \expandafter\def\csname LT1\endcsname{\color{black}}%
      \expandafter\def\csname LT2\endcsname{\color{black}}%
      \expandafter\def\csname LT3\endcsname{\color{black}}%
      \expandafter\def\csname LT4\endcsname{\color{black}}%
      \expandafter\def\csname LT5\endcsname{\color{black}}%
      \expandafter\def\csname LT6\endcsname{\color{black}}%
      \expandafter\def\csname LT7\endcsname{\color{black}}%
      \expandafter\def\csname LT8\endcsname{\color{black}}%
    \fi
  \fi
    \setlength{\unitlength}{0.0500bp}%
    \ifx\gptboxheight\undefined%
      \newlength{\gptboxheight}%
      \newlength{\gptboxwidth}%
      \newsavebox{\gptboxtext}%
    \fi%
    \setlength{\fboxrule}{0.5pt}%
    \setlength{\fboxsep}{1pt}%
    \definecolor{tbcol}{rgb}{1,1,1}%
\begin{picture}(7200.00,4320.00)%
    \gplgaddtomacro\gplbacktext{%
      \csname LTb\endcsname%%
      \put(616,633){\makebox(0,0)[r]{\strut{}$0$}}%
      \csname LTb\endcsname%%
      \put(616,1260){\makebox(0,0)[r]{\strut{}$0.1$}}%
      \csname LTb\endcsname%%
      \put(616,1888){\makebox(0,0)[r]{\strut{}$0.2$}}%
      \csname LTb\endcsname%%
      \put(616,2516){\makebox(0,0)[r]{\strut{}$0.3$}}%
      \csname LTb\endcsname%%
      \put(616,3144){\makebox(0,0)[r]{\strut{}$0.4$}}%
      \csname LTb\endcsname%%
      \put(616,3772){\makebox(0,0)[r]{\strut{}$0.5$}}%
      \csname LTb\endcsname%%
      \put(1056,386){\makebox(0,0){\strut{}$-9.5$}}%
      \csname LTb\endcsname%%
      \put(1437,386){\makebox(0,0){\strut{}$-8.5$}}%
      \csname LTb\endcsname%%
      \put(1817,386){\makebox(0,0){\strut{}$-7.5$}}%
      \csname LTb\endcsname%%
      \put(2198,386){\makebox(0,0){\strut{}$-6.5$}}%
      \csname LTb\endcsname%%
      \put(2578,386){\makebox(0,0){\strut{}$-5.5$}}%
      \csname LTb\endcsname%%
      \put(2959,386){\makebox(0,0){\strut{}$-4.5$}}%
      \csname LTb\endcsname%%
      \put(3339,386){\makebox(0,0){\strut{}$-3.5$}}%
      \csname LTb\endcsname%%
      \put(3720,386){\makebox(0,0){\strut{}$-2.5$}}%
      \csname LTb\endcsname%%
      \put(4100,386){\makebox(0,0){\strut{}$-1.5$}}%
      \csname LTb\endcsname%%
      \put(4481,386){\makebox(0,0){\strut{}$-0.5$}}%
      \csname LTb\endcsname%%
      \put(4862,386){\makebox(0,0){\strut{}$0.5$}}%
      \csname LTb\endcsname%%
      \put(5242,386){\makebox(0,0){\strut{}$1.5$}}%
      \csname LTb\endcsname%%
      \put(5623,386){\makebox(0,0){\strut{}$2.5$}}%
    }%
    \gplgaddtomacro\gplfronttext{%
      \csname LTb\endcsname%%
      \put(1056,817){\makebox(0,0){\strut{}1}}%
      \csname LTb\endcsname%%
      \put(1817,879){\makebox(0,0){\strut{}8}}%
      \csname LTb\endcsname%%
      \put(2198,852){\makebox(0,0){\strut{}5}}%
      \csname LTb\endcsname%%
      \put(2578,879){\makebox(0,0){\strut{}8}}%
      \csname LTb\endcsname%%
      \put(2959,967){\makebox(0,0){\strut{}18}}%
      \csname LTb\endcsname%%
      \put(3339,1248){\makebox(0,0){\strut{}50}}%
      \csname LTb\endcsname%%
      \put(3720,1520){\makebox(0,0){\strut{}81}}%
      \csname LTb\endcsname%%
      \put(4100,2047){\makebox(0,0){\strut{}141}}%
      \csname LTb\endcsname%%
      \put(4481,3601){\makebox(0,0){\strut{}318}}%
      \csname LTb\endcsname%%
      \put(5242,1406){\makebox(0,0){\strut{}68}}%
      \csname LTb\endcsname%%
      \put(5623,958){\makebox(0,0){\strut{}17}}%
      \csname LTb\endcsname%%
      \put(1372,3622){\makebox(0,0)[l]{\strut{}centroid (-1.649)}}%
      \csname LTb\endcsname%%
      \put(161,2202){\rotatebox{-270.00}{\makebox(0,0){\strut{}Normalised density}}}%
      \csname LTb\endcsname%%
      \put(3339,123){\makebox(0,0){\strut{}Sentiment value}}%
      \csname LTb\endcsname%%
      \put(6457,880){\makebox(0,0)[l]{\strut{}$-8$}}%
      \csname LTb\endcsname%%
      \put(6457,1211){\makebox(0,0)[l]{\strut{}$-6$}}%
      \csname LTb\endcsname%%
      \put(6457,1541){\makebox(0,0)[l]{\strut{}$-4$}}%
      \csname LTb\endcsname%%
      \put(6457,1872){\makebox(0,0)[l]{\strut{}$-2$}}%
      \csname LTb\endcsname%%
      \put(6457,2202){\makebox(0,0)[l]{\strut{}$0$}}%
      \csname LTb\endcsname%%
      \put(6457,2533){\makebox(0,0)[l]{\strut{}$2$}}%
      \csname LTb\endcsname%%
      \put(6457,2863){\makebox(0,0)[l]{\strut{}$4$}}%
      \csname LTb\endcsname%%
      \put(6457,3193){\makebox(0,0)[l]{\strut{}$6$}}%
      \csname LTb\endcsname%%
      \put(6457,3524){\makebox(0,0)[l]{\strut{}$8$}}%
      \csname LTb\endcsname%%
      \put(6701,2202){\rotatebox{-270.00}{\makebox(0,0){\strut{}Sentiment value}}}%
      \csname LTb\endcsname%%
      \put(3339,4036){\makebox(0,0){\strut{}Charles Sobhraj MINOS}}%
    }%
    \gplbacktext
    \put(0,0){\includegraphics[width={360.00bp},height={216.00bp}]{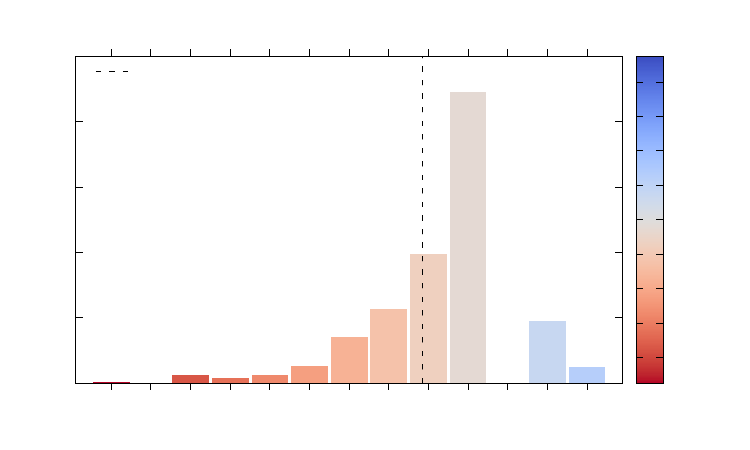}}%
    \gplfronttext
  \end{picture}%
\endgroup

%% file: figures/charles_sobhraj/vader_compound.tex
% GNUPLOT: LaTeX picture with Postscript
\begingroup
  \makeatletter
  \providecommand\color[2][]{%
    \GenericError{(gnuplot) \space\space\space\@spaces}{%
      Package color not loaded in conjunction with
      terminal option `colourtext'%
    }{See the gnuplot documentation for explanation.%
    }{Either use 'blacktext' in gnuplot or load the package
      color.sty in LaTeX.}%
    \renewcommand\color[2][]{}%
  }%
  \providecommand\includegraphics[2][]{%
    \GenericError{(gnuplot) \space\space\space\@spaces}{%
      Package graphicx or graphics not loaded%
    }{See the gnuplot documentation for explanation.%
    }{The gnuplot epslatex terminal needs graphicx.sty or graphics.sty.}%
    \renewcommand\includegraphics[2][]{}%
  }%
  \providecommand\rotatebox[2]{#2}%
  \@ifundefined{ifGPcolor}{%
    \newif\ifGPcolor
    \GPcolortrue
  }{}%
  \@ifundefined{ifGPblacktext}{%
    \newif\ifGPblacktext
    \GPblacktextfalse
  }{}%
  % define a \g@addto@macro without @ in the name:
  \let\gplgaddtomacro\g@addto@macro
  % define empty templates for all commands taking text:
  \gdef\gplbacktext{}%
  \gdef\gplfronttext{}%
  \makeatother
  \ifGPblacktext
    % no textcolor at all
    \def\colorrgb#1{}%
    \def\colorgray#1{}%
  \else
    % gray or color?
    \ifGPcolor
      \def\colorrgb#1{\color[rgb]{#1}}%
      \def\colorgray#1{\color[gray]{#1}}%
      \expandafter\def\csname LTw\endcsname{\color{white}}%
      \expandafter\def\csname LTb\endcsname{\color{black}}%
      \expandafter\def\csname LTa\endcsname{\color{black}}%
      \expandafter\def\csname LT0\endcsname{\color[rgb]{1,0,0}}%
      \expandafter\def\csname LT1\endcsname{\color[rgb]{0,1,0}}%
      \expandafter\def\csname LT2\endcsname{\color[rgb]{0,0,1}}%
      \expandafter\def\csname LT3\endcsname{\color[rgb]{1,0,1}}%
      \expandafter\def\csname LT4\endcsname{\color[rgb]{0,1,1}}%
      \expandafter\def\csname LT5\endcsname{\color[rgb]{1,1,0}}%
      \expandafter\def\csname LT6\endcsname{\color[rgb]{0,0,0}}%
      \expandafter\def\csname LT7\endcsname{\color[rgb]{1,0.3,0}}%
      \expandafter\def\csname LT8\endcsname{\color[rgb]{0.5,0.5,0.5}}%
    \else
      % gray
      \def\colorrgb#1{\color{black}}%
      \def\colorgray#1{\color[gray]{#1}}%
      \expandafter\def\csname LTw\endcsname{\color{white}}%
      \expandafter\def\csname LTb\endcsname{\color{black}}%
      \expandafter\def\csname LTa\endcsname{\color{black}}%
      \expandafter\def\csname LT0\endcsname{\color{black}}%
      \expandafter\def\csname LT1\endcsname{\color{black}}%
      \expandafter\def\csname LT2\endcsname{\color{black}}%
      \expandafter\def\csname LT3\endcsname{\color{black}}%
      \expandafter\def\csname LT4\endcsname{\color{black}}%
      \expandafter\def\csname LT5\endcsname{\color{black}}%
      \expandafter\def\csname LT6\endcsname{\color{black}}%
      \expandafter\def\csname LT7\endcsname{\color{black}}%
      \expandafter\def\csname LT8\endcsname{\color{black}}%
    \fi
  \fi
    \setlength{\unitlength}{0.0500bp}%
    \ifx\gptboxheight\undefined%
      \newlength{\gptboxheight}%
      \newlength{\gptboxwidth}%
      \newsavebox{\gptboxtext}%
    \fi%
    \setlength{\fboxrule}{0.5pt}%
    \setlength{\fboxsep}{1pt}%
    \definecolor{tbcol}{rgb}{1,1,1}%
\begin{picture}(7200.00,4320.00)%
    \gplgaddtomacro\gplbacktext{%
      \csname LTb\endcsname%%
      \put(714,633){\makebox(0,0)[r]{\strut{}$0$}}%
      \csname LTb\endcsname%%
      \put(714,981){\makebox(0,0)[r]{\strut{}$0.02$}}%
      \csname LTb\endcsname%%
      \put(714,1330){\makebox(0,0)[r]{\strut{}$0.04$}}%
      \csname LTb\endcsname%%
      \put(714,1679){\makebox(0,0)[r]{\strut{}$0.06$}}%
      \csname LTb\endcsname%%
      \put(714,2028){\makebox(0,0)[r]{\strut{}$0.08$}}%
      \csname LTb\endcsname%%
      \put(714,2377){\makebox(0,0)[r]{\strut{}$0.1$}}%
      \csname LTb\endcsname%%
      \put(714,2725){\makebox(0,0)[r]{\strut{}$0.12$}}%
      \csname LTb\endcsname%%
      \put(714,3074){\makebox(0,0)[r]{\strut{}$0.14$}}%
      \csname LTb\endcsname%%
      \put(714,3423){\makebox(0,0)[r]{\strut{}$0.16$}}%
      \csname LTb\endcsname%%
      \put(714,3772){\makebox(0,0)[r]{\strut{}$0.18$}}%
      \csname LTb\endcsname%%
      \put(812,386){\makebox(0,0){\strut{}$-1$}}%
      \csname LTb\endcsname%%
      \put(1355,386){\makebox(0,0){\strut{}$-0.8$}}%
      \csname LTb\endcsname%%
      \put(1899,386){\makebox(0,0){\strut{}$-0.6$}}%
      \csname LTb\endcsname%%
      \put(2442,386){\makebox(0,0){\strut{}$-0.4$}}%
      \csname LTb\endcsname%%
      \put(2986,386){\makebox(0,0){\strut{}$-0.2$}}%
      \csname LTb\endcsname%%
      \put(3529,386){\makebox(0,0){\strut{}$0$}}%
      \csname LTb\endcsname%%
      \put(4073,386){\makebox(0,0){\strut{}$0.2$}}%
      \csname LTb\endcsname%%
      \put(4616,386){\makebox(0,0){\strut{}$0.4$}}%
      \csname LTb\endcsname%%
      \put(5160,386){\makebox(0,0){\strut{}$0.6$}}%
      \csname LTb\endcsname%%
      \put(5703,386){\makebox(0,0){\strut{}$0.8$}}%
    }%
    \gplgaddtomacro\gplfronttext{%
      \csname LTb\endcsname%%
      \put(948,3555){\makebox(0,0){\strut{}143}}%
      \csname LTb\endcsname%%
      \put(1219,3651){\makebox(0,0){\strut{}148}}%
      \csname LTb\endcsname%%
      \put(1491,2230){\makebox(0,0){\strut{}74}}%
      \csname LTb\endcsname%%
      \put(1763,3267){\makebox(0,0){\strut{}128}}%
      \csname LTb\endcsname%%
      \put(2035,2115){\makebox(0,0){\strut{}68}}%
      \csname LTb\endcsname%%
      \put(2306,1961){\makebox(0,0){\strut{}60}}%
      \csname LTb\endcsname%%
      \put(2578,1385){\makebox(0,0){\strut{}30}}%
      \csname LTb\endcsname%%
      \put(2850,1654){\makebox(0,0){\strut{}44}}%
      \csname LTb\endcsname%%
      \put(3122,1596){\makebox(0,0){\strut{}41}}%
      \csname LTb\endcsname%%
      \put(3393,1097){\makebox(0,0){\strut{}15}}%
      \csname LTb\endcsname%%
      \put(3665,1116){\makebox(0,0){\strut{}16}}%
      \csname LTb\endcsname%%
      \put(3937,1039){\makebox(0,0){\strut{}12}}%
      \csname LTb\endcsname%%
      \put(4209,1173){\makebox(0,0){\strut{}19}}%
      \csname LTb\endcsname%%
      \put(4480,1193){\makebox(0,0){\strut{}20}}%
      \csname LTb\endcsname%%
      \put(4752,1462){\makebox(0,0){\strut{}34}}%
      \csname LTb\endcsname%%
      \put(5024,1385){\makebox(0,0){\strut{}30}}%
      \csname LTb\endcsname%%
      \put(5295,1020){\makebox(0,0){\strut{}11}}%
      \csname LTb\endcsname%%
      \put(5567,943){\makebox(0,0){\strut{}7}}%
      \csname LTb\endcsname%%
      \put(5839,962){\makebox(0,0){\strut{}8}}%
      \csname LTb\endcsname%%
      \put(1469,3622){\makebox(0,0)[l]{\strut{}centroid (-0.471)}}%
      \csname LTb\endcsname%%
      \put(161,2202){\rotatebox{-270.00}{\makebox(0,0){\strut{}Normalised density}}}%
      \csname LTb\endcsname%%
      \put(3393,123){\makebox(0,0){\strut{}Sentiment value}}%
      \csname LTb\endcsname%%
      \put(6460,880){\makebox(0,0)[l]{\strut{}$-0.8$}}%
      \csname LTb\endcsname%%
      \put(6460,1211){\makebox(0,0)[l]{\strut{}$-0.6$}}%
      \csname LTb\endcsname%%
      \put(6460,1541){\makebox(0,0)[l]{\strut{}$-0.4$}}%
      \csname LTb\endcsname%%
      \put(6460,1872){\makebox(0,0)[l]{\strut{}$-0.2$}}%
      \csname LTb\endcsname%%
      \put(6460,2202){\makebox(0,0)[l]{\strut{}$0$}}%
      \csname LTb\endcsname%%
      \put(6460,2533){\makebox(0,0)[l]{\strut{}$0.2$}}%
      \csname LTb\endcsname%%
      \put(6460,2863){\makebox(0,0)[l]{\strut{}$0.4$}}%
      \csname LTb\endcsname%%
      \put(6460,3193){\makebox(0,0)[l]{\strut{}$0.6$}}%
      \csname LTb\endcsname%%
      \put(6460,3524){\makebox(0,0)[l]{\strut{}$0.8$}}%
      \csname LTb\endcsname%%
      \put(6900,2202){\rotatebox{-270.00}{\makebox(0,0){\strut{}Sentiment value}}}%
      \csname LTb\endcsname%%
      \put(3393,4036){\makebox(0,0){\strut{}Charles Sobhraj VADER (compound)}}%
    }%
    \gplbacktext
    \put(0,0){\includegraphics[width={360.00bp},height={216.00bp}]{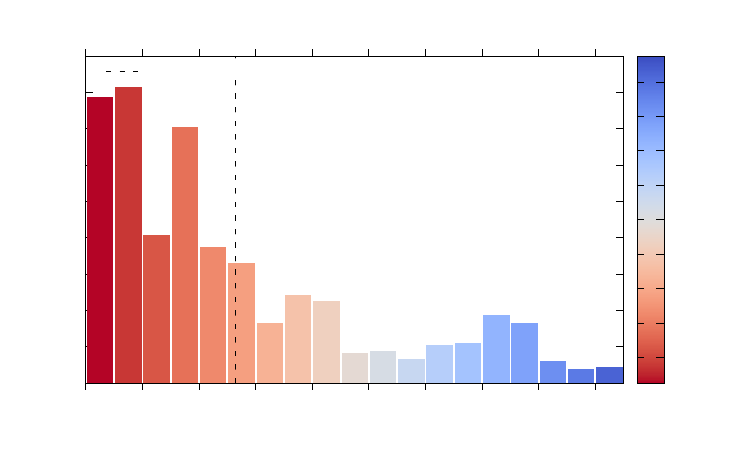}}%
    \gplfronttext
  \end{picture}%
\endgroup

%% file: figures/charles_sobhraj/vader_simple.tex
% GNUPLOT: LaTeX picture with Postscript
\begingroup
  \makeatletter
  \providecommand\color[2][]{%
    \GenericError{(gnuplot) \space\space\space\@spaces}{%
      Package color not loaded in conjunction with
      terminal option `colourtext'%
    }{See the gnuplot documentation for explanation.%
    }{Either use 'blacktext' in gnuplot or load the package
      color.sty in LaTeX.}%
    \renewcommand\color[2][]{}%
  }%
  \providecommand\includegraphics[2][]{%
    \GenericError{(gnuplot) \space\space\space\@spaces}{%
      Package graphicx or graphics not loaded%
    }{See the gnuplot documentation for explanation.%
    }{The gnuplot epslatex terminal needs graphicx.sty or graphics.sty.}%
    \renewcommand\includegraphics[2][]{}%
  }%
  \providecommand\rotatebox[2]{#2}%
  \@ifundefined{ifGPcolor}{%
    \newif\ifGPcolor
    \GPcolortrue
  }{}%
  \@ifundefined{ifGPblacktext}{%
    \newif\ifGPblacktext
    \GPblacktextfalse
  }{}%
  % define a \g@addto@macro without @ in the name:
  \let\gplgaddtomacro\g@addto@macro
  % define empty templates for all commands taking text:
  \gdef\gplbacktext{}%
  \gdef\gplfronttext{}%
  \makeatother
  \ifGPblacktext
    % no textcolor at all
    \def\colorrgb#1{}%
    \def\colorgray#1{}%
  \else
    % gray or color?
    \ifGPcolor
      \def\colorrgb#1{\color[rgb]{#1}}%
      \def\colorgray#1{\color[gray]{#1}}%
      \expandafter\def\csname LTw\endcsname{\color{white}}%
      \expandafter\def\csname LTb\endcsname{\color{black}}%
      \expandafter\def\csname LTa\endcsname{\color{black}}%
      \expandafter\def\csname LT0\endcsname{\color[rgb]{1,0,0}}%
      \expandafter\def\csname LT1\endcsname{\color[rgb]{0,1,0}}%
      \expandafter\def\csname LT2\endcsname{\color[rgb]{0,0,1}}%
      \expandafter\def\csname LT3\endcsname{\color[rgb]{1,0,1}}%
      \expandafter\def\csname LT4\endcsname{\color[rgb]{0,1,1}}%
      \expandafter\def\csname LT5\endcsname{\color[rgb]{1,1,0}}%
      \expandafter\def\csname LT6\endcsname{\color[rgb]{0,0,0}}%
      \expandafter\def\csname LT7\endcsname{\color[rgb]{1,0.3,0}}%
      \expandafter\def\csname LT8\endcsname{\color[rgb]{0.5,0.5,0.5}}%
    \else
      % gray
      \def\colorrgb#1{\color{black}}%
      \def\colorgray#1{\color[gray]{#1}}%
      \expandafter\def\csname LTw\endcsname{\color{white}}%
      \expandafter\def\csname LTb\endcsname{\color{black}}%
      \expandafter\def\csname LTa\endcsname{\color{black}}%
      \expandafter\def\csname LT0\endcsname{\color{black}}%
      \expandafter\def\csname LT1\endcsname{\color{black}}%
      \expandafter\def\csname LT2\endcsname{\color{black}}%
      \expandafter\def\csname LT3\endcsname{\color{black}}%
      \expandafter\def\csname LT4\endcsname{\color{black}}%
      \expandafter\def\csname LT5\endcsname{\color{black}}%
      \expandafter\def\csname LT6\endcsname{\color{black}}%
      \expandafter\def\csname LT7\endcsname{\color{black}}%
      \expandafter\def\csname LT8\endcsname{\color{black}}%
    \fi
  \fi
    \setlength{\unitlength}{0.0500bp}%
    \ifx\gptboxheight\undefined%
      \newlength{\gptboxheight}%
      \newlength{\gptboxwidth}%
      \newsavebox{\gptboxtext}%
    \fi%
    \setlength{\fboxrule}{0.5pt}%
    \setlength{\fboxsep}{1pt}%
    \definecolor{tbcol}{rgb}{1,1,1}%
\begin{picture}(7200.00,4320.00)%
    \gplgaddtomacro\gplbacktext{%
      \csname LTb\endcsname%%
      \put(714,633){\makebox(0,0)[r]{\strut{}$0$}}%
      \csname LTb\endcsname%%
      \put(714,1081){\makebox(0,0)[r]{\strut{}$0.05$}}%
      \csname LTb\endcsname%%
      \put(714,1530){\makebox(0,0)[r]{\strut{}$0.1$}}%
      \csname LTb\endcsname%%
      \put(714,1978){\makebox(0,0)[r]{\strut{}$0.15$}}%
      \csname LTb\endcsname%%
      \put(714,2426){\makebox(0,0)[r]{\strut{}$0.2$}}%
      \csname LTb\endcsname%%
      \put(714,2875){\makebox(0,0)[r]{\strut{}$0.25$}}%
      \csname LTb\endcsname%%
      \put(714,3323){\makebox(0,0)[r]{\strut{}$0.3$}}%
      \csname LTb\endcsname%%
      \put(714,3772){\makebox(0,0)[r]{\strut{}$0.35$}}%
      \csname LTb\endcsname%%
      \put(812,386){\makebox(0,0){\strut{}$-0.7$}}%
      \csname LTb\endcsname%%
      \put(1209,386){\makebox(0,0){\strut{}$-0.6$}}%
      \csname LTb\endcsname%%
      \put(1606,386){\makebox(0,0){\strut{}$-0.5$}}%
      \csname LTb\endcsname%%
      \put(2003,386){\makebox(0,0){\strut{}$-0.4$}}%
      \csname LTb\endcsname%%
      \put(2400,386){\makebox(0,0){\strut{}$-0.3$}}%
      \csname LTb\endcsname%%
      \put(2798,386){\makebox(0,0){\strut{}$-0.2$}}%
      \csname LTb\endcsname%%
      \put(3195,386){\makebox(0,0){\strut{}$-0.1$}}%
      \csname LTb\endcsname%%
      \put(3592,386){\makebox(0,0){\strut{}$0$}}%
      \csname LTb\endcsname%%
      \put(3989,386){\makebox(0,0){\strut{}$0.1$}}%
      \csname LTb\endcsname%%
      \put(4386,386){\makebox(0,0){\strut{}$0.2$}}%
      \csname LTb\endcsname%%
      \put(4783,386){\makebox(0,0){\strut{}$0.3$}}%
      \csname LTb\endcsname%%
      \put(5181,386){\makebox(0,0){\strut{}$0.4$}}%
      \csname LTb\endcsname%%
      \put(5578,386){\makebox(0,0){\strut{}$0.5$}}%
      \csname LTb\endcsname%%
      \put(5975,386){\makebox(0,0){\strut{}$0.6$}}%
    }%
    \gplgaddtomacro\gplfronttext{%
      \csname LTb\endcsname%%
      \put(1010,818){\makebox(0,0){\strut{}1}}%
      \csname LTb\endcsname%%
      \put(1407,858){\makebox(0,0){\strut{}5}}%
      \csname LTb\endcsname%%
      \put(1805,976){\makebox(0,0){\strut{}17}}%
      \csname LTb\endcsname%%
      \put(2202,1678){\makebox(0,0){\strut{}88}}%
      \csname LTb\endcsname%%
      \put(2599,2626){\makebox(0,0){\strut{}184}}%
      \csname LTb\endcsname%%
      \put(2996,3574){\makebox(0,0){\strut{}280}}%
      \csname LTb\endcsname%%
      \put(3393,2468){\makebox(0,0){\strut{}168}}%
      \csname LTb\endcsname%%
      \put(3790,1638){\makebox(0,0){\strut{}84}}%
      \csname LTb\endcsname%%
      \put(4188,1362){\makebox(0,0){\strut{}56}}%
      \csname LTb\endcsname%%
      \put(4585,1006){\makebox(0,0){\strut{}20}}%
      \csname LTb\endcsname%%
      \put(4982,828){\makebox(0,0){\strut{}2}}%
      \csname LTb\endcsname%%
      \put(5379,818){\makebox(0,0){\strut{}1}}%
      \csname LTb\endcsname%%
      \put(5776,828){\makebox(0,0){\strut{}2}}%
      \csname LTb\endcsname%%
      \put(1469,3622){\makebox(0,0)[l]{\strut{}centroid (-0.129)}}%
      \csname LTb\endcsname%%
      \put(161,2202){\rotatebox{-270.00}{\makebox(0,0){\strut{}Normalised density}}}%
      \csname LTb\endcsname%%
      \put(3393,123){\makebox(0,0){\strut{}Sentiment value}}%
      \csname LTb\endcsname%%
      \put(6460,753){\makebox(0,0)[l]{\strut{}$-0.6$}}%
      \csname LTb\endcsname%%
      \put(6460,1236){\makebox(0,0)[l]{\strut{}$-0.4$}}%
      \csname LTb\endcsname%%
      \put(6460,1719){\makebox(0,0)[l]{\strut{}$-0.2$}}%
      \csname LTb\endcsname%%
      \put(6460,2202){\makebox(0,0)[l]{\strut{}$0$}}%
      \csname LTb\endcsname%%
      \put(6460,2685){\makebox(0,0)[l]{\strut{}$0.2$}}%
      \csname LTb\endcsname%%
      \put(6460,3168){\makebox(0,0)[l]{\strut{}$0.4$}}%
      \csname LTb\endcsname%%
      \put(6460,3651){\makebox(0,0)[l]{\strut{}$0.6$}}%
      \csname LTb\endcsname%%
      \put(6900,2202){\rotatebox{-270.00}{\makebox(0,0){\strut{}Sentiment value}}}%
      \csname LTb\endcsname%%
      \put(3393,4036){\makebox(0,0){\strut{}Charles Sobhraj VADER (simple)}}%
    }%
    \gplbacktext
    \put(0,0){\includegraphics[width={360.00bp},height={216.00bp}]{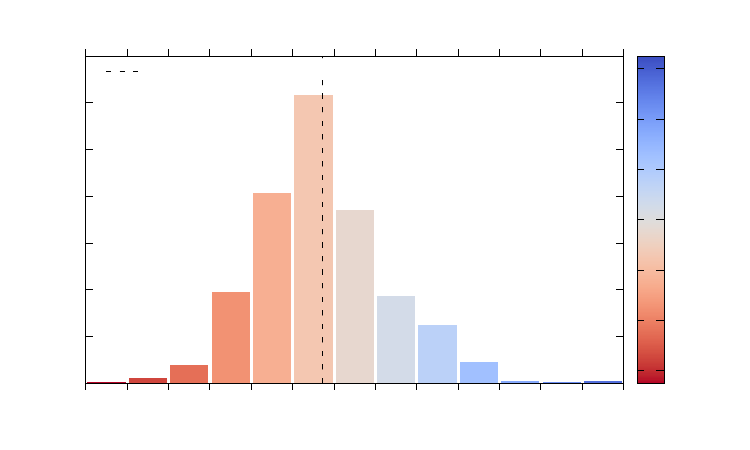}}%
    \gplfronttext
  \end{picture}%
\endgroup

%% file: figures/rolf_harris/afinn.tex
% GNUPLOT: LaTeX picture with Postscript
\begingroup
  \makeatletter
  \providecommand\color[2][]{%
    \GenericError{(gnuplot) \space\space\space\@spaces}{%
      Package color not loaded in conjunction with
      terminal option `colourtext'%
    }{See the gnuplot documentation for explanation.%
    }{Either use 'blacktext' in gnuplot or load the package
      color.sty in LaTeX.}%
    \renewcommand\color[2][]{}%
  }%
  \providecommand\includegraphics[2][]{%
    \GenericError{(gnuplot) \space\space\space\@spaces}{%
      Package graphicx or graphics not loaded%
    }{See the gnuplot documentation for explanation.%
    }{The gnuplot epslatex terminal needs graphicx.sty or graphics.sty.}%
    \renewcommand\includegraphics[2][]{}%
  }%
  \providecommand\rotatebox[2]{#2}%
  \@ifundefined{ifGPcolor}{%
    \newif\ifGPcolor
    \GPcolortrue
  }{}%
  \@ifundefined{ifGPblacktext}{%
    \newif\ifGPblacktext
    \GPblacktextfalse
  }{}%
  % define a \g@addto@macro without @ in the name:
  \let\gplgaddtomacro\g@addto@macro
  % define empty templates for all commands taking text:
  \gdef\gplbacktext{}%
  \gdef\gplfronttext{}%
  \makeatother
  \ifGPblacktext
    % no textcolor at all
    \def\colorrgb#1{}%
    \def\colorgray#1{}%
  \else
    % gray or color?
    \ifGPcolor
      \def\colorrgb#1{\color[rgb]{#1}}%
      \def\colorgray#1{\color[gray]{#1}}%
      \expandafter\def\csname LTw\endcsname{\color{white}}%
      \expandafter\def\csname LTb\endcsname{\color{black}}%
      \expandafter\def\csname LTa\endcsname{\color{black}}%
      \expandafter\def\csname LT0\endcsname{\color[rgb]{1,0,0}}%
      \expandafter\def\csname LT1\endcsname{\color[rgb]{0,1,0}}%
      \expandafter\def\csname LT2\endcsname{\color[rgb]{0,0,1}}%
      \expandafter\def\csname LT3\endcsname{\color[rgb]{1,0,1}}%
      \expandafter\def\csname LT4\endcsname{\color[rgb]{0,1,1}}%
      \expandafter\def\csname LT5\endcsname{\color[rgb]{1,1,0}}%
      \expandafter\def\csname LT6\endcsname{\color[rgb]{0,0,0}}%
      \expandafter\def\csname LT7\endcsname{\color[rgb]{1,0.3,0}}%
      \expandafter\def\csname LT8\endcsname{\color[rgb]{0.5,0.5,0.5}}%
    \else
      % gray
      \def\colorrgb#1{\color{black}}%
      \def\colorgray#1{\color[gray]{#1}}%
      \expandafter\def\csname LTw\endcsname{\color{white}}%
      \expandafter\def\csname LTb\endcsname{\color{black}}%
      \expandafter\def\csname LTa\endcsname{\color{black}}%
      \expandafter\def\csname LT0\endcsname{\color{black}}%
      \expandafter\def\csname LT1\endcsname{\color{black}}%
      \expandafter\def\csname LT2\endcsname{\color{black}}%
      \expandafter\def\csname LT3\endcsname{\color{black}}%
      \expandafter\def\csname LT4\endcsname{\color{black}}%
      \expandafter\def\csname LT5\endcsname{\color{black}}%
      \expandafter\def\csname LT6\endcsname{\color{black}}%
      \expandafter\def\csname LT7\endcsname{\color{black}}%
      \expandafter\def\csname LT8\endcsname{\color{black}}%
    \fi
  \fi
    \setlength{\unitlength}{0.0500bp}%
    \ifx\gptboxheight\undefined%
      \newlength{\gptboxheight}%
      \newlength{\gptboxwidth}%
      \newsavebox{\gptboxtext}%
    \fi%
    \setlength{\fboxrule}{0.5pt}%
    \setlength{\fboxsep}{1pt}%
    \definecolor{tbcol}{rgb}{1,1,1}%
\begin{picture}(7200.00,4320.00)%
    \gplgaddtomacro\gplbacktext{%
      \csname LTb\endcsname%%
      \put(714,633){\makebox(0,0)[r]{\strut{}$0$}}%
      \csname LTb\endcsname%%
      \put(714,1260){\makebox(0,0)[r]{\strut{}$0.05$}}%
      \csname LTb\endcsname%%
      \put(714,1888){\makebox(0,0)[r]{\strut{}$0.1$}}%
      \csname LTb\endcsname%%
      \put(714,2516){\makebox(0,0)[r]{\strut{}$0.15$}}%
      \csname LTb\endcsname%%
      \put(714,3144){\makebox(0,0)[r]{\strut{}$0.2$}}%
      \csname LTb\endcsname%%
      \put(714,3772){\makebox(0,0)[r]{\strut{}$0.25$}}%
      \csname LTb\endcsname%%
      \put(973,386){\makebox(0,0){\strut{}$-14.5$}}%
      \csname LTb\endcsname%%
      \put(1332,386){\makebox(0,0){\strut{}$-12.5$}}%
      \csname LTb\endcsname%%
      \put(1690,386){\makebox(0,0){\strut{}$-10.5$}}%
      \csname LTb\endcsname%%
      \put(2049,386){\makebox(0,0){\strut{}$-8.5$}}%
      \csname LTb\endcsname%%
      \put(2407,386){\makebox(0,0){\strut{}$-6.5$}}%
      \csname LTb\endcsname%%
      \put(2766,386){\makebox(0,0){\strut{}$-4.5$}}%
      \csname LTb\endcsname%%
      \put(3124,386){\makebox(0,0){\strut{}$-2.5$}}%
      \csname LTb\endcsname%%
      \put(3483,386){\makebox(0,0){\strut{}$-0.5$}}%
      \csname LTb\endcsname%%
      \put(3841,386){\makebox(0,0){\strut{}$1.5$}}%
      \csname LTb\endcsname%%
      \put(4200,386){\makebox(0,0){\strut{}$3.5$}}%
      \csname LTb\endcsname%%
      \put(4559,386){\makebox(0,0){\strut{}$5.5$}}%
      \csname LTb\endcsname%%
      \put(4917,386){\makebox(0,0){\strut{}$7.5$}}%
      \csname LTb\endcsname%%
      \put(5276,386){\makebox(0,0){\strut{}$9.5$}}%
      \csname LTb\endcsname%%
      \put(5634,386){\makebox(0,0){\strut{}$11.5$}}%
    }%
    \gplgaddtomacro\gplfronttext{%
      \csname LTb\endcsname%%
      \put(973,846){\makebox(0,0){\strut{}1}}%
      \csname LTb\endcsname%%
      \put(1152,846){\makebox(0,0){\strut{}1}}%
      \csname LTb\endcsname%%
      \put(1869,846){\makebox(0,0){\strut{}1}}%
      \csname LTb\endcsname%%
      \put(2049,957){\makebox(0,0){\strut{}4}}%
      \csname LTb\endcsname%%
      \put(2228,1068){\makebox(0,0){\strut{}7}}%
      \csname LTb\endcsname%%
      \put(2407,957){\makebox(0,0){\strut{}4}}%
      \csname LTb\endcsname%%
      \put(2587,1253){\makebox(0,0){\strut{}12}}%
      \csname LTb\endcsname%%
      \put(2766,1697){\makebox(0,0){\strut{}24}}%
      \csname LTb\endcsname%%
      \put(2945,1549){\makebox(0,0){\strut{}20}}%
      \csname LTb\endcsname%%
      \put(3124,1883){\makebox(0,0){\strut{}29}}%
      \csname LTb\endcsname%%
      \put(3304,3179){\makebox(0,0){\strut{}64}}%
      \csname LTb\endcsname%%
      \put(3483,1697){\makebox(0,0){\strut{}24}}%
      \csname LTb\endcsname%%
      \put(3841,2216){\makebox(0,0){\strut{}38}}%
      \csname LTb\endcsname%%
      \put(4021,2327){\makebox(0,0){\strut{}41}}%
      \csname LTb\endcsname%%
      \put(4200,1957){\makebox(0,0){\strut{}31}}%
      \csname LTb\endcsname%%
      \put(4379,1364){\makebox(0,0){\strut{}15}}%
      \csname LTb\endcsname%%
      \put(4559,1216){\makebox(0,0){\strut{}11}}%
      \csname LTb\endcsname%%
      \put(4738,920){\makebox(0,0){\strut{}3}}%
      \csname LTb\endcsname%%
      \put(4917,920){\makebox(0,0){\strut{}3}}%
      \csname LTb\endcsname%%
      \put(5096,957){\makebox(0,0){\strut{}4}}%
      \csname LTb\endcsname%%
      \put(5814,883){\makebox(0,0){\strut{}2}}%
      \csname LTb\endcsname%%
      \put(1469,3622){\makebox(0,0)[l]{\strut{}centroid (-0.688)}}%
      \csname LTb\endcsname%%
      \put(161,2202){\rotatebox{-270.00}{\makebox(0,0){\strut{}Normalised density}}}%
      \csname LTb\endcsname%%
      \put(3393,123){\makebox(0,0){\strut{}Sentiment value}}%
      \csname LTb\endcsname%%
      \put(6460,1120){\makebox(0,0)[l]{\strut{}$-10$}}%
      \csname LTb\endcsname%%
      \put(6460,1661){\makebox(0,0)[l]{\strut{}$-5$}}%
      \csname LTb\endcsname%%
      \put(6460,2202){\makebox(0,0)[l]{\strut{}$0$}}%
      \csname LTb\endcsname%%
      \put(6460,2743){\makebox(0,0)[l]{\strut{}$5$}}%
      \csname LTb\endcsname%%
      \put(6460,3285){\makebox(0,0)[l]{\strut{}$10$}}%
      \csname LTb\endcsname%%
      \put(6803,2202){\rotatebox{-270.00}{\makebox(0,0){\strut{}Sentiment value}}}%
      \csname LTb\endcsname%%
      \put(3393,4036){\makebox(0,0){\strut{}Rolf Harris AFINN}}%
    }%
    \gplbacktext
    \put(0,0){\includegraphics[width={360.00bp},height={216.00bp}]{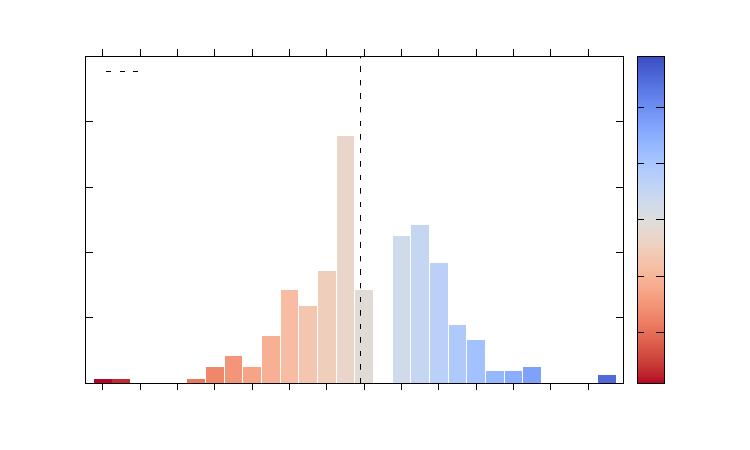}}%
    \gplfronttext
  \end{picture}%
\endgroup

%% file: figures/rolf_harris/minos.tex
% GNUPLOT: LaTeX picture with Postscript
\begingroup
  \makeatletter
  \providecommand\color[2][]{%
    \GenericError{(gnuplot) \space\space\space\@spaces}{%
      Package color not loaded in conjunction with
      terminal option `colourtext'%
    }{See the gnuplot documentation for explanation.%
    }{Either use 'blacktext' in gnuplot or load the package
      color.sty in LaTeX.}%
    \renewcommand\color[2][]{}%
  }%
  \providecommand\includegraphics[2][]{%
    \GenericError{(gnuplot) \space\space\space\@spaces}{%
      Package graphicx or graphics not loaded%
    }{See the gnuplot documentation for explanation.%
    }{The gnuplot epslatex terminal needs graphicx.sty or graphics.sty.}%
    \renewcommand\includegraphics[2][]{}%
  }%
  \providecommand\rotatebox[2]{#2}%
  \@ifundefined{ifGPcolor}{%
    \newif\ifGPcolor
    \GPcolortrue
  }{}%
  \@ifundefined{ifGPblacktext}{%
    \newif\ifGPblacktext
    \GPblacktextfalse
  }{}%
  % define a \g@addto@macro without @ in the name:
  \let\gplgaddtomacro\g@addto@macro
  % define empty templates for all commands taking text:
  \gdef\gplbacktext{}%
  \gdef\gplfronttext{}%
  \makeatother
  \ifGPblacktext
    % no textcolor at all
    \def\colorrgb#1{}%
    \def\colorgray#1{}%
  \else
    % gray or color?
    \ifGPcolor
      \def\colorrgb#1{\color[rgb]{#1}}%
      \def\colorgray#1{\color[gray]{#1}}%
      \expandafter\def\csname LTw\endcsname{\color{white}}%
      \expandafter\def\csname LTb\endcsname{\color{black}}%
      \expandafter\def\csname LTa\endcsname{\color{black}}%
      \expandafter\def\csname LT0\endcsname{\color[rgb]{1,0,0}}%
      \expandafter\def\csname LT1\endcsname{\color[rgb]{0,1,0}}%
      \expandafter\def\csname LT2\endcsname{\color[rgb]{0,0,1}}%
      \expandafter\def\csname LT3\endcsname{\color[rgb]{1,0,1}}%
      \expandafter\def\csname LT4\endcsname{\color[rgb]{0,1,1}}%
      \expandafter\def\csname LT5\endcsname{\color[rgb]{1,1,0}}%
      \expandafter\def\csname LT6\endcsname{\color[rgb]{0,0,0}}%
      \expandafter\def\csname LT7\endcsname{\color[rgb]{1,0.3,0}}%
      \expandafter\def\csname LT8\endcsname{\color[rgb]{0.5,0.5,0.5}}%
    \else
      % gray
      \def\colorrgb#1{\color{black}}%
      \def\colorgray#1{\color[gray]{#1}}%
      \expandafter\def\csname LTw\endcsname{\color{white}}%
      \expandafter\def\csname LTb\endcsname{\color{black}}%
      \expandafter\def\csname LTa\endcsname{\color{black}}%
      \expandafter\def\csname LT0\endcsname{\color{black}}%
      \expandafter\def\csname LT1\endcsname{\color{black}}%
      \expandafter\def\csname LT2\endcsname{\color{black}}%
      \expandafter\def\csname LT3\endcsname{\color{black}}%
      \expandafter\def\csname LT4\endcsname{\color{black}}%
      \expandafter\def\csname LT5\endcsname{\color{black}}%
      \expandafter\def\csname LT6\endcsname{\color{black}}%
      \expandafter\def\csname LT7\endcsname{\color{black}}%
      \expandafter\def\csname LT8\endcsname{\color{black}}%
    \fi
  \fi
    \setlength{\unitlength}{0.0500bp}%
    \ifx\gptboxheight\undefined%
      \newlength{\gptboxheight}%
      \newlength{\gptboxwidth}%
      \newsavebox{\gptboxtext}%
    \fi%
    \setlength{\fboxrule}{0.5pt}%
    \setlength{\fboxsep}{1pt}%
    \definecolor{tbcol}{rgb}{1,1,1}%
\begin{picture}(7200.00,4320.00)%
    \gplgaddtomacro\gplbacktext{%
      \csname LTb\endcsname%%
      \put(714,633){\makebox(0,0)[r]{\strut{}$0$}}%
      \csname LTb\endcsname%%
      \put(714,1081){\makebox(0,0)[r]{\strut{}$0.05$}}%
      \csname LTb\endcsname%%
      \put(714,1530){\makebox(0,0)[r]{\strut{}$0.1$}}%
      \csname LTb\endcsname%%
      \put(714,1978){\makebox(0,0)[r]{\strut{}$0.15$}}%
      \csname LTb\endcsname%%
      \put(714,2426){\makebox(0,0)[r]{\strut{}$0.2$}}%
      \csname LTb\endcsname%%
      \put(714,2875){\makebox(0,0)[r]{\strut{}$0.25$}}%
      \csname LTb\endcsname%%
      \put(714,3323){\makebox(0,0)[r]{\strut{}$0.3$}}%
      \csname LTb\endcsname%%
      \put(714,3772){\makebox(0,0)[r]{\strut{}$0.35$}}%
      \csname LTb\endcsname%%
      \put(1175,386){\makebox(0,0){\strut{}$-5.5$}}%
      \csname LTb\endcsname%%
      \put(1578,386){\makebox(0,0){\strut{}$-4.5$}}%
      \csname LTb\endcsname%%
      \put(1982,386){\makebox(0,0){\strut{}$-3.5$}}%
      \csname LTb\endcsname%%
      \put(2385,386){\makebox(0,0){\strut{}$-2.5$}}%
      \csname LTb\endcsname%%
      \put(2788,386){\makebox(0,0){\strut{}$-1.5$}}%
      \csname LTb\endcsname%%
      \put(3192,386){\makebox(0,0){\strut{}$-0.5$}}%
      \csname LTb\endcsname%%
      \put(3595,386){\makebox(0,0){\strut{}$0.5$}}%
      \csname LTb\endcsname%%
      \put(3998,386){\makebox(0,0){\strut{}$1.5$}}%
      \csname LTb\endcsname%%
      \put(4402,386){\makebox(0,0){\strut{}$2.5$}}%
      \csname LTb\endcsname%%
      \put(4805,386){\makebox(0,0){\strut{}$3.5$}}%
      \csname LTb\endcsname%%
      \put(5208,386){\makebox(0,0){\strut{}$4.5$}}%
      \csname LTb\endcsname%%
      \put(5612,386){\makebox(0,0){\strut{}$5.5$}}%
    }%
    \gplgaddtomacro\gplfronttext{%
      \csname LTb\endcsname%%
      \put(1175,837){\makebox(0,0){\strut{}1}}%
      \csname LTb\endcsname%%
      \put(1578,866){\makebox(0,0){\strut{}2}}%
      \csname LTb\endcsname%%
      \put(1982,1066){\makebox(0,0){\strut{}9}}%
      \csname LTb\endcsname%%
      \put(2385,1439){\makebox(0,0){\strut{}22}}%
      \csname LTb\endcsname%%
      \put(2788,2127){\makebox(0,0){\strut{}46}}%
      \csname LTb\endcsname%%
      \put(3192,3473){\makebox(0,0){\strut{}93}}%
      \csname LTb\endcsname%%
      \put(3998,3273){\makebox(0,0){\strut{}86}}%
      \csname LTb\endcsname%%
      \put(4402,1869){\makebox(0,0){\strut{}37}}%
      \csname LTb\endcsname%%
      \put(4805,1095){\makebox(0,0){\strut{}10}}%
      \csname LTb\endcsname%%
      \put(5208,980){\makebox(0,0){\strut{}6}}%
      \csname LTb\endcsname%%
      \put(5612,837){\makebox(0,0){\strut{}1}}%
      \csname LTb\endcsname%%
      \put(1469,3622){\makebox(0,0)[l]{\strut{}centroid (-0.248)}}%
      \csname LTb\endcsname%%
      \put(161,2202){\rotatebox{-270.00}{\makebox(0,0){\strut{}Normalised density}}}%
      \csname LTb\endcsname%%
      \put(3393,123){\makebox(0,0){\strut{}Sentiment value}}%
      \csname LTb\endcsname%%
      \put(6460,1061){\makebox(0,0)[l]{\strut{}$-4$}}%
      \csname LTb\endcsname%%
      \put(6460,1631){\makebox(0,0)[l]{\strut{}$-2$}}%
      \csname LTb\endcsname%%
      \put(6460,2202){\makebox(0,0)[l]{\strut{}$0$}}%
      \csname LTb\endcsname%%
      \put(6460,2773){\makebox(0,0)[l]{\strut{}$2$}}%
      \csname LTb\endcsname%%
      \put(6460,3344){\makebox(0,0)[l]{\strut{}$4$}}%
      \csname LTb\endcsname%%
      \put(6705,2202){\rotatebox{-270.00}{\makebox(0,0){\strut{}Sentiment value}}}%
      \csname LTb\endcsname%%
      \put(3393,4036){\makebox(0,0){\strut{}Rolf Harris MINOS}}%
    }%
    \gplbacktext
    \put(0,0){\includegraphics[width={360.00bp},height={216.00bp}]{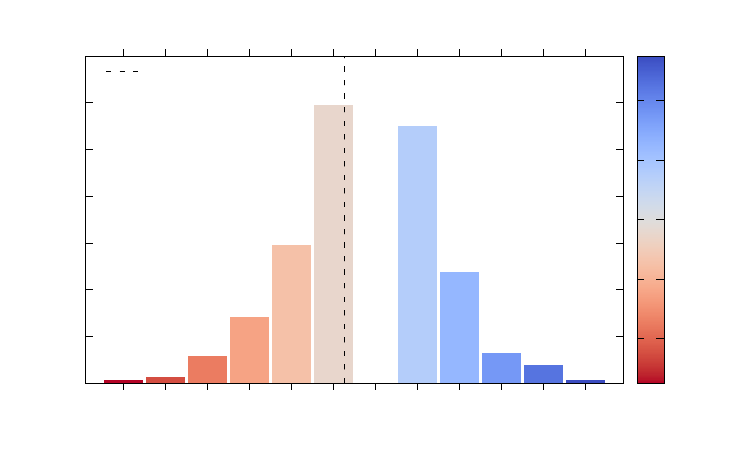}}%
    \gplfronttext
  \end{picture}%
\endgroup

%% file: figures/rolf_harris/vader_compound.tex
% GNUPLOT: LaTeX picture with Postscript
\begingroup
  \makeatletter
  \providecommand\color[2][]{%
    \GenericError{(gnuplot) \space\space\space\@spaces}{%
      Package color not loaded in conjunction with
      terminal option `colourtext'%
    }{See the gnuplot documentation for explanation.%
    }{Either use 'blacktext' in gnuplot or load the package
      color.sty in LaTeX.}%
    \renewcommand\color[2][]{}%
  }%
  \providecommand\includegraphics[2][]{%
    \GenericError{(gnuplot) \space\space\space\@spaces}{%
      Package graphicx or graphics not loaded%
    }{See the gnuplot documentation for explanation.%
    }{The gnuplot epslatex terminal needs graphicx.sty or graphics.sty.}%
    \renewcommand\includegraphics[2][]{}%
  }%
  \providecommand\rotatebox[2]{#2}%
  \@ifundefined{ifGPcolor}{%
    \newif\ifGPcolor
    \GPcolortrue
  }{}%
  \@ifundefined{ifGPblacktext}{%
    \newif\ifGPblacktext
    \GPblacktextfalse
  }{}%
  % define a \g@addto@macro without @ in the name:
  \let\gplgaddtomacro\g@addto@macro
  % define empty templates for all commands taking text:
  \gdef\gplbacktext{}%
  \gdef\gplfronttext{}%
  \makeatother
  \ifGPblacktext
    % no textcolor at all
    \def\colorrgb#1{}%
    \def\colorgray#1{}%
  \else
    % gray or color?
    \ifGPcolor
      \def\colorrgb#1{\color[rgb]{#1}}%
      \def\colorgray#1{\color[gray]{#1}}%
      \expandafter\def\csname LTw\endcsname{\color{white}}%
      \expandafter\def\csname LTb\endcsname{\color{black}}%
      \expandafter\def\csname LTa\endcsname{\color{black}}%
      \expandafter\def\csname LT0\endcsname{\color[rgb]{1,0,0}}%
      \expandafter\def\csname LT1\endcsname{\color[rgb]{0,1,0}}%
      \expandafter\def\csname LT2\endcsname{\color[rgb]{0,0,1}}%
      \expandafter\def\csname LT3\endcsname{\color[rgb]{1,0,1}}%
      \expandafter\def\csname LT4\endcsname{\color[rgb]{0,1,1}}%
      \expandafter\def\csname LT5\endcsname{\color[rgb]{1,1,0}}%
      \expandafter\def\csname LT6\endcsname{\color[rgb]{0,0,0}}%
      \expandafter\def\csname LT7\endcsname{\color[rgb]{1,0.3,0}}%
      \expandafter\def\csname LT8\endcsname{\color[rgb]{0.5,0.5,0.5}}%
    \else
      % gray
      \def\colorrgb#1{\color{black}}%
      \def\colorgray#1{\color[gray]{#1}}%
      \expandafter\def\csname LTw\endcsname{\color{white}}%
      \expandafter\def\csname LTb\endcsname{\color{black}}%
      \expandafter\def\csname LTa\endcsname{\color{black}}%
      \expandafter\def\csname LT0\endcsname{\color{black}}%
      \expandafter\def\csname LT1\endcsname{\color{black}}%
      \expandafter\def\csname LT2\endcsname{\color{black}}%
      \expandafter\def\csname LT3\endcsname{\color{black}}%
      \expandafter\def\csname LT4\endcsname{\color{black}}%
      \expandafter\def\csname LT5\endcsname{\color{black}}%
      \expandafter\def\csname LT6\endcsname{\color{black}}%
      \expandafter\def\csname LT7\endcsname{\color{black}}%
      \expandafter\def\csname LT8\endcsname{\color{black}}%
    \fi
  \fi
    \setlength{\unitlength}{0.0500bp}%
    \ifx\gptboxheight\undefined%
      \newlength{\gptboxheight}%
      \newlength{\gptboxwidth}%
      \newsavebox{\gptboxtext}%
    \fi%
    \setlength{\fboxrule}{0.5pt}%
    \setlength{\fboxsep}{1pt}%
    \definecolor{tbcol}{rgb}{1,1,1}%
\begin{picture}(7200.00,4320.00)%
    \gplgaddtomacro\gplbacktext{%
      \csname LTb\endcsname%%
      \put(714,633){\makebox(0,0)[r]{\strut{}$0$}}%
      \csname LTb\endcsname%%
      \put(714,1081){\makebox(0,0)[r]{\strut{}$0.02$}}%
      \csname LTb\endcsname%%
      \put(714,1530){\makebox(0,0)[r]{\strut{}$0.04$}}%
      \csname LTb\endcsname%%
      \put(714,1978){\makebox(0,0)[r]{\strut{}$0.06$}}%
      \csname LTb\endcsname%%
      \put(714,2426){\makebox(0,0)[r]{\strut{}$0.08$}}%
      \csname LTb\endcsname%%
      \put(714,2875){\makebox(0,0)[r]{\strut{}$0.1$}}%
      \csname LTb\endcsname%%
      \put(714,3323){\makebox(0,0)[r]{\strut{}$0.12$}}%
      \csname LTb\endcsname%%
      \put(714,3772){\makebox(0,0)[r]{\strut{}$0.14$}}%
      \csname LTb\endcsname%%
      \put(812,386){\makebox(0,0){\strut{}$-1.05$}}%
      \csname LTb\endcsname%%
      \put(1303,386){\makebox(0,0){\strut{}$-0.85$}}%
      \csname LTb\endcsname%%
      \put(1795,386){\makebox(0,0){\strut{}$-0.65$}}%
      \csname LTb\endcsname%%
      \put(2287,386){\makebox(0,0){\strut{}$-0.45$}}%
      \csname LTb\endcsname%%
      \put(2779,386){\makebox(0,0){\strut{}$-0.25$}}%
      \csname LTb\endcsname%%
      \put(3270,386){\makebox(0,0){\strut{}$-0.05$}}%
      \csname LTb\endcsname%%
      \put(3762,386){\makebox(0,0){\strut{}$0.15$}}%
      \csname LTb\endcsname%%
      \put(4254,386){\makebox(0,0){\strut{}$0.35$}}%
      \csname LTb\endcsname%%
      \put(4746,386){\makebox(0,0){\strut{}$0.55$}}%
      \csname LTb\endcsname%%
      \put(5237,386){\makebox(0,0){\strut{}$0.75$}}%
      \csname LTb\endcsname%%
      \put(5729,386){\makebox(0,0){\strut{}$0.95$}}%
    }%
    \gplgaddtomacro\gplfronttext{%
      \csname LTb\endcsname%%
      \put(1181,972){\makebox(0,0){\strut{}3}}%
      \csname LTb\endcsname%%
      \put(1426,1681){\makebox(0,0){\strut{}16}}%
      \csname LTb\endcsname%%
      \put(1672,2227){\makebox(0,0){\strut{}26}}%
      \csname LTb\endcsname%%
      \put(1918,1572){\makebox(0,0){\strut{}14}}%
      \csname LTb\endcsname%%
      \put(2164,2336){\makebox(0,0){\strut{}28}}%
      \csname LTb\endcsname%%
      \put(2410,2063){\makebox(0,0){\strut{}23}}%
      \csname LTb\endcsname%%
      \put(2656,1681){\makebox(0,0){\strut{}16}}%
      \csname LTb\endcsname%%
      \put(2902,1736){\makebox(0,0){\strut{}17}}%
      \csname LTb\endcsname%%
      \put(3147,1518){\makebox(0,0){\strut{}13}}%
      \csname LTb\endcsname%%
      \put(3393,1409){\makebox(0,0){\strut{}11}}%
      \csname LTb\endcsname%%
      \put(3639,1354){\makebox(0,0){\strut{}10}}%
      \csname LTb\endcsname%%
      \put(3885,1463){\makebox(0,0){\strut{}12}}%
      \csname LTb\endcsname%%
      \put(4131,2282){\makebox(0,0){\strut{}27}}%
      \csname LTb\endcsname%%
      \put(4377,3209){\makebox(0,0){\strut{}44}}%
      \csname LTb\endcsname%%
      \put(4623,3427){\makebox(0,0){\strut{}48}}%
      \csname LTb\endcsname%%
      \put(4868,2227){\makebox(0,0){\strut{}26}}%
      \csname LTb\endcsname%%
      \put(5114,2336){\makebox(0,0){\strut{}28}}%
      \csname LTb\endcsname%%
      \put(5360,2118){\makebox(0,0){\strut{}24}}%
      \csname LTb\endcsname%%
      \put(5606,1736){\makebox(0,0){\strut{}17}}%
      \csname LTb\endcsname%%
      \put(5852,1245){\makebox(0,0){\strut{}8}}%
      \csname LTb\endcsname%%
      \put(1469,3622){\makebox(0,0)[l]{\strut{}centroid (+0.087)}}%
      \csname LTb\endcsname%%
      \put(161,2202){\rotatebox{-270.00}{\makebox(0,0){\strut{}Normalised density}}}%
      \csname LTb\endcsname%%
      \put(3393,123){\makebox(0,0){\strut{}Sentiment value}}%
      \csname LTb\endcsname%%
      \put(6460,633){\makebox(0,0)[l]{\strut{}$-1$}}%
      \csname LTb\endcsname%%
      \put(6460,1417){\makebox(0,0)[l]{\strut{}$-0.5$}}%
      \csname LTb\endcsname%%
      \put(6460,2202){\makebox(0,0)[l]{\strut{}$0$}}%
      \csname LTb\endcsname%%
      \put(6460,2987){\makebox(0,0)[l]{\strut{}$0.5$}}%
      \csname LTb\endcsname%%
      \put(6460,3772){\makebox(0,0)[l]{\strut{}$1$}}%
      \csname LTb\endcsname%%
      \put(6900,2202){\rotatebox{-270.00}{\makebox(0,0){\strut{}Sentiment value}}}%
      \csname LTb\endcsname%%
      \put(3393,4036){\makebox(0,0){\strut{}Rolf Harris VADER (compound)}}%
    }%
    \gplbacktext
    \put(0,0){\includegraphics[width={360.00bp},height={216.00bp}]{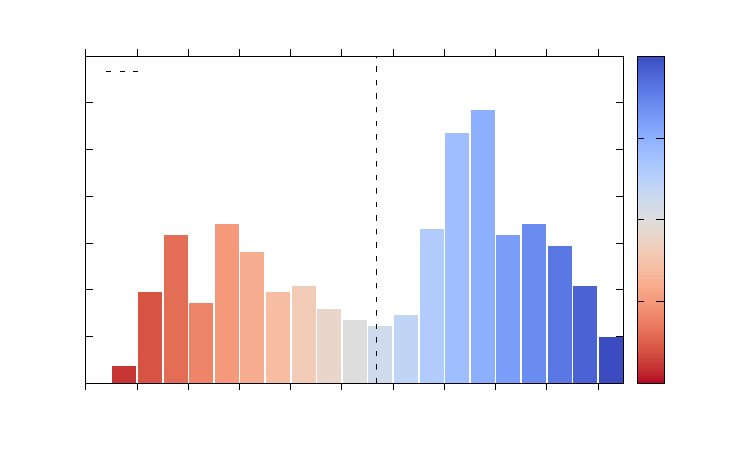}}%
    \gplfronttext
  \end{picture}%
\endgroup

%% file: figures/rolf_harris/vader_simple.tex
% GNUPLOT: LaTeX picture with Postscript
\begingroup
  \makeatletter
  \providecommand\color[2][]{%
    \GenericError{(gnuplot) \space\space\space\@spaces}{%
      Package color not loaded in conjunction with
      terminal option `colourtext'%
    }{See the gnuplot documentation for explanation.%
    }{Either use 'blacktext' in gnuplot or load the package
      color.sty in LaTeX.}%
    \renewcommand\color[2][]{}%
  }%
  \providecommand\includegraphics[2][]{%
    \GenericError{(gnuplot) \space\space\space\@spaces}{%
      Package graphicx or graphics not loaded%
    }{See the gnuplot documentation for explanation.%
    }{The gnuplot epslatex terminal needs graphicx.sty or graphics.sty.}%
    \renewcommand\includegraphics[2][]{}%
  }%
  \providecommand\rotatebox[2]{#2}%
  \@ifundefined{ifGPcolor}{%
    \newif\ifGPcolor
    \GPcolortrue
  }{}%
  \@ifundefined{ifGPblacktext}{%
    \newif\ifGPblacktext
    \GPblacktextfalse
  }{}%
  % define a \g@addto@macro without @ in the name:
  \let\gplgaddtomacro\g@addto@macro
  % define empty templates for all commands taking text:
  \gdef\gplbacktext{}%
  \gdef\gplfronttext{}%
  \makeatother
  \ifGPblacktext
    % no textcolor at all
    \def\colorrgb#1{}%
    \def\colorgray#1{}%
  \else
    % gray or color?
    \ifGPcolor
      \def\colorrgb#1{\color[rgb]{#1}}%
      \def\colorgray#1{\color[gray]{#1}}%
      \expandafter\def\csname LTw\endcsname{\color{white}}%
      \expandafter\def\csname LTb\endcsname{\color{black}}%
      \expandafter\def\csname LTa\endcsname{\color{black}}%
      \expandafter\def\csname LT0\endcsname{\color[rgb]{1,0,0}}%
      \expandafter\def\csname LT1\endcsname{\color[rgb]{0,1,0}}%
      \expandafter\def\csname LT2\endcsname{\color[rgb]{0,0,1}}%
      \expandafter\def\csname LT3\endcsname{\color[rgb]{1,0,1}}%
      \expandafter\def\csname LT4\endcsname{\color[rgb]{0,1,1}}%
      \expandafter\def\csname LT5\endcsname{\color[rgb]{1,1,0}}%
      \expandafter\def\csname LT6\endcsname{\color[rgb]{0,0,0}}%
      \expandafter\def\csname LT7\endcsname{\color[rgb]{1,0.3,0}}%
      \expandafter\def\csname LT8\endcsname{\color[rgb]{0.5,0.5,0.5}}%
    \else
      % gray
      \def\colorrgb#1{\color{black}}%
      \def\colorgray#1{\color[gray]{#1}}%
      \expandafter\def\csname LTw\endcsname{\color{white}}%
      \expandafter\def\csname LTb\endcsname{\color{black}}%
      \expandafter\def\csname LTa\endcsname{\color{black}}%
      \expandafter\def\csname LT0\endcsname{\color{black}}%
      \expandafter\def\csname LT1\endcsname{\color{black}}%
      \expandafter\def\csname LT2\endcsname{\color{black}}%
      \expandafter\def\csname LT3\endcsname{\color{black}}%
      \expandafter\def\csname LT4\endcsname{\color{black}}%
      \expandafter\def\csname LT5\endcsname{\color{black}}%
      \expandafter\def\csname LT6\endcsname{\color{black}}%
      \expandafter\def\csname LT7\endcsname{\color{black}}%
      \expandafter\def\csname LT8\endcsname{\color{black}}%
    \fi
  \fi
    \setlength{\unitlength}{0.0500bp}%
    \ifx\gptboxheight\undefined%
      \newlength{\gptboxheight}%
      \newlength{\gptboxwidth}%
      \newsavebox{\gptboxtext}%
    \fi%
    \setlength{\fboxrule}{0.5pt}%
    \setlength{\fboxsep}{1pt}%
    \definecolor{tbcol}{rgb}{1,1,1}%
\begin{picture}(7200.00,4320.00)%
    \gplgaddtomacro\gplbacktext{%
      \csname LTb\endcsname%%
      \put(714,633){\makebox(0,0)[r]{\strut{}$0$}}%
      \csname LTb\endcsname%%
      \put(714,1156){\makebox(0,0)[r]{\strut{}$0.05$}}%
      \csname LTb\endcsname%%
      \put(714,1679){\makebox(0,0)[r]{\strut{}$0.1$}}%
      \csname LTb\endcsname%%
      \put(714,2202){\makebox(0,0)[r]{\strut{}$0.15$}}%
      \csname LTb\endcsname%%
      \put(714,2725){\makebox(0,0)[r]{\strut{}$0.2$}}%
      \csname LTb\endcsname%%
      \put(714,3249){\makebox(0,0)[r]{\strut{}$0.25$}}%
      \csname LTb\endcsname%%
      \put(714,3772){\makebox(0,0)[r]{\strut{}$0.3$}}%
      \csname LTb\endcsname%%
      \put(812,386){\makebox(0,0){\strut{}$-0.5$}}%
      \csname LTb\endcsname%%
      \put(1242,386){\makebox(0,0){\strut{}$-0.4$}}%
      \csname LTb\endcsname%%
      \put(1672,386){\makebox(0,0){\strut{}$-0.3$}}%
      \csname LTb\endcsname%%
      \put(2103,386){\makebox(0,0){\strut{}$-0.2$}}%
      \csname LTb\endcsname%%
      \put(2533,386){\makebox(0,0){\strut{}$-0.1$}}%
      \csname LTb\endcsname%%
      \put(2963,386){\makebox(0,0){\strut{}$0$}}%
      \csname LTb\endcsname%%
      \put(3393,386){\makebox(0,0){\strut{}$0.1$}}%
      \csname LTb\endcsname%%
      \put(3824,386){\makebox(0,0){\strut{}$0.2$}}%
      \csname LTb\endcsname%%
      \put(4254,386){\makebox(0,0){\strut{}$0.3$}}%
      \csname LTb\endcsname%%
      \put(4684,386){\makebox(0,0){\strut{}$0.4$}}%
      \csname LTb\endcsname%%
      \put(5114,386){\makebox(0,0){\strut{}$0.5$}}%
      \csname LTb\endcsname%%
      \put(5545,386){\makebox(0,0){\strut{}$0.6$}}%
      \csname LTb\endcsname%%
      \put(5975,386){\makebox(0,0){\strut{}$0.7$}}%
    }%
    \gplgaddtomacro\gplfronttext{%
      \csname LTb\endcsname%%
      \put(1027,834){\makebox(0,0){\strut{}1}}%
      \csname LTb\endcsname%%
      \put(1457,1190){\makebox(0,0){\strut{}15}}%
      \csname LTb\endcsname%%
      \put(1887,1496){\makebox(0,0){\strut{}27}}%
      \csname LTb\endcsname%%
      \put(2318,2387){\makebox(0,0){\strut{}62}}%
      \csname LTb\endcsname%%
      \put(2748,2336){\makebox(0,0){\strut{}60}}%
      \csname LTb\endcsname%%
      \put(3178,3304){\makebox(0,0){\strut{}98}}%
      \csname LTb\endcsname%%
      \put(3608,3278){\makebox(0,0){\strut{}97}}%
      \csname LTb\endcsname%%
      \put(4039,1751){\makebox(0,0){\strut{}37}}%
      \csname LTb\endcsname%%
      \put(4469,1012){\makebox(0,0){\strut{}8}}%
      \csname LTb\endcsname%%
      \put(4899,936){\makebox(0,0){\strut{}5}}%
      \csname LTb\endcsname%%
      \put(5760,834){\makebox(0,0){\strut{}1}}%
      \csname LTb\endcsname%%
      \put(1469,3622){\makebox(0,0)[l]{\strut{}centroid (+0.023)}}%
      \csname LTb\endcsname%%
      \put(161,2202){\rotatebox{-270.00}{\makebox(0,0){\strut{}Normalised density}}}%
      \csname LTb\endcsname%%
      \put(3393,123){\makebox(0,0){\strut{}Sentiment value}}%
      \csname LTb\endcsname%%
      \put(6460,753){\makebox(0,0)[l]{\strut{}$-0.6$}}%
      \csname LTb\endcsname%%
      \put(6460,1236){\makebox(0,0)[l]{\strut{}$-0.4$}}%
      \csname LTb\endcsname%%
      \put(6460,1719){\makebox(0,0)[l]{\strut{}$-0.2$}}%
      \csname LTb\endcsname%%
      \put(6460,2202){\makebox(0,0)[l]{\strut{}$0$}}%
      \csname LTb\endcsname%%
      \put(6460,2685){\makebox(0,0)[l]{\strut{}$0.2$}}%
      \csname LTb\endcsname%%
      \put(6460,3168){\makebox(0,0)[l]{\strut{}$0.4$}}%
      \csname LTb\endcsname%%
      \put(6460,3651){\makebox(0,0)[l]{\strut{}$0.6$}}%
      \csname LTb\endcsname%%
      \put(6900,2202){\rotatebox{-270.00}{\makebox(0,0){\strut{}Sentiment value}}}%
      \csname LTb\endcsname%%
      \put(3393,4036){\makebox(0,0){\strut{}Rolf Harris VADER (simple)}}%
    }%
    \gplbacktext
    \put(0,0){\includegraphics[width={360.00bp},height={216.00bp}]{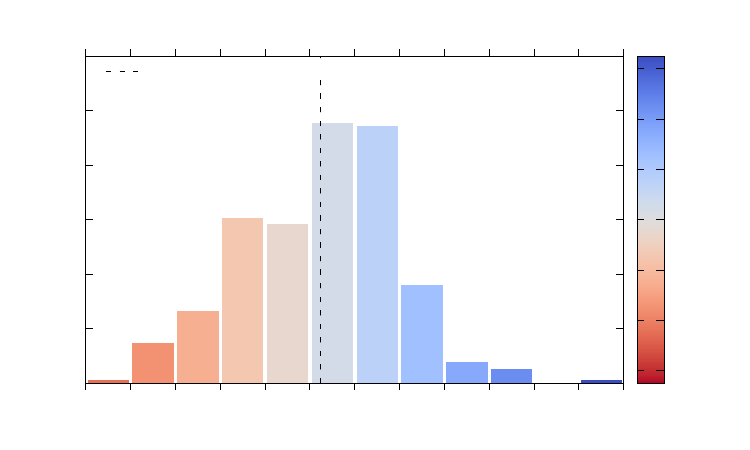}}%
    \gplfronttext
  \end{picture}%
\endgroup

%% file: conclusions_and_future_work.tex
\section{Conclusions and Future Work}
\label{sec:conclusions}

We utilised several natural language processing techniques to maximise the information available to a human sifter:
\begin{itemize}
	\item Web-scraping allows us to collect attributable, open-source information at scale
	\item Tokenisation divided articles of up to $\sim 10^7$ characters\footnote{This is approximately the length of the entire \textit{Harry Potter} series, impossible for a human to consume at pace.} into individual sense units, enabling us to discard irrelevant information
	\item Co-reference resolution was critical to identifying all information relevant to the \soi{}, without which we would have thrown away signal as well as noise, creating misleading results
	\item Sentiment analysis provided a rich range of posterior sentiment distributions which allowed us to make informed judgements on which algorithm works most effectively across all \soi{} groups
\end{itemize}
Bayesian inference broadly allowed us to create a summary score for each \soi{} from a full awareness of our prior assumptions.  Despite inherent limitations regarding data accessibility, our methodology presents a robust, scalable solution capable of addressing the challenges inherent in maintaining public trust in the Honours system. It is accepted that the results from this work might have some biases based on the articles about any individual found on the internet. Therefore, a human-in-the-loop 
is essential to make the final decisions regarding an award.
Future research directions include:
\begin{itemize}
	\item periodic reassessment of current Honours recipients to proactively detect emerging risks
	\item broadening the intelligence picture via better access to news services and subscription-based platforms
	\item refining our \minos{} sentiment model to better capture the context of the sentence as a whole (similar to \vader{} compound)
	\item expanding \minos{}'s vocabulary via semantic embeddings, which would allow us to assign positive and negative scores to synonyms of the current wordlist
	\item exploring the applicability of our methodology to other awards and recognition frameworks
\end{itemize}
This provides diverse opportunities to capitalise on the promising results in this proof-of-concept.

Our research highlights the potential for data science to significantly enhance the Honours System.  It increases the efficiency and transparency of the selection process by gathering the most positive and negative information available in the public domain.  Unlike the manual vetting approach used currently, our parallel, automated methodology can scale to the $\sim 200\,000$ living recipients of Honours.  This opens the possibility to assess forfeiture risk dynamically, protecting the integrity of the Honours System as the information picture of recipients changes.  The objective is not to replace human decision-making but to provide committees with an impartial and thorough intelligence picture, from which to make informed judgments.  Our approach promises significant improvements in fairness, transparency, and public confidence across various high-profile applications.